\documentclass{article}

\usepackage[main,final]{neurips_2026}

\usepackage[utf8]{inputenc} 
\usepackage[T1]{fontenc}    
\usepackage{hyperref}       
\usepackage{url}            
\usepackage{booktabs}       
\usepackage{amsfonts}       
\usepackage{nicefrac}       
\usepackage{microtype}      
\usepackage{xcolor}         

\usepackage{amsmath}
\usepackage{amssymb}
\usepackage{mathtools}
\usepackage{amsthm}
\usepackage[capitalise]{cleveref} 
\usepackage{tikz}
\usetikzlibrary{positioning,arrows.meta}

\title{Accuracy vs.\ Accuracy: Computational Tradeoffs Between Classification Rates and Utility}

\author{Noga Amit\thanks{Authors listed alphabetically.} \\
    MIT \\
    \And
    Omer Reingold\footnotemark[1]{} \\
  Stanford \\
  \And
   Guy N. Rothblum\footnotemark[1]{} \\
   Apple \\
}

\newtheorem{theorem}{Theorem}[section]       
\newtheorem{definition}[theorem]{Definition}
\newtheorem{corollary}[theorem]{Corollary}
\newtheorem{claim}[theorem]{Claim}
\newtheorem{remark}[theorem]{Remark}
\newtheorem{lemma}[theorem]{Lemma}

\newcommand{\mD}{\mathcal{D}}
\newcommand{\zo}{\ensuremath{\{0,1\}}}

\newcommand{\X}{\mathcal{X}}
\newcommand{\N}{\mathbb{N}}
\newcommand{\R}{\mathbb{R}}
\DeclareMathOperator*{\expectation}{\mathbb{E}}

\newcommand{\defeq}{\vcentcolon=}

\newcommand{\tth}{^\text{th}}
\newcommand{\DX}{\mathcal{D}_{\X}}
\newcommand{\UX}{\mathcal{U}_{\X}}

\newcommand{\C}{\mathcal{C}}
\newcommand{\set}[1] {\left\{ #1 \right\}}
\newcommand{\stochasticVecs}{\mathcal{Y}}
\newcommand{\eps}{\varepsilon}
\newcommand{\negli}{\mathsf{neg}}
\newcommand{\cupdot}{\mathbin{\mathaccent\cdot\cup}}
\newcommand{\tilX}{\tilde{\X}}
\newcommand{\tilGa}{\tilde{\gamma}}
\newcommand{\noisyGa}{\gamma_\nu}
\newcommand{\decRule}{\rho}
\newcommand{\learner}{\mathcal{L}}
\newcommand{\poly}{\mathsf{poly}}
\DeclareMathOperator*{\argmax}{arg\,max}
\DeclareMathOperator*{\argmin}{arg\,min}
\newcommand{\numOfVecs}{m}

\newcommand{\loss}{\ell}
\newcommand{\lossFamily}{\mathcal{L}}
\newcommand{\expLoss}{\mathsf{exp\text{-}loss}_\mD}
\newcommand{\actionFunctionsClass}{\mathcal{H}}
\newcommand{\actionFunction}{h}

\newcommand{\type}{t}
\newcommand{\setOfTypes}{\mathcal{T}}
\newcommand{\lossEps}{\eps_{\loss}}
\newcommand{\ACEps}{\eps_{\mathsf{AC}}}
\newcommand{\tierone}{\mathsf{tier\text{-}1}}
\newcommand{\tiertwo}{\mathsf{tier\text{-}2}}
\newcommand{\tierthree}{\mathsf{tier\text{-}3}}

\begin{document}

\maketitle

\begin{abstract}
We revisit the foundations of fairness and its interplay with utility and efficiency in supervised learning settings where training labels are richer than binary outcomes, such as risk estimates (probabilities), individual types, or rankings.
We introduce new notions of accurate classification rates for subgroups in the population, defined by comparing the induced positive-classification rates to those under the Bayes-optimal rich predictor.
Our main contributions are computational impossibility results: we show that simultaneously achieving these rate-accuracy guarantees and natural desiderata such as calibration or loss minimization is, in some cases, computationally infeasible, even when training examples are labeled by the Bayes-optimal rich predictor.
Unlike prior impossibility results in this area, these desiderata are simultaneously satisfied by the Bayes-optimal predictor, and each can be achieved efficiently in isolation.
\end{abstract}

\section{Introduction}\label{sec:intro}

Many deployed ML systems follow a predict-then-decide pipeline. A model outputs a prediction about an individual that is richer than a Boolean outcome---for example, a risk estimate (a probability), a distribution over types, or a ranking---and a downstream decision rule maps this prediction to a binary action (approve/deny, admit/reject, intervene/not intervene). In such settings, it is natural to evaluate not only predictive quality, but also the binary decisions induced by the prediction, including their aggregate behavior on subpopulations.

A simple lending example illustrates a gap between predictor-level and decision-level guarantees. Suppose a predictor outputs repayment probabilities and the deployed decision rule approves loans above a fixed cutoff. A predictor can be well-calibrated and yet induce approval rates that differ substantially from those induced by the Bayes-optimal predictor: calibration is an on-average guarantee about predictions, but approval rates depend on how many individuals end up on each side of the cutoff. This motivates asking when guarantees about rich predictions translate into guarantees about the downstream decisions they induce.

Why consider rich predictions at all, rather than learning the binary decision directly? In many applications, the prediction is meant to be reusable across downstream choices: the same learned predictor may be used for multiple downstream tasks, decision rules may change over time (due to constraints, business goals, or regulation), and audits may target both predictive properties (e.g., calibration) and decision-level outcomes (e.g., positive-decision rates). 
This modularity motivates treating the rich prediction itself as the object to be learned: it preserves information that can support future decisions and audits, whereas a binary decision commits to a single downstream use.

\paragraph{Setting.}
Individuals have features $x$ drawn from a distribution $\DX$ over $\X$. We let $\C$ denote a collection of (possibly intersecting) subpopulations, which specifies the groups on which we require guarantees. Throughout, our multigroup notions are required to hold for every $S\in\C$. Nature specifies a Bayes-optimal rich predictor $R^*:\X\to\stochasticVecs$ (where $\stochasticVecs$ may represent probabilities, rankings, or distributions over a finite set of abstract categories, which we call types). When the rich label is a single probability for a binary outcome, $R^*(x)$ is simply $p^*(x)=\Pr[y=1\mid x]$.
The learner receives samples labeled by $R^*(x)$ and outputs a (rich) predictor $\tilde R:\X\to\stochasticVecs$. 

We recall two standard multigroup notions from the algorithmic fairness literature. A predictor $\tilde R$ is \emph{multi-accurate} if, for every group $S\in\C$, its average predicted type distribution is close coordinate-wise to the average type distribution specified by $R^*$. Informally, for every $S\in\C$ and type $t$, the averages of $R^*_t(x)$ and $\tilde R_t(x)$ over $S$ are close. \emph{Multi-calibration} \citep{MC} is stronger: it requires such agreement to hold even after conditioning on the prediction made by $\tilde R$.

A \emph{decision rule} is a randomized map $\decRule:\stochasticVecs\to\zo$ that converts a rich prediction into a binary decision, whereas an \emph{action function} is a randomized map $\actionFunction:\X\to\zo$ that makes a binary decision directly from features (action functions are also commonly called {\em binary classifiers}, and sometimes {\em policies}). Given a loss function $\loss$, we denote the Bayes-optimal loss-minimizing action function by $\actionFunction^*_\loss$. Namely, $h^*_\ell(x)$ denotes the action minimizing expected loss under $R^*(x)$. 

\subsection{Our contributions} \label{sec:intro_contributions}
We introduce two accuracy notions that assess a learned rich predictor via the subpopulation-level positive-decision/classification rates it induces. Along the way, we relate our new notions to standard multigroup guarantees such as multi-accuracy and multi-calibration. Highlights include:

\textbf{Decision-accuracy vs.\ calibration.}
Let $\decRule:\stochasticVecs\to\zo$ be a decision rule that (randomly) maps rich predictions to binary decisions. We formalize \emph{decision-accuracy} by requiring that for each group, the positive-decision rate induced by applying $\decRule$ to a learned predictor $\tilde R$ is close to the rate induced by applying $\decRule$ to the Bayes-optimal rich predictor $R^*$. That is, for every $S\in \C$,
\begin{equation}  \label{eq:MAD}
    \left| \expectation [ \decRule(\tilde{R}(x)) \mid x \in S ] - \expectation [ \decRule(R^*(x)) \mid x \in S ] \right| \le \eps.
\end{equation}
For general decision rules, we prove a computational separation: there are settings where no efficient learner can output a predictor that is both decision-accurate and calibrated, even when training examples are labeled by $R^*(x)$.
We complement this with a tight positive result: for affine decision rules (defined next in \Cref{sec:MAD_vs_Cali}), calibration (and even weaker conditions) implies decision-accuracy, yielding a  characterization of the decision rules that admit decision-accuracy.

\textbf{Classification-accuracy vs.\ loss minimization.} Let $\actionFunction:\X\to\zo$ be an action function that (randomly) maps individuals to binary classifications. Motivated by the above separation for predictors, we define \emph{classification-accuracy} by requiring that, for each group, the positive-classification rate of $\actionFunction$ matches the rate of the Bayes-optimal loss-minimizing action function $\actionFunction^*_\loss$ (which is induced by $R^*$ and a loss function $\loss$). That is, for every $S\in \C$,
\begin{equation} \label{eq:MAC}
    \left| \expectation \left[ \actionFunction(x) \mid x \in S \right] - \expectation \left[ \actionFunction^*_\loss(x) \mid x \in S \right] \right| \le \eps.
\end{equation}
We show a strong negative result: for non-trivial loss functions, it can be computationally infeasible to learn an action function that is classification-accurate and also has loss competitive with the minimal benchmark class of constant functions. In other words, to obtain classification accuracy one may incur expected loss strictly larger than  the constant functions (classify everyone as $0$ or everyone as $1$). As above, this infeasibility result holds even when training examples are labeled by $R^*(x)$.

\textbf{Positive results: one desideratum at a time.}
Although the above pairs of guarantees can be incompatible, we give efficient algorithms achieving each desideratum on its own.
We show how multi-accuracy can be used to obtain decision-accuracy and classification-accuracy, and leverage strong results from prior works on multi-calibration and omniprediction \citep{omni} to show how to learn action functions with competitive loss.

The notions and their associated objects are summarized in \Cref{tab:cheat-sheet}. 
The key message is that the desiderata we consider are Bayes-consistent and each is efficiently achievable on its own, yet certain combinations are computationally incompatible. In particular, our negative results apply even in settings where the Bayes-optimal predictor (and the Bayes-optimal action function) have succinct descriptions; a super-polynomial-time learner can recover $R^*$ from a polynomial-size training set and satisfy all desiderata, but no efficient learner can.

\begin{table}
\caption{New accuracy notions: Multi-Accuracy-on-Decision (MAD, \Cref{eq:MAD}) and  Multi-Accuracy-on-Classification (MAC, \Cref{eq:MAC}).}
  \label{tab:cheat-sheet}
  \centering
  \begin{tabular}{p{0.32\linewidth}p{0.60\linewidth}}
    \toprule
    Objects     & Accuracy notion     \\
    \midrule
    predictor 
$\tilde R:\X\to\stochasticVecs$, 
decision rule   $\decRule:\stochasticVecs\to\zo$ & {\textbf{MAD:} On every group, the positive-decision rate induced by applying  $\decRule$ to $\tilde R$ matches that induced by applying $\decRule$ to $R^*$.}      \\
    action function $\actionFunction:\X\to\zo$     & \textbf{MAC:} On every group, the positive-classification rate of $\actionFunction$ matches that of the loss-minimizing action function.       \\
    \bottomrule
  \end{tabular}
\end{table}

\paragraph{Implications for fairness.}
Although our desiderata are Bayes-consistent accuracy guarantees, they are naturally motivated by fairness concerns: subpopulations can be harmed by both false positives and false negatives, and these harms are reflected in both decision rates and loss-based objectives \citep{LDRSH18}. Our positive results show that rich labels enable strong subgroup-level protections such as calibration, loss minimization, and our new rate-accuracy notions, each achievable efficiently on its own. Our negative results show that, even with rich labels, certain pairs of these protections can be computationally incompatible, forcing a policy maker to choose which protections to provide.

\section{Related work}\label{sec:related_work}

\paragraph{Prior incompatibility results.}
Classic impossibility results in algorithmic fairness, most notably those of \citet{kleinberg17} and \citet{Cho17}, identify inherent statistical tensions between calibration and error-rate parity constraints across demographic groups; these tensions persist even given complete knowledge of the Bayes-optimal conditional probabilities. The importance of these separations lies in exposing conceptual contradictions among the notions. Recent work has also studied fairness--accuracy and fairness--fairness tradeoffs through Pareto-frontier formulations~\citep{chairethinking}. In contrast, our separations are computational in nature, and illustrate incompatibilities between different notions that are all grounded in, and in line with, accuracy.

\paragraph{Multi-accuracy and multi-calibration for predictions.}
The literature studies several notions of calibration in the multi-type (or multi-class) setting (see \citet{EfficientMultiClassMC} for a full discussion). We highlight two extremes among these: \emph{Canonical} (or \emph{distribution) calibration} \citep{KF15,WLZ19} conditions on every possible predicted type distribution, and yields a robust accuracy guarantee, but the relevant event space grows exponentially with the number of types, leading to exponential sample complexity even for a single group. In contrast, \emph{class-wise} (or \emph{coordinate-wise) calibration} \citep{KPNK} conditions separately on each coordinate of the predicted distribution (a polynomial-size family of events) and can be obtained efficiently. In \Cref{sec:MA_defs,sec:MC} we include formal definitions of the variants we study and the relevant learning algorithms for achieving them.

\paragraph{Other variants of multi-class calibration.}
Canonical calibration requires predicted label probabilities to match the true label distribution among instances with the same prediction. As discussed in the previous paragraph, this is often infeasible in multi-class settings due to exponential sample complexity in the number of labels. To address this, \citet{decisionCali} propose \emph{decision calibration}, which ensures that predicted and true distributions are indistinguishable to decision-makers with a bounded number of actions. Their recalibration algorithm runs with sample complexity polynomial in both the number of actions and labels. \citet{EfficientMultiClassMC} strengthen this by introducing \emph{projected smooth calibration}, which guarantees robustness for all binary classification tasks of the form ``is the label in a subset $T \subseteq \setOfTypes$?'' and achieves it efficiently in both time and samples. Our work also supports downstream binary tasks, whereas our emphasis is not on efficient recalibration per se, but rather on characterizing which decision or classification guarantees can be achieved efficiently.

\paragraph{Uncertainty decomposition via higher-order calibration.} 
The recent work of \citet{Charlotte} introduces higher-order calibration for predictors that output distributions over label distributions. They aim to decompose uncertainty into \emph{epistemic} and \emph{aleatoric} components via higher-order calibration, while relying on access to multiple independent labels per instance (as Nature itself is probabilistic). This enables a rigorous separation between model-induced uncertainty and inherent randomness. 
With the tension between the two kinds of uncertainty in mind, one can understand our work as embedding uncertainty into the notion of MAC, which penalizes \emph{epistemic} uncertainty---when the model fails to separate different types in the data---while accepting \emph{aleatoric} uncertainty as intrinsic.\footnote{For instance, if all applicants in a subgroup have Bayes-optimal rich label assigning probability $1/2$ to being $\tierone$ and probability $1/2$ to being $\tierthree$, then MAC may tolerate rejecting them under a threshold-based decision rule. But if the subgroup instead consists of a mixture of applicants whose Bayes-optimal rich labels are pure $\tierone$ and pure $\tierthree$, and the model fails to distinguish them, then MAC flags this as a failure of learning.}
This contrasts with fairness notions like balance or equalized odds, which penalize even purely aleatoric deviations. In our framework, aleatoric uncertainty defines the baseline---we do not ask the model to outperform the inherent randomness of Nature, only to capture structure when it exists. In this way, our approach captures the epistemic/aleatoric distinction implicitly, by requiring consistency with the Bayes-optimal predictor, and without relying on multiple labels per instance.

\paragraph{Evidence-based rankings.}
\citet{ranking} study the problem of learning rankings from binary outcomes alone, without access to individual probabilities. Their framework defines a ranking as a real-valued function over the domain $\X$, consistent with our formulation up to the fact that their range is continuous (i.e., interval $[0,1]$), whereas we focus on coarse, discrete rankings (i.e., a rank is a number in $\{1,\dots,k\}$). Additionally, our definitions are extended to allow probabilistic outputs, i.e., distributions over ranks. In their setup, the quality of a learned ranking is evaluated indirectly: rather than comparing it to a ground-truth ranking, as in our case, they assess the ranking through the best predictor it can induce.
They propose \emph{evidence-consistency} and \emph{domination-compatibility}, where the former is stronger than the latter. Since their assumptions are more restrictive (due to limited supervision), the corresponding guarantees are also weaker: the notion we consider of coordinate-wise-multi-accuracy (which assumes access to individual ranks) implies evidence-consistency.

\paragraph{Relation to linearizing statistics.} 
Multi-accuracy can be read as keeping the group averages of $\tilde{R}$ correct: for each group $S \in \C$ and type $\type \in \setOfTypes$, the average of $\tilde{R}_\type(x)$ over $S$ matches that of $R^*_\type(x)$. 
For affine $\decRule$, the groupwise decision rates depend only on these averages, so multi-accuracy implies decision-accuracy (\Cref{thm:affine-MA-MD}). This mirrors the recipe in  \citet{OmniForRegression}: predict the right statistics and keep them calibrated on groups to obtain near-optimal actions.

\section{Overview of results and techniques}\label{sec:intro:tech-overview}
We provide a more detailed overview of our results and their proofs, focusing on the negative results.

\begin{figure*}
\centering
\begin{tikzpicture}[
  node distance=1.0cm and 2.5cm,
  >={Stealth}
]

\node[draw, rounded corners] (MC)                {MC};
\node[draw, rounded corners, right=of MC] (MA)  {MA};
\node[draw, rounded corners, right=of MA] (MAD) {MAD};

\node[draw, rounded corners, below=of MC] (Loss) {Loss Minimization};
\node[draw, rounded corners, below=of MAD] (MAC) {MAC};

\node[anchor=east, font=\bfseries] at ([xshift=-2.5cm]MC.west) {Predictors:};
\node[anchor=east, font=\bfseries] at ([xshift=-0.4cm]Loss.west) {Action Functions:};

\draw[->, semithick, green!50!black] (MC) -- node[below] {\Cref{clm:MC_implies_MA}} (MA);
\draw[->, semithick, green!50!black] (MA) -- node[below] {\Cref{thm:affine-MA-MD}}
node[midway, text=black] {$(*)$}(MAD);
\draw[dashed, ->, semithick, green!50!black] (MC) -- node[left] {\Cref{thm:MC_for_loss_mini}} (Loss);
\draw[dashed, ->, semithick, green!50!black] (MAD) -- node[right] {\Cref{cor:MA_implies_MAC}} (MAC);

\draw[<->, red, semithick, bend left=25]
  (MC) to
  node[above, text=red] {Computational Hardness (\Cref{thm:non-affine-negative})}
  node[midway, text=black] {$(**)$}
  (MAD);
\draw[<->, red, semithick, bend left=25]
  (MAC) to node[below, text=red] {Computational Hardness (\Cref{thm:MAC-lossMin-negative})} (Loss);

\end{tikzpicture}
\caption{Positive and negative results.  MC: Multi-Calibration, MA: Multi-Accuracy, MAD: Multi-Accuracy-on-Decision, MAC: Multi-Accuracy-on-Classification. The top row contains properties defined over a predictor and the bottom row contains properties defined over the action function obtained by applying a decision rule. A green arrow from $X$ to $Y$ means that satisfying $X$ implies satisfying $Y$. A green dashed arrow from $X$ to $Y$ means that applying a decision rule to a predictor satisfying $X$ yields an action function satisfying $Y$. A red arrow between $X$ and $Y$ means that achieving both simultaneously is computationally hard. The implication marked $(*)$ holds for affine decision rules, and the impossibility marked $(**)$ holds for rules that are far from affine.
The dashed implication from MAD to MAC additionally assumes that Nature is deterministic.
}
\label{fig:results}
\end{figure*}
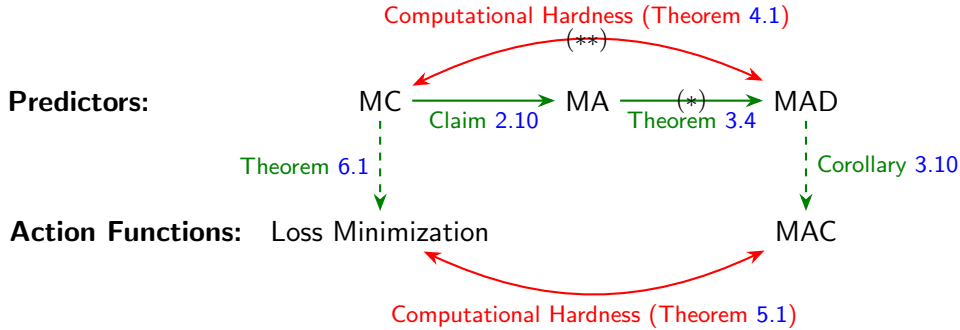

\subsection{Decision-accuracy vs.\ calibration}\label{sec:MAD_vs_Cali}

\paragraph{Multi-accurate decisions.}
Recall that in the multi-type setting, a predictor's output is a distribution over a finite type space $\setOfTypes$, and $\stochasticVecs$ denotes the set of such distributions. A \emph{decision rule} $\decRule:\stochasticVecs\to\zo$ maps a prediction (a distribution over $\setOfTypes$) to a binary decision. In deployments, $\decRule$ is typically designed with the Bayes-optimal rich predictor $R^*$ in mind, but is applied to a learned predictor $\tilde R$.

Recall the formalization of multi-accuracy-on-decision (MAD) from \Cref{eq:MAD}: 
MAD compares the decision rate induced by $\tilde R$ to the decision rate induced by the Bayes-optimal rich predictor $R^*$. Thus, to define or audit MAD, one needs access to the rich labels $R^*(x)$, not merely to binary outcomes. For example, suppose a loan rule approves applicants whose repayment probability is at least $0.95$, and in a group $S$ half of the observed loans are repaid. This observation is consistent both with the case where every applicant has repayment probability $0.5$, in which case the Bayes-optimal rule rejects everyone, and with the case where half the applicants have repayment probability $1$ and half have probability $0$, in which case the Bayes-optimal rule approves half the group. These two cases induce different MAD benchmarks, even though they may be indistinguishable from binary outcomes alone. This motivates our focus on rich-label settings, where training examples reveal each individual's rich label $R^*(x)$.

Decision-accuracy (with respect to some decision rule) and calibrated predictions are both desirable properties for a predictor, and enforcing them for subgroups can address certain fairness concerns. Can we simultaneously achieve both of these natural notions? We give a tight characterization of the decision rules for which this is and is not possible: close-to-affine rules allow coexistence, while far-from-affine rules lead to computational hardness.

\paragraph{Affine decision rules: multi-accuracy implies decision-accuracy.} We say that a decision rule is {\em affine} if, on input a probability distribution over  types, it operates by sampling a type from the distribution and applying a fixed function that (randomly) maps each type to a decision in $\{0,1\}$.\footnote{This formulation is equivalent to the standard definition of affineness: applying the rule to a convex combination of two distributions over types yields the same convex combination of the outcomes obtained by applying the rule to each distribution separately. We refer the reader to \Cref{def:close_to_affine} for the full formal definition of being affine and close to affine.}

\emph{Example.} Suppose a predictor $\tilde{R}$ outputs, for each applicant, a distribution over three future student types: $\tierone$, $\tiertwo$, and $\tierthree$. Consider an affine decision rule that samples a type from this distribution and then admits with probability $1$ for $\tierone$, $0.5$ for $\tiertwo$, and $0$ for $\tierthree$; equivalently, the admission probability is the predicted mass on $\tierone$ plus half the predicted mass on $\tiertwo$. Suppose that on some subgroup $S$, the Bayes-optimal predictor $R^*$ has average type distribution $(0.4,0.2,0.4)$ (for example, it assigns $(0.5,0.2,0.3)$ for half of the applicants and $(0.3,0.2,0.5)$ for the other half). The admission rate induced by $R^*$ on $S$ is $0.4 + 0.5 \cdot 0.2 = 0.5$.
But any predictor $\tilde{R}$ that is multi-accurate has essentially the same average type distribution on $S$, namely $(0.4,0.2,0.4)$, as multi-accuracy requires. Therefore, under the same affine rule, $\tilde{R}$ induces essentially the same admission rate on $S$, namely about $50\%$.

We show that this intuition holds generally: for affine decision rules, multi-accurate predictors (and so do multi-calibrated predictors) induce multi-accurate decisions (MAD).

\begin{theorem}[Informal, see  \Cref{thm:MA_implies_MAD}]
\label{thm:affine-MA-MD}
Let $\decRule : \stochasticVecs \to \zo$ be an affine decision rule. If a predictor $\tilde{R}$ is multi-accurate w.r.t.\ a collection of groups $\C$, then it is also decision-accurate for every group in $\C$.
\end{theorem}

We note that the error in the decision-accuracy guarantee depends on the error parameter in the multi-accuracy guarantee. Moreover, it suffices that the decision rule be \emph{close} to affine; the resulting closeness parameter also contributes to the overall decision-accuracy error. This means that our characterization is \emph{tight}: \Cref{thm:affine-MA-MD} shows that rules that are close to affine still allow coexistence, whereas \Cref{thm:non-affine-negative} presented next will show hardness for rules that are far from affine. 

\paragraph{Non-affine rules: calibration and decision-accuracy are in conflict.}
In many settings, affine decision rules are too restrictive.  
For example, a lender may wish to approve only applicants who are nearly certain to repay the loan under the predicted rich label, rather than applicants who have, say, only a $90\%$ chance of repayment; this kind of ``near-certainty'' requirement is inherently non-affine in the predicted distribution. Can we obtain predictors that are both calibrated and decision-accurate for such non-affine decision rules?

We prove a general negative result: for decision rules that are far from affine (informally, they do not approximately preserve convex combinations), there are data distributions for which it is computationally infeasible to learn a predictor that is simultaneously calibrated and decision-accurate. 
Importantly, the conflict is neither conceptual nor a consequence of simultaneously enforcing many group constraints: the Bayes-optimal rich predictor $R^*$ is both calibrated and decision-accurate, but cannot be learned efficiently (nor can any other predictor satisfying both guarantees). At the same time, each guarantee is natural and efficiently achievable in isolation; the hardness only arises when we require both at once.

\emph{Example.} Let us consider our running example of lending. Suppose half of the applicants belong to a subpopulation $\X_1$ with true repayment probability $0.99$, and the other half belong to a subpopulation $\X_2$ with true repayment probability $0.90$, but the two are computationally hidden from the learner's features. Then an efficient calibrated predictor $\tilde R$ may be forced to average over the hidden partition and assign all applicants repayment probability $0.945$. If the bank approves only applicants with predicted repayment probability above the threshold $0.95$ (this rule is far from affine), then the Bayes-optimal predictor $R^*$ approves all of $\X_1$ and none of $\X_2$ (overall approval rate $50\%$), while $\tilde{R}$ rejects everyone (overall approval rate $0\%$).\footnote{Strictly speaking, this threshold rule is not Lipschitz, as required by \Cref{thm:non-affine-negative}.
However, this is easily resolved by smoothing the rule: we can replace it with a randomized Lipschitz rule whose approval probability agrees with the threshold rule outside the interval $[0.947,0.953]$ and works linearly within this interval.
This leaves the approval rates in the example unchanged.}

The calibration notion we consider is coordinate-wise, a very relaxed requirement; proving impossibility even under this weak notion strengthens the result. Moreover, recall that we work in the rich-label model: the learner has sample access to pairs $(x,R^*(x))$, where $R^*(x)\in\stochasticVecs$ is a distribution over types. This further strengthens the lower bound by showing that the obstruction is not insufficient supervision from binary outcomes, but computational hardness.

\begin{theorem}[Informal, see \Cref{thm:impossibility_result}]
\label{thm:non-affine-negative}
Assume the existence of a computationally hidden subpopulation. Let $\decRule : \stochasticVecs \to \zo$ be a decision rule that is Lipschitz and far from affine. There exists a data distribution for which no polynomial-time learner can output a predictor that is both coordinate-wise calibrated and decision-accurate w.r.t.\ $\decRule$ (except with negligible probability).
\end{theorem}

The assumption says that there is a subpopulation of the required constant density whose membership is not inferable from the learner's observed features. We formalize it via computational indistinguishability: no efficient procedure can reliably tell whether an  individual belongs to the subpopulation or to its complement.
We give two formal realizations. The first is information-theoretic, modeling latent attributes, missing covariates, or privacy-constrained access to features: labels may be efficiently generated from the full feature set, while the learner's observed features make the relevant partition inaccessible. The second is cryptographic, based on one-way functions, a lightweight and well-established assumption. See \Cref{sec:indistin_subsets} for the formal definitions and proofs.

\begin{proof}[Proof overview for \Cref{thm:non-affine-negative}.] 
Given a non-affine decision rule $\decRule$, we construct a distribution for which no efficient learner can output a predictor that is both calibrated and decision-accurate, even for a single subpopulation. The idea behind the proof is to construct a setting where any efficiently learned calibrated predictor must smooth out distinctions between inputs, even when their true predictions are different. To do so, we consider the following setup. Since $\decRule$ violates affineness, there exist two vectors (representing distributions over labels) $y$ and $y'$ such that \begin{equation}\label{eq:rule_not_affine} \frac{1}{2} \decRule(y) + \frac{1}{2} \decRule(y') \neq \decRule\left( \frac{1}{2} y + \frac{1}{2} y' \right).\end{equation} For simplicity, we use the constant $1/2$ in this argument, though the same reasoning applies to any convex combination. We construct the domain as $\X = \X_1 \cupdot \X_2$, where each subset constitutes half of $\X$ and the two are computationally indistinguishable, relying on the hidden-subpopulation assumption. The Bayes-optimal predictor $R^*$ assigns the distribution $y$ on $\X_1$ and $y'$ on $\X_2$. Informally, any efficiently learned calibrated predictor $\tilde R$ cannot reliably distinguish $\X_1$ from $\X_2$; calibration then forces it to behave like the ``squashed'' predictor that outputs the mixture $\frac{1}{2}y+\frac{1}{2}y'$ rather than the group-dependent labels $y$ and $y'$, on any $x \in \X$. 

To evaluate decision-accuracy, we compare the expected value of $\decRule$ applied to $\tilde{R}$ and to $R^*$. For $R^*$, this expectation equals $\frac{1}{2} \decRule(y) + \frac{1}{2} \decRule(y')$, whereas for $\tilde{R}$ it equals $\decRule\left( \frac{1}{2} y + \frac{1}{2} y' \right)$. By \Cref{eq:rule_not_affine}, these values differ, and thus $\tilde{R}$ is not decision-accurate with respect to~$\decRule$.

The formal proof of the general case is the most technically involved part of this work, as it introduces several technical challenges. First, $\tilde{R}$ is only \emph{approximately} calibrated, and may behave arbitrarily on small subsets. This interacts subtly with the weak calibration requirement we assume, under which the level sets\footnote{This term is from \citet{MC}. In our relaxed \emph{coordinate-wise} multi-calibration setting, a ``level set'' groups individuals whose predicted probability for a fixed type is approximately the same. These sets can overlap across different types, since each individual is assigned a full distribution over types.} of $\tilde{R}$ do not form a partition, complicating the identification of a large subset where $\tilde{R}$ fails to be decision-accurate. Second, since we aim to refute {\em approximate} decision-accuracy, we must ensure that $\decRule$ is bounded away from any affine function. To control this, we impose a Lipschitz condition on $\decRule$, and  argue that without such regularity, the resulting lower bound would be vacuous. Finally, since $\tilde{R}$ must be discretized to be learned efficiently, care is required to ensure the argument is robust to any choice of discretization.
\end{proof}

Non-affine rules arise naturally from loss minimization. Given a loss function, the loss-minimizing decision rule takes the positive action (e.g., approve/admit/intervene) precisely when its expected loss is smaller than that of the negative action (e.g., deny/reject/not intervene). In a medical setting, for example, a treatment with severe side effects may be prescribed only when the predicted distribution places overwhelming mass on types corresponding to very high medical risk. The loss function trades off the harm of prescribing an unnecessary or dangerous treatment against the harm of failing to treat a patient who is in fact at high risk, and the induced rule prescribes treatment exactly when the expected benefit outweighs the expected side-effect cost. Such thresholding of expected loss differences is inherently far from affine. 

In fact, loss-minimizing decision rules are far from affine for every non-trivial\footnote{\label{footnote1}We only require that the loss is not degenerate: there exists a type for which one action is significantly better than the other, but no single action is optimal for all types (otherwise the problem is trivial, since a constant action minimizes loss).} loss function (see \Cref{lem:loss_min_dec_rule_not_affine,rem:loss_min_dec_rule_not_affine_as_cor}).
Combined with the negative result, this implies that learning a calibrated predictor that is decision-accurate with respect to the loss-minimizing rule may be computationally infeasible.\footnote{Technically, the loss-minimizing decision rule is not Lipschitz, so the impossibility result cannot apply directly. However, in \Cref{clm:loss_min_dec_rule_impossibility}, we show that it extends for these rules, for any non-trivial loss function.} 
This leads to the next question: if applying a loss-minimizing rule to a learned predictor may be problematic, can we instead learn an action function directly while preserving classification-accuracy?

\subsection{Classification-accuracy vs.\ loss minimization}

\paragraph{Multi-accurate classifications.}
Binary decisions need not be produced by applying a decision rule to a learned predictor. Accordingly, we study general \emph{action functions} $\actionFunction:\X\to\zo$, which map features directly to a binary action. We work in a loss-minimization framework: a loss function $\loss$ assigns a loss $\loss(t,a)$ to taking action $a\in\zo$ when the individual's type is $t\in\setOfTypes$. 

Given an individual $x$ and an action $a$, the relevant quantity to consider is the expected loss under the individual's rich label $R^*(x)$, namely $\expectation_{t\sim R^*(x)}[\loss(t,a)]$. If we had access to $R^*$, we could choose, for each individual $x$ separately, the action $a\in\zo$ that minimizes $\expectation_{t\sim R^*(x)}[\loss(t,a)]$, yielding the Bayes-optimal loss-minimizing action function $h^*_{\loss}$. For a subpopulation $S$, this benchmark induces a positive-classification rate on $S$. 
Recalling \Cref{eq:MAC}, we say that an action function is \emph{multi-accurate-on-classification} (MAC) for a collection of groups $\C$ if it matches these groupwise positive-classification rates for every $S\in\C$. Note that $h^*_{\loss}$ both minimizes loss and (trivially) satisfies MAC for any collection $\C$; thus the negative result of \Cref{thm:MAC-lossMin-negative} presented next is purely computational, as was the previous negative result.

\paragraph{Loss minimization and classification-accuracy are in conflict.}
A second natural objective when learning action functions is to (approximately) minimize expected loss relative to a benchmark class. In contrast to our multigroup rate notions, loss minimization is a global objective. Our second negative result shows that, for essentially any non-trivial loss function (see \Cref{footnote1}), loss minimization can be incompatible with classification-accuracy. Indeed, the incompatibility already appears when the benchmark class for loss minimization consists only of the two constant action functions.

\emph{Example.} Let us consider our running example of preventive medicine. Suppose $75\%$ of patients belong to a  subpopulation $\X_1$ for whom treatment is harmful, with loss $1$ for ``treat'' and loss $0$ for ``do not treat'', while the remaining $25\%$ belong to a  subpopulation $\X_2$ for whom treatment is beneficial, with loss $0$ for ``treat'' and loss $0.2$ for ``do not treat''. 
The Bayes-optimal action function treats exactly $\X_2$, so the treatment rate is $25\%$. But if $\X_1$ and $\X_2$ are computationally hidden from the learner's features, then any efficiently learnable action function that also treats about $25\%$ of patients cannot reliably pick out $\X_2$, and instead treats an essentially random $25\%$ of the population. Its expected loss is then
\[
0.75 \cdot 0.25 \cdot 1 + 0.25 \cdot 0.75 \cdot 0.2 = 0.225,
\]
which is worse than the constant action function ``treat nobody'', whose loss is only $0.25 \cdot 0.2 = 0.05$.

The following theorem formalizes this phenomenon: for every non-trivial loss function, classification-accuracy can force any efficient action function to have loss worse than a constant benchmark.

\begin{theorem}[Informal, see \Cref{thm:impossibility_result_of_loss}]
\label{thm:MAC-lossMin-negative}
Assume the existence of a computationally hidden subpopulation. For any non-trivial loss function $\loss$ (as above), there exists a data distribution such that any polynomial-time action function that satisfies classification-accuracy must incur expected loss strictly larger than one of the constant action functions (classify everyone as $0$ or everyone as $1$).
\end{theorem}

\begin{proof}[Proof overview for \Cref{thm:MAC-lossMin-negative}]

For any non-trivial loss function, we construct a Nature under which any efficiently computable action function that is classification-accurate fails to minimize loss. We prove this with respect to the most restricted hypothesis class, which contains only the two constant functions that either accept or reject all individuals.
The proof idea is as follows. The non-triviality of the loss implies the existence of two labels, or types, $\type$ and $\type'$, such that {\em rejecting} type $\type$ results in a significantly lower loss compared to accepting it, with a difference of at least $\alpha > 0$, while {\em accepting} type $\type'$ is preferable in terms of loss, though the difference may be smaller. Here, ``reject'' and ``accept'' correspond to the binary actions $0$ and $1$, but any other binary action can work. 

We define the domain to contain an indistinguishable subset $\X_1 \subseteq \X$ of size $3/4$ of $\X$, relying on the hidden-subpopulation assumption. Let $R^*$ assign type $\type$ to each $x \in \X_1$ and type $\type'$ to the remaining individuals. Consequently, the loss-optimal action function rejects everyone in $\X_1$ and accepts the rest, resulting in an expected acceptance rate of $1/4$ over $\X$. Thus, any action function $\actionFunction$ that is classification-accurate must accept roughly $1/4$ of the domain. 

By indistinguishability, the set of accepted individuals must intersect $\X_1$ in approximately a $3/4$ fraction: if the efficiently computable action function $\actionFunction$ accepted a noticeably different fraction from $\X_1$ than from its complement, then $\actionFunction$ itself would distinguish $\X_1$ from $\X\setminus\X_1$. This implies that the expected loss of $\actionFunction$ conditioned on the set $\{\actionFunction(x) = 1\}$ matches the expected loss of accepting the entire domain. Let $L_1$ denote this quantity and $L_0$ denote the analogous quantity for rejecting everyone. Then, the expected loss of $\actionFunction$ is given by $3L_0/4  + L_1/4 $, and the difference between this loss and that of the function that rejects everyone is $(L_1 - L_0)/4$. On the other hand, by the non-triviality of the loss, $L_1 - L_0$ is lower-bounded by $\alpha / 2$, allowing us to lower-bound the loss-minimization error of $\actionFunction$ by at least $\alpha/8$. Hence no polynomial-time action function can satisfy classification-accuracy and simultaneously match (or beat) the best constant action.
\end{proof}

\subsection{Positive results: one desideratum at a time}
While our negative results show that certain pairs of natural desiderata can be computationally incompatible, we also give positive results showing that each desideratum is efficiently achievable on its own. Our positive results build on predictors satisfying multi-accuracy or multi-calibration. We summarize these results here, emphasizing the case of loss minimization, which is the most subtle.

\paragraph{Decision and classification accuracy.} First, as already proved in \Cref{thm:affine-MA-MD}, MAD can be achieved by multi-accuracy. Moreover, in the natural special case when Nature's rich predictor $R^*$ is deterministic, multi-accuracy suffices to obtain both MAD and MAC. Concretely, randomly instantiating a multi-accurate predictor (sampling a type for each individual from its predicted distribution) yields MAD, 
and applying the loss-minimizing decision rule yields MAC. 
After random instantiation, the rule maps the label to $\zo$. MAD is achieved with any such mapping, while MAC is achieved when this mapping is the loss-minimizing action for the realized label (i.e., it chooses the loss-minimizing action given the sampled label). Unlike direct loss minimization, in the MAC path the loss-minimizing rule is applied only {\em after} random instantiation.  \Cref{fig:loss_vs_MAC} visualizes  the two paths.

\paragraph{Loss minimization.}
Finally, we study learning an action function whose expected loss is competitive with a benchmark class. Building on the theory of omnipredictors~\citep{omni}, we prove that applying a loss-minimizing decision rule to a predictor that is multi-calibrated for a collection of sets $\C$ yields an action function competitive with a benchmark class closely related to $\C$.

\begin{theorem}[Informal, see \Cref{thm:MC_for_loss_mini}]
\label{theorem:MC-loss-minimization}
Let $\actionFunctionsClass$ be a hypothesis class over $\X$, and let $\C$ be a collection of subsets induced by $\actionFunctionsClass$, i.e., $\{S_\actionFunction: \actionFunction \in \actionFunctionsClass\} \subseteq \C$ where $S_\actionFunction = \{ x \in \X : \actionFunction(x)=1\}$.
For every loss function $\loss$, if $\tilde{R}$ satisfies canonical $\C$-multi-calibration, then applying the loss-minimizing decision rule to $\tilde{R}$ yields an action function whose expected loss is competitive with every $\actionFunction \in \actionFunctionsClass$.
\end{theorem}

A full comparison to the setting of \citet{omni} can be found in \Cref{sec:MC_for_loss}; in what follows, we focus on the results that go beyond prior work. Since the calibration guarantee used in \Cref{theorem:MC-loss-minimization} is strong and (as discussed above) may be expensive to obtain, we also study relaxed settings in which weaker calibration guarantees, and thus more efficient learning, suffice for loss minimization. In \Cref{sec:relaxing_the_MC_for_probs}, we focus on type spaces corresponding to success probabilities (as in the loan repayment example) and losses that are linear in the predicted probability; in this case, it suffices to multi-calibrate the predicted expectations, reducing the problem to one-dimensional prediction. In \Cref{sec:relaxing_the_MC_for_one_loss}, we allow general type spaces but fix a single loss function. Finally, \Cref{sec:relaxing_the_MC_for_many_losses} gives a reduction from loss minimization to auditing multi-calibration; while this can still be computationally demanding, it avoids the exponential sample complexity inherent in canonical multi-calibration.

\subsection{Limitations and implications}

\paragraph{Limitations.}
First, our results are theoretical: the paper proves that the incompatibilities we identify can occur, but does not show that they arise in typical deployed systems or measure their frequency empirically. 
Second, the lower bounds rely on the existence of a hidden subpopulation, whose membership is inaccessible to efficient learners. While we give both information-theoretic and cryptographic realizations of this assumption, the results should be interpreted as identifying a computational obstruction rather than as claiming that the obstruction appears in every deployment. Moreover, our positive results  inherit the computational and sample-complexity requirements of multi-accuracy and multi-calibration.

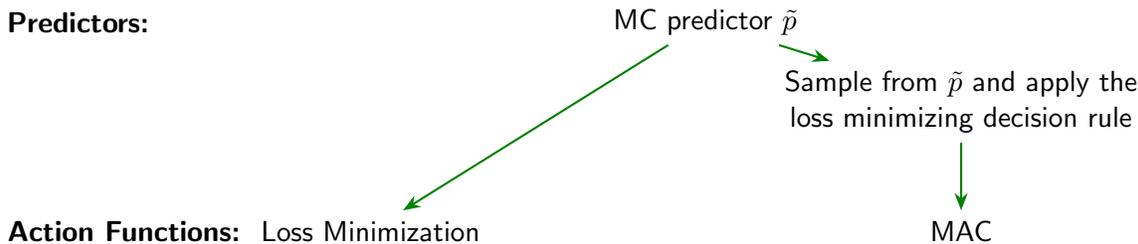
\begin{figure}
\centering
\begin{tikzpicture}[
  x=1cm, y=1cm,
  >={Stealth}
]

\node[anchor=west, font=\bfseries] (PredLbl) at (0,0) {Predictors:};
\node[anchor=west, font=\bfseries] (ActLbl)  at (0,-2.5) {Action Functions:};

\node[anchor=west] (MCtildep) at (6,0) {MC predictor $\tilde{R}$};

\node[anchor=west] (Loss) at (3.3,-2.5) {Loss Minimization};
\node[anchor=west] (MAC)  at (11,-2.5) {MAC};

\node[anchor=west, align=center, text width=4.5cm] (desc2) at (7,-1.3)
  {Sample from $\tilde{R}$ and apply the loss minimizing decision rule};

\draw[->, semithick, green!50!black] (MCtildep) -- (Loss);
\draw[->, semithick, green!50!black] (MCtildep) -- (desc2);
\draw[->, semithick, green!50!black] (desc2) -- (MAC);

\end{tikzpicture}
\caption{Choice in how to use a predictor. While a single predictor may enable both loss minimization and MAC, these goals cannot always be achieved simultaneously. This leads to two distinct paths: applying the loss-minimizing decision rule to the full distribution $\tilde{R}$ to directly minimize loss (left), or first sampling from $\tilde{R}$ and then applying the decision rule to achieve MAC (right). The MAC guarantee for the right-hand path assumes that Nature is deterministic.
}
\label{fig:loss_vs_MAC}
\end{figure}

\paragraph{Implications.}
\Cref{fig:loss_vs_MAC}, which builds on \Cref{fig:results}, highlights the central choice exposed by our results. Given a multi-calibrated predictor $\tilde{R}$, one can either apply the loss-minimizing decision rule (a non-affine rule) to directly obtain a loss-minimizing action function, or use a decision rule that first randomly instantiates $\tilde{R}$ (an affine rule) to achieve MAD or MAC. 
Thus, calibration plays a central and somewhat counter-intuitive role. On the one hand, multi-calibrated predictors provide a versatile starting point from which we can efficiently obtain strong guarantees---including decision-accuracy, classification-accuracy, and loss competitiveness---depending on how we post-process the predictions. On the other hand, our separations show that these guarantees can be computationally incompatible when demanded simultaneously. 

The resolution is that calibration is not itself the end goal; different downstream goals require using the predictor in different ways at decision time. Since both approaches operate on the same predictor $\tilde{R}$, the choice can be deferred to deployment: one can train a multi-calibrated predictor and later decide whether to prioritize MAC or loss competitiveness.

\newpage
\begin{ack}
We thank Shafi Goldwasser for her collaboration in early stages of this project, which motivated our work. We also thank Parikshit Gopalan for many useful discussions, and Charlotte Gates Peale for helpful feedback on an earlier draft. Omer Reingold is supported by the Simons Foundation Collaboration on the Theory of Algorithmic Fairness, the Sloan Foundation Grant 2020-13941 and the Simons Foundation investigators award 689988. The authors declare no competing interests.

\end{ack}

\bibliography{refs}

@inproceedings{TrevisanTV09,
  author       = {Luca Trevisan and
                  Madhur Tulsiani and
                  Salil P. Vadhan},
  title        = {Regularity, Boosting, and Efficiently Simulating Every High-Entropy
                  Distribution},
  booktitle    = {Proceedings of the 24th Annual {IEEE} Conference on Computational
                  Complexity},
  pages        = {126--136},
  publisher    = {{IEEE} Computer Society},
  year         = {2009},

}

@inproceedings{BeyondBer,
  author       = {Cynthia Dwork and
                  Michael P. Kim and
                  Omer Reingold and
                  Guy N. Rothblum and
                  Gal Yona},
  editor       = {Sanjoy Dasgupta and
                  Nika Haghtalab},
  title        = {Beyond Bernoulli: Generating Random Outcomes that cannot be Distinguished
                  from Nature},
  booktitle    = {International Conference on Algorithmic Learning Theory, 29 March
                  - 1 April 2022, Paris, France},
  series       = {Proceedings of Machine Learning Research},
  volume       = {167},
  pages        = {342--380},
  publisher    = {{PMLR}},
  year         = {2022},
  url          = {https://proceedings.mlr.press/v167/dwork22a.html},
  bibsource    = {dblp computer science bibliography, https://dblp.org}
}

@inproceedings{MC,
  author       = {{\'{U}}rsula H{\'{e}}bert{-}Johnson and
                  Michael P. Kim and
                  Omer Reingold and
                  Guy N. Rothblum},
  editor       = {Jennifer G. Dy and
                  Andreas Krause},
  title        = {Multicalibration: Calibration for the (Computationally-Identifiable)
                  Masses},
  booktitle    = {Proceedings of the 35th International Conference on Machine Learning,
                  {ICML} 2018, Stockholmsm{\"{a}}ssan, Stockholm, Sweden, July
                  10-15, 2018},
  series       = {Proceedings of Machine Learning Research},
  volume       = {80},
  pages        = {1944--1953},
  publisher    = {{PMLR}},
  year         = {2018},
  url          = {http://proceedings.mlr.press/v80/hebert-johnson18a.html},
  bibsource    = {dblp computer science bibliography, https://dblp.org}
}

@inproceedings{OI,
  author       = {Cynthia Dwork and
                  Michael P. Kim and
                  Omer Reingold and
                  Guy N. Rothblum and
                  Gal Yona},
  editor       = {Samir Khuller and
                  Virginia Vassilevska Williams},
  title        = {Outcome indistinguishability},
  booktitle    = {{STOC} '21: 53rd Annual {ACM} {SIGACT} Symposium on Theory of Computing,
                  Virtual Event, Italy, June 21-25, 2021},
  pages        = {1095--1108},
  publisher    = {{ACM}},
  year         = {2021},
  url          = {https://doi.org/10.1145/3406325.3451064},
  doi          = {10.1145/3406325.3451064},
  bibsource    = {dblp computer science bibliography, https://dblp.org}
}

@InProceedings{kleinberg17,
  author =	{Kleinberg, Jon and Mullainathan, Sendhil and Raghavan, Manish},
  title =	{{Inherent Trade-Offs in the Fair Determination of Risk Scores}},
  booktitle =	{8th Innovations in Theoretical Computer Science Conference (ITCS 2017)},
  pages =	{43:1--43:23},
  series =	{Leibniz International Proceedings in Informatics (LIPIcs)},
  ISBN =	{978-3-95977-029-3},
  ISSN =	{1868-8969},
  year =	{2017},
  volume =	{67},
  editor =	{Papadimitriou, Christos H.},
  publisher =	{Schloss Dagstuhl -- Leibniz-Zentrum f{\"u}r Informatik},
  address =	{Dagstuhl, Germany},
  URL =		{https://drops.dagstuhl.de/entities/document/10.4230/LIPIcs.ITCS.2017.43},
  URN =		{urn:nbn:de:0030-drops-81560},
  doi =		{10.4230/LIPIcs.ITCS.2017.43}
}

@article{Cho17,
  author = {Chouldechova, Alexandra},
  journal = {Big Data},
  number = 2,
  pages = {153--163},
  publisher = {Mary Ann Liebert, Inc. 140 Huguenot Street, 3rd Floor New Rochelle, NY 10801 USA},
  title = {Fair Prediction with Disparate Impact: A Study of Bias in Recidivism Prediction Instruments.},
  volume = 5,
  year = 2017
}

@article{Daw17,
	author = {Philip Dawid},
	doi = {10.1007/s11229-015-0953-4},
	journal = {Synthese},
	number = {9},
	pages = {3445--3474},
	publisher = {Springer Nature},
	title = {On Individual Risk},
	volume = {194},
	year = {2017}
}

@InProceedings{omni,
  author =	{Gopalan, Parikshit and Kalai, Adam Tauman and Reingold, Omer and Sharan, Vatsal and Wieder, Udi},
  title =	{{Omnipredictors}},
  booktitle =	{13th Innovations in Theoretical Computer Science Conference (ITCS 2022)},
  pages =	{79:1--79:21},
  series =	{Leibniz International Proceedings in Informatics (LIPIcs)},
  ISBN =	{978-3-95977-217-4},
  ISSN =	{1868-8969},
  year =	{2022},
  volume =	{215},
  editor =	{Braverman, Mark},
  publisher =	{Schloss Dagstuhl -- Leibniz-Zentrum f{\"u}r Informatik},
  address =	{Dagstuhl, Germany},
  URL =		{https://drops.dagstuhl.de/entities/document/10.4230/LIPIcs.ITCS.2022.79},
  URN =		{urn:nbn:de:0030-drops-156755},
  doi =		{10.4230/LIPIcs.ITCS.2022.79}
}

@article{HILL99,
author = {H\r{a}stad, Johan and Impagliazzo, Russell and Levin, Leonid A. and Luby, Michael},
title = {A Pseudorandom Generator from any One-way Function},
year = {1999},
issue_date = {Aug. 1999},
publisher = {Society for Industrial and Applied Mathematics},
address = {USA},
volume = {28},
number = {4},
issn = {0097-5397},
url = {https://doi.org/10.1137/S0097539793244708},
doi = {10.1137/S0097539793244708},
journal = {SIAM J. Comput.},
month = mar,
pages = {1364--1396},
numpages = {33}
}

@article{GGM86,
author = {Goldreich, Oded and Goldwasser, Shafi and Micali, Silvio},
title = {How to construct random functions},
year = {1986},
issue_date = {Oct. 1986},
publisher = {Association for Computing Machinery},
address = {New York, NY, USA},
volume = {33},
number = {4},
issn = {0004-5411},
url = {https://doi.org/10.1145/6490.6503},
doi = {10.1145/6490.6503},
journal = {J. ACM},
month = aug,
pages = {792--807},
numpages = {16}
}

@INPROCEEDINGS{ranking,
  author={Dwork, Cynthia and Kim, Michael P. and Reingold, Omer and Rothblum, Guy N. and Yona, Gal},
  booktitle={2019 IEEE 60th Annual Symposium on Foundations of Computer Science (FOCS)}, 
  title={Learning from Outcomes: Evidence-Based Rankings}, 
  year={2019},
  volume={},
  number={},
  pages={106-125},
  doi={10.1109/FOCS.2019.00016}}

@InProceedings{LDRSH18,
  title = 	 {Delayed Impact of Fair Machine Learning},
  author =       {Liu, Lydia T. and Dean, Sarah and Rolf, Esther and Simchowitz, Max and Hardt, Moritz},
  booktitle = 	 {Proceedings of the 35th International Conference on Machine Learning},
  pages = 	 {3150--3158},
  year = 	 {2018},
  editor = 	 {Dy, Jennifer and Krause, Andreas},
  volume = 	 {80},
  series = 	 {Proceedings of Machine Learning Research},
  month = 	 {10--15 Jul},
  publisher =    {PMLR},
  url = 	 {https://proceedings.mlr.press/v80/liu18c.html},
}

@inproceedings{Charlotte,
  title={Provable Uncertainty Decomposition via Higher-Order Calibration},
  author={Ahdritz, Gustaf and Gollakota, Aravind and Gopalan, Parikshit and Peale, Charlotte and Wieder, Udi},
  booktitle={International Conference on Learning Representations},
  year={2025},
  note={To appear}
}

@inproceedings{decisionCali,
 author = {Zhao, Shengjia and Kim, Michael and Sahoo, Roshni and Ma, Tengyu and Ermon, Stefano},
 booktitle = {Advances in Neural Information Processing Systems},
 editor = {M. Ranzato and A. Beygelzimer and Y. Dauphin and P.S. Liang and J. Wortman Vaughan},
 pages = {22313--22324},
 publisher = {Curran Associates, Inc.},
 title = {Calibrating Predictions to Decisions: A Novel Approach to Multi-Class Calibration},
 url = {https://proceedings.neurips.cc/paper_files/paper/2021/file/bbc92a647199b832ec90d7cf57074e9e-Paper.pdf},
 volume = {34},
 year = {2021}
}

@InProceedings{EfficientMultiClassMC,
  title = 	 {On Computationally Efficient Multi-Class Calibration},
  author =       {Gopalan, Parikshit and Hu, Lunjia and Rothblum, Guy N.},
  booktitle = 	 {Proceedings of Thirty Seventh Conference on Learning Theory},
  pages = 	 {1983--2026},
  year = 	 {2024},
  editor = 	 {Agrawal, Shipra and Roth, Aaron},
  volume = 	 {247},
  series = 	 {Proceedings of Machine Learning Research},
  month = 	 {30 Jun--03 Jul},
  publisher =    {PMLR},
  }

@inproceedings{KF15,
author = {Kull, Meelis and Flach, Peter},
title = {Novel decompositions of proper scoring rules for classification: score adjustment as precursor to calibration},
year = {2015},
isbn = {9783319235271},
publisher = {Springer},
address = {Gewerbestrasse 11 CH-6330, Cham (ZG), CHE},
url = {https://doi.org/10.1007/978-3-319-23528-8_5},
doi = {10.1007/978-3-319-23528-8_5},
booktitle = {Proceedings of the 2015th European Conference on Machine Learning and Knowledge Discovery in Databases - Volume Part I},
pages = {68--85},
numpages = {18},
location = {Porto, Portugal},
series = {ECMLPKDD'15}
}

@inproceedings{WLZ19,
  title={Calibration tests in multi-class classification: A unifying framework},
  author={David Widmann and Fredrik Lindsten and Dave Zachariah},
  booktitle={Neural Information Processing Systems},
  year={2019},
  url={https://api.semanticscholar.org/CorpusID:202770409}
}

@inproceedings{KPNK,
author = {Kull, Meelis and Perello-Nieto, Miquel and K\"{a}ngsepp, Markus and Filho, Telmo Silva and Song, Hao and Flach, Peter},
title = {Beyond temperature scaling: obtaining well-calibrated multiclass probabilities with Dirichlet calibration},
year = {2019},
publisher = {Curran Associates Inc.},
address = {Red Hook, NY, USA},
booktitle = {Proceedings of the 33rd International Conference on Neural Information Processing Systems},
articleno = {1103},
numpages = {11}
}

@inproceedings{OmniForRegression,
  title={Omnipredictors for Regression and the Approximate Rank of Convex Functions},
  author={Parikshit Gopalan and Princewill Okoroafor and Prasad Raghavendra and Abhishek Shetty and Mihir Singhal},
  booktitle={Annual Conference on Computational Learning Theory},
  year={2024},
  url={https://api.semanticscholar.org/CorpusID:267301310}
}

@inproceedings{chairethinking,
  title={Rethinking Pareto Frontier: On the Optimal Trade-offs in Fair Classification},
  author={Chai, Junyi and Lu, Shenyu and Wang, Xiaoqian},
  booktitle={The Fourteenth International Conference on Learning Representations},
  year = {2026},
}
\bibliographystyle{plainnat}


\appendix

\section{Appendix roadmap} 
The remainder of the paper is organized as follows. In \Cref{sec:rich_labels} we further discuss the motivation for learning with rich labels and contrast it with the standard Boolean setting. \Cref{sec:prelims} provides the full definitions and preliminaries. In \Cref{sec:MAD} we show how multi-accuracy guarantees for rich predictors translate into decision-accuracy and classification-accuracy guarantees for the induced binary decisions. In \Cref{sec:impossibility_cali_and_AD_requires_affine} we show that calibration can contradict decision-accuracy. In \Cref{sec:impossibility_AC_and_loss_min} we show that loss minimization can contradict classification-accuracy. Finally, in \Cref{sec:MC_for_loss} we show how multi-calibration can be translated into loss minimization via post-processing.

\section{Further discussion on rich labels}\label{sec:rich_labels}

\paragraph{Background: Boolean labels.} Much of the algorithmic fairness literature focuses on scenarios where the labels correspond to Boolean outcomes. The training data comes from a distribution $\mD$ on pairs $(x, y^*)$ where $x$ is a feature vector from a space $\X$ and  $y^* \in \zo$ is a label. For example, $x$ could correspond to a patient's medical history, symptoms, and test results, and $y^* $ could correspond to whether they have experienced significant cardiovascular disease in the subsequent decade. Alternatively, as in the running example given in \Cref{sec:intro}, $x$ could represent the information of a loan applicant, such as income, credit history, and other financial factors, and $y^* $ corresponds to whether the loan was repaid. The purpose of the learning model in such scenarios is to decide on an appropriate action. For example, deciding on a medical intervention such as prescribing cholesterol medication or on whether to approve a loan application. A natural first step in such decisions is to estimate the probability of the outcome conditioned on the features. For example, ``the patient has an 8\% chance of having a heart attack in the next decade'' or ``there is a 30\% chance that the applicant will default on the loan''. We refer to this task as {\em prediction}. Decisions are often based on risk predictions: given a predicted level of risk, a particular decision may or may not be warranted.

In this model of learning, the true probability $p^*(x)=\Pr[y^*=1 \mid x]$ is never provided to the learning algorithm (here, $p^*$ is the Bayes-optimal predictor). Indeed, $p^*$ is often considered completely out of reach, and in fact the meaning of individual probabilities has deep philosophical implications and has been debated in Statistics for generations \citep{Daw17}, with more recent computational perspectives \citep{OI}. Given these challenges, a typical approach is to approximate $p^*$ by an efficient predictor $\tilde{p}$ and to execute some decision rule on $\tilde{p}$ that we would consider appropriate if applied on $p^*$. For example, we could rank all the patients based on $\tilde{p}$ (indicating probability of future heart attack) and invite those with high values to a consultation with their primary care team on making lifestyle changes. Similarly, we can accept loan applications whenever $\tilde{p}$ is larger than 0.95, indicating a tolerable risk for the bank. 

Let us revisit the example from \Cref{sec:intro:tech-overview}, that motivated the study of Decision-Accuracy. For a set of individuals $S$, multi-accuracy-on-decision requires that members of $S$ see the same distribution of decisions when the decision rule is applied to $\tilde{p}$ as they would when it is applied to $p^*$. While this is a very natural notion of accuracy and of fairness, it is often way beyond the reach of a learning algorithm. Consider for example a sample of the outcomes, where half of the members of $S$ repaid the loan and half did not. If there are no efficiently learnable patterns that correlate with loan repayment in $S$, then it could be that every member of $S$ has $p^*$ being half, or perhaps half of the values are one and half are zeros, or perhaps two thirds of the values are $3/4$ and the rest are zeros. Each of these cases could induce a very different distribution on the decisions to members of $S$, say, under the decision rule that gives a positive decision to individuals whose probability of repaying the loan is at least $95\%$.

\paragraph{Our setting: rich labels.}
The above example highlights a fundamental challenge in achieving strong notions of accurate decision rates: we aim to base decisions on risk scores, yet we never observe the true probabilities, not even during training. When comparing two historical loan applicants, for instance, we cannot determine the correct ranking between them based solely on observed outcomes (e.g., repaid vs.\ defaulted). Even if Alice repaid her loan and Bob did not, it is possible that Bob had a higher true probability of repayment under the ground truth distribution $p^*$, but was simply unlucky, while Alice was fortunate despite a lower $p^*$ value. In other words, $p^*$ encodes uncertainty that we cannot directly access; we only observe one realization of it per individual through historical outcomes.

In this work, we study the task of learning given access to richer labels, that reflect this underlying uncertainty, and our goal is to use this richer data to support  more accurate decision and classification rates.  One natural approach is to include the true probabilities $p^*$ for individuals in the training data. An alternative is to include rankings of individuals. If these rankings are derived from probabilities, the two approaches are effectively equivalent. However, rankings can also encode broader considerations beyond simple outcome probabilities. For example, in college admissions, we might assess applicants not only by their probability of graduation but also by their expected contribution to their cohort or to society more broadly. Even more generally, we can think of each individual as having a ``type'' that supports multiple kinds of rankings, depending on the institution's goals, such as different prioritization of student applicants by a university.

Where could rich labels come from? A natural source is historical data annotated by one or more domain experts. Of course, even expert judgments can carry biases, and the very act of defining which ``types'' individuals are categorized into may have fairness implications. Still, experts may possess information unavailable to algorithms---for example, the nuanced understanding a physician might have about a patient's condition. Additionally, the expertise used to generate rich labels can include knowledge of fairness considerations specific to the task or domain. In such cases, these labels can be understood not as representations of an objective ground truth, but rather as a way to encode human values into the learning process.

\section{Preliminaries} \label{sec:prelims}
We model Nature as a joint distribution over individuals and labels, denoted by $\mD$. Individuals are drawn from a discrete domain $\X$; throughout, we assume that each $x \in \X$ can be represented as a Boolean string $x \in \zo^n$, which encodes the features of the individual. For simplicity, we write $n = \log |\X|$ to mean $n = \lceil \log |\X| \rceil$. We use $x \sim \DX$ to denote a sample from Nature's marginal distribution over individuals.

The labels are drawn from an abstract outcome space $\setOfTypes$, which is assumed to be finite, i.e., $k \defeq \left|\setOfTypes\right| < \infty$. We refer to elements of $\setOfTypes$ as {\em types}. Distributions over $\setOfTypes$ are represented using stochastic vectors, denoted by $\stochasticVecs \in \Delta(\setOfTypes)$, where $\Delta\left(\setOfTypes\right)$ is the set of all discrete probability distributions over $\setOfTypes$. The following definition formalizes $\stochasticVecs$.  

\begin{definition}[Stochastic Vectors]\label{def:stochastic_vecs}  
    Let $\setOfTypes$ be a finite outcome space of size $k \in \mathbb{N}$.  
    The set of $k$-dimensional stochastic vectors is defined as  
    \[  
    \stochasticVecs \defeq \left\{ y \in [0,1]^k \text{ such that } \sum_{i=1}^{k} y_i = 1 \right\}.  
    \]    
    The probability of sampling $\type \in \setOfTypes$ from $y \in \stochasticVecs$ is given by its $\type\tth$ coordinate, $y_\type$. We denote this by $\type \sim y$. If $\setOfTypes$ admits a total order, we assume that this order determines the coordinate indexing.
\end{definition}

We define $R^*: \X \to \stochasticVecs$ as the predictor induced by Nature. If Nature is deterministic, then $R^*$ outputs a unit vector in $\stochasticVecs$, corresponding to a one-hot encoding. We denote the $\type\tth$ unit vector by $e_\type$.  
Overall, we express Nature's joint distribution as  
\[
\mD = \DX \times R^*(\DX).
\]
The primary objects of study in this work are {\em probabilistic predictors},  
\[
\tilde{R}: \X \to \stochasticVecs,
\]  
which are intended to approximate $R^*$.  

For a predictor $R$, we use $R(x) \in \stochasticVecs$ to denote its estimated distribution for an individual $x$. A random label sampled according to this distribution is denoted by $\type(x) \sim R(x)$. We assume that each sample from $R$ is drawn independently. The predictor $R$ is said to be deterministic if, for every $x \in \X$, it outputs only unit vectors.

Throughout this work, we evaluate notions of group fairness, which are parameterized by a collection of sets $\C \subseteq \zo^{|\X|}$. Each subset $S \in \C$ is referred to as a {\em subpopulation} or a {\em group}.

\paragraph{Probabilistic predictors for a deterministic Nature.}
The study of probabilistic predictors is well-motivated even when Nature is entirely deterministic. Consider the following example: a subpopulation $S \in \C$ consists of two disjoint groups, $S = S_1 \cup S_2$, where $|S_1| = |S_2| = \frac{1}{2}|S|$, such that $S \in \C$ but $S_1$ and $S_2$ are computationally indistinguishable from each other (in particular, $S_1, S_2 \notin \C$). Suppose that the ground truth predictor $R^*$ assigns label $\type$ to all individuals in $S_1$ and label $\type'$ to all individuals in $S_2$.  

A {\em deterministic} predictor $\tilde{R}$, which outputs only unit vectors, cannot achieve better accuracy or calibration than simply selecting a (pseudo-)random subset of $S$ of half its size to assign label $\type$, while assigning label $\type'$ to the remaining individuals. However, if this predictor were used repeatedly---for instance, by multiple universities in their admission processes---the same individuals would consistently receive the same label, potentially leading to systematic unfairness.  

In contrast, a {\em probabilistic} predictor $\tilde{R}$ could assign label $\type$ or $\type'$ to each individual with equal probability (i.e., $1/2$). As a result, each application of $\tilde{R}$ would yield different labels for the same individuals, ensuring that unfair outcomes do not persist across multiple usages.

\subsection{Multi-Accuracy}\label{sec:MA_defs}
Multi-Accuracy for (deterministic) predictors \citep{MC,TrevisanTV09}, requires a predictor to be accurate across multiple subpopulations of the domain. The goal is to ensure that a predictor's expected output is unbiased with respect to Nature, not just globally but also on structured subpopulations. Multi-accuracy is parameterized by a class of subpopulations $\C$, which controls the strength of the guarantee: the richer the class, the stronger the requirement. 
The following definition extends this notion to the probabilistic setting and requires approximate accuracy on each label in $\setOfTypes$ individually. We therefore refer to it as {\em coordinate-wise} multi-accuracy.

\begin{definition}[Coordinate-Wise-Multi-Accuracy for Probabilistic Predictors]\label{def:eq_MA_for_types}
Let $\C \subseteq \zo^{|\X|}$ be a collection of sets. 
For any error parameter $\alpha \ge 0$, we say that $\tilde{R}$ satisfies $(\C,\alpha)$-coordinate-wise-multi-accuracy w.r.t.\ $\mD$ if for all $S \in \C$ and all $\type \in \setOfTypes$,
\[ 
\left| 
\Pr_{\substack{x \sim \DX \\ \type^*(x) \sim R^*(x)}} \left[\type^*(x) = \type \land x \in S \right] - 
\Pr_{\substack{x \sim \DX \\ \tilde{\type}(x) \sim \tilde{R}(x)}}
\left[\tilde{\type}(x) = \type \land x \in S \right] \right| \le \alpha.
\]
\end{definition}
For the special case where $\setOfTypes$ is an ordered set, we introduce an alternative (and stronger) definition called threshold-multi-accuracy. We denote $\setOfTypes = [k]$ and interpret it as a set of ranks. Note that
we refer to $1$ as the best rank and to $k$ as the worst. To distinguish between sampling a rank and a type, we use the notation $r(x) \sim R(x)$ instead of $\type(x) \sim R(x)$. When Nature is deterministic, we often denote it by $r^*$ and assume that it outputs a rank directly rather than a unit vector. In \Cref{sec:MA_for_rankings}, we provide a detailed discussion on the relationship between different definitions for this special case.

\begin{definition}[Threshold-Multi-Accuracy for Probabilistic Rankings] \label{def:MA_for_prob_ranking}
Let $\C \subseteq \zo^{|\X|}$ be a collection of sets. 
For any error parameter $\alpha \ge 0$, we say that $\tilde{R}$ satisfies  $(\C,\alpha)$-threshold-multi-accuracy w.r.t.\ $\mD$ if for every $S \in \C$ and every $\tau \in [k]$,
\[ 
\left| 
\Pr_{\substack{x \sim \DX \\ r^*(x) \sim R^*(x)}} \left[r^*(x) > \tau \land x \in S \right] - 
\Pr_{\substack{x \sim \DX \\ \tilde{r}(x) \sim \tilde{R}(x)}}
\left[\tilde{r}(x) > \tau \land x \in S \right] \right| \le \alpha.
\]
\end{definition}

\begin{remark}\label{rem:arbitrary_subset_MA}
In \Cref{def:MA_for_prob_ranking}, we define multi-accuracy with respect to threshold events of the form $\{r(x) > \tau\}$. However, both this threshold condition and the coordinate-wise condition in \Cref{def:eq_MA_for_types} can be generalized to arbitrary subsets of labels, without increasing the complexity of learning. This can be useful in cases a loss function is known in advance. See \Cref{rem:avoiding_k_with_arbitrary_subset_MA} for further discussion of the implications for error bounds and runtime.
\end{remark}

In \Cref{sec:MA_for_rankings} we discuss the relationship between threshold-multi-accuracy and coordinate-wise-multi-accuracy. In particular, \Cref{clm:MA_implies_equality_MA} and \Cref{clm:coordinate_wise_implies_thres_with_k_blowup} prove that the threshold definition is strictly stronger. 

Next, we show that accuracy on a set implies accuracy on any large enough subset of it. The claim applies to both multi-accuracy definitions presented above; however, for concreteness, we state it in terms of coordinate-wise-multi-accuracy for types (\Cref{def:eq_MA_for_types}). Its proof appears in \Cref{sec:deferred_MA_MC}.
\begin{claim}\label{clm:accuracy_holds_on_a_large_subset}
    Let $S, S'$ be two sets such that $S' \subseteq S$. If $\tilde{R}$ is $\alpha$-coordinate-wise-accurate w.r.t. $\mD$ on $S$, then it is $\alpha'$-coordinate-wise-accurate w.r.t. $\mD$ on $S'$, where $\alpha' = \alpha + \Pr_{x \sim \DX} \left[ x \in S \setminus S' \right]$.
\end{claim}

Finally, we show that these notions can be learned efficiently. \citet{MC} demonstrated a connection between learning multi-accurate and multi-calibrated predictors and weak agnostic learning over the collection $\C$. In particular, this suggests that structural properties of $\C$ can be exploited, if present. Following their approach, we show that efficient weak agnostic learning over $\C$ implies efficient learning of multi-accurate and multi-calibrated ($k$-dimensional) predictors on $\C$.

\begin{theorem}[\Cref{thm:learning_MA}, simplified]\label{thm:learning_MA_simplified}
If there is a weak agnostic learner for $\C$ that runs in time $T$, then there is an algorithm for learning $(\C,\alpha)$-coordinate-wise-multi-accurate probabilistic predictor $\tilde{R}$ that runs in time $\tilde{O}(k \cdot T / \alpha^6)$ and uses $\tilde{O}(1 / \alpha^2)$ samples.
\end{theorem}
As noted in \Cref{thm:learning_MA}, these complexity bounds also apply to learning threshold-multi-accuracy probabilistic rankings. We refer the reader to \Cref{sec:learning_MA_MC} for full details.

\subsection{Multi-Calibration}\label{sec:MC}

Next, we define Multi-Calibration, a natural strengthening of Multi-Accuracy. While calibration remains meaningful even when allowing arbitrary predictions $\tilde{R}(x) \in \stochasticVecs$, computational feasibility in our constructions necessitates maintaining a certain level of discretization. Formally, we introduce the following definition of $\lambda$-discretization of the $[0,1]$ interval.

\begin{definition}[$\lambda$-discretization] \label{def:lambda}
Let $\lambda > 0$. The \emph{$\lambda$-discretization} of $[0,1]$, denoted by $\Lambda[0,1] = \set{\frac{\lambda}{2},\frac{3\lambda}{2},\dots, 1-\frac{\lambda}{2}}$, is the set of $1/\lambda$ evenly spaced real values over $[0,1]$.  For $v \in \Lambda[0,1]$, let \[\lambda(v) = [v-\lambda/2,v+\lambda/2)\] be the $\lambda$-interval centered around $v$ (except for the final interval, which is $[1-\lambda,1]$).
\end{definition}

To define the values on which we condition to measure approximate calibration, akin to the ``level sets'' in \citet{MC}, we consider two approaches. The first approach conditions on all possible $\lambda$-discretized outputs of the predictor, and is therefore referred to as {\em full}-multi-calibration.
\begin{definition}[Full-Multi-Calibration for Probabilistic Predictors]\label{def:full_MC_for_types}
Let $\C \subseteq \zo^{|\X|}$ be a collection of sets and $\lambda \in (0,1)$ a discretization parameter. 
For any set $S \in \C$ and any $\lambda$-discretized vector $R \in (\Lambda[0,1])^k$, we define $S_R$ w.r.t. $\tilde{R}: \X \to \stochasticVecs$ as:
\[
S_R = \left\{ x \in S \text{ such that } \forall \type \in \setOfTypes,\ \Pr_{\tilde{\type}(x) \sim \tilde{R}(x)} \left[\tilde{\type}(x) = \type\right] \in \lambda\left(R_\type\right) \right\}.
\]
For any error parameter $\alpha \ge 0$, we say that $\tilde{R}$ satisfies $(\C,\alpha,\lambda)$-full-multi-calibration w.r.t. $\mD$ if for all $S \in \C$, $R \in (\Lambda[0,1])^k$ and $\type \in \setOfTypes$,
\[ 
\left| 
\Pr_{\substack{x \sim \DX \\ \type^*(x) \sim R^*(x)}} \left[\type^*(x) = \type \land x \in S_R \right] - 
\Pr_{\substack{x \sim \DX \\ \tilde{\type}(x) \sim \tilde{R}(x)}} \left[\tilde{\type}(x) = \type \land x \in S_R \right] \right| \le \alpha.
\]
\end{definition}
The number of sets to consider is exponential in $k$, i.e., $\left|\left(S_R\right)_{R \in (\Lambda[0,1])^k}\right| = \left( \frac{1}{\lambda} \right)^k$, which for some settings of the parameters could be computationally infeasible. For this reason, we deploy relaxed notions of multi-calibration when they are sufficient. We propose a relaxed definition where approximate calibration is measured after sampling a label from the probabilistic predictor, hence the number of sets to consider is $\left|(S_{\type,\beta})_{\type \in \setOfTypes, \beta \in \Lambda[0,1]} \right| \le  k / \lambda$, which admits an efficient learning procedure as described in \Cref{thm:learning_MC}.

Beyond serving as a natural relaxation of full-multi-calibration, that allows for efficient computation, this notion is used in our first negative result (\Cref{thm:impossibility_result}). Notably, it strengthens the impossibility result by demonstrating it with respect to this relaxed notion rather than full-multi-calibration.  

\begin{definition}[Coordinate-Wise-Multi-Calibration for Probabilistic Predictors]\label{def:relaxed_MC_for_types}
Let $\C \subseteq \zo^{|\X|}$ be a collection of sets and $\lambda \in (0,1)$ a discretization parameter. 
For any set $S \in \C$, label $\type \in \setOfTypes$, and discretized probability $\beta \in \Lambda[0,1]$, define $S_{\type, \beta}$ w.r.t. $\tilde{R}: \X \to \stochasticVecs$ as:
\[
S_{\type, \beta} \defeq \left\{ x \in S \text{ such that }  \Pr_{\tilde{\type}(x) \sim \tilde{R}(x)}[\tilde{\type}(x) = \type] \in \lambda(\beta) \right\}.
\]
For any error parameter $\alpha \ge 0$, we say that $\tilde{R}$ satisfies $(\C,\alpha,\lambda)$-coordinate-wise-multi-calibration w.r.t. $\mD$ if for all $S \in \C$,   $\type \in \setOfTypes$ and $\beta \in \Lambda[0,1]$,
\[ 
\left| 
\Pr_{\substack{x \sim \DX \\ \type^*(x) \sim R^*(x)}} \left[\type^*(x) = \type \land x \in S_{\type, \beta} \right] - 
\Pr_{\substack{x \sim \DX \\ \tilde{\type}(x) \sim \tilde{R}(x)}}
\left[\tilde{\type}(x) = \type \land x \in S_{\type, \beta} \right] \right| \le \alpha.
\]
\end{definition}



The following claim establishes that coordinate-wise-multi-calibration (and, in particular, full multi-calibration) is a stronger requirement than coordinate-wise-multi-accuracy. The proof applies to both the coordinate-wise and threshold-based definitions and is presented in the coordinate-wise setting for concreteness.

\begin{claim}[Coordinate-Wise-Multi-Calibration Implies Coordinate-Wise-Multi-Accuracy] \label{clm:MC_implies_MA}
    In the setting of \Cref{def:relaxed_MC_for_types}, if $\tilde{R}$ is $(\C,\alpha,\lambda)$-coordinate-wise-multi-calibrated, then $\tilde{R}$ is also $(\C,\alpha/\lambda)$-coordinate-wise-multi-accurate.
\end{claim}

To illustrate the motivation for Multi-Calibration, consider again the example where $S = S_1 \cup S_2$, with $S_1, S_2$ indistinguishable and Nature defined such that $R^*(x) = \type$ for all $x \in S_1$ and $R^*(x) = \type'$ for all $x \in S_2$. Compare two predictors:
$(1)$ $\tilde{R}_1$: assigns $\type'$ to all $x \in S_1$ and $\type$ to all $x \in S_2$.
$(2)$ $\tilde{R}_2$: for all $x \in S$,
\[
\tilde{R}_2(x) =
\begin{cases}
\type\; \text{ w.p. } 1/2, \\
\type' \text{ w.p. } 1/2.
\end{cases}
\]
The predictor $\tilde{R}_2$ satisfies perfect coordinate-wise-multi-calibration on $S$ (that is, with error $\alpha = 0$), and even perfect full-multi-calibration (and thus perfect coordinate-wise-multi-accuracy), whereas $\tilde{R}_1$ has a coordinate-wise-multi-calibration error of $1/2$ on $S$, despite being perfectly coordinate-wise-multi-accurate. This example illustrates that while coordinate-wise-multi-accuracy ensures unbiased predictions over subpopulations, only multi-calibration prevents systematic misallocation of probabilities, ensuring that predicted distributions align with the true conditional distributions.  

\bigskip
We conclude this section by showing that coordinate-wise-multi-calibration can be learned efficiently. As in \Cref{thm:learning_MA_simplified}, we formulate this in terms of weak agnostic learning over $\C$.

\begin{theorem}[\Cref{thm:learning_MC}, simplified]\label{thm:learning_MC_simplified}
If there is a weak agnostic learner for $\C$ that runs in time $T$, then there is an algorithm for learning $(\C,\alpha,\lambda)$-coordinate-wise-multi-calibrated probabilistic predictor $\tilde{R}$ that runs in time $\tilde{O}(k \cdot T / (\lambda \alpha^6))$ and uses $\tilde{O}(1 / \alpha^4)$ samples.
\end{theorem}

\subsection{Decision Rules} \label{sec:prelims_for_dec_rules}

The following definitions and claims consider a randomized predicate $f: \stochasticVecs \to \zo$, which we later refer to as a {\em decision rule}.

\begin{definition}[Lipschitz]\label{def:lip_rule}
    Let $M \in \R^+$ be a positive constant. 
    A randomized predicate $f: \stochasticVecs \to \zo$ is called {\em $M$-Lipschitz} if $\forall y,y' \in \stochasticVecs$, 
    \[
    \left| \expectation_{f} \left[ f(y) - f(y') \right] \right| \le M \cdot ||y-y'||_{\infty}.
    \]
\end{definition}

\begin{definition}[$\eps$-Close to Affine]\label{def:close_to_affine}
    Let $\eps \in [0,1]$. 
    A randomized predicate $f: \stochasticVecs \to \zo$ is called {\em $\eps$-close to affine} if for any $y,y' \in \stochasticVecs$ and any $\gamma \in [0,1]$,
    \[
    \left| \expectation_{f} \left[ f\left(\gamma y + (1-\gamma) y'\right) \right] - \left( \gamma \expectation_{f}\left[f(y)\right] + (1-\gamma) \expectation_{f}\left[f(y')\right] \right) \right| \le \eps.
    \]
\end{definition}
A predicate is affine if it is $0$-close to affine, and it is $\eps$-far from affine if it is not $\eps$-close to affine. The following claim shows that the definition with two points (\Cref{def:close_to_affine}) suffices to ensure that the inequality holds with any number of points, with an increase of at most a factor of $2$. The proof of the claim and its corollary can be found in \Cref{sec:deferred_dec_rules}.

\begin{claim}\label{clm:two_points_in_convex_combi_are_enough}
    Let $\numOfVecs \in \N$, $\gamma_1,\dots,\gamma_\numOfVecs \in [0,1]$ s.t. $\sum_{i=1}^\numOfVecs \gamma_i = 1$, and $y_1,\dots,y_\numOfVecs \in \stochasticVecs$. Let $\eps \in [0,1]$ and assume that $f: \stochasticVecs \to \zo$ is $\eps$-close to affine. Then, 
    \[
    \left| \expectation_{f}\left[f\left(\sum_{i=1}^\numOfVecs \gamma_i y_i\right) \right]- \sum_{i=1}^\numOfVecs \gamma_i \expectation_{f}\left[f(y_i)\right] \right| \le 2 \eps.
    \]
\end{claim}

\begin{corollary}\label{cor:close_affine_function}
    If $f: \stochasticVecs \to \zo$ is $\eps$-close to affine, then there exists an affine randomized predicate $f': \stochasticVecs \to \zo$ such that $\forall y \in \stochasticVecs$,
    \[
    |\expectation_{f}[f(y)] - \expectation_{f'}[f'(y)]| \le 2\eps.
    \]
\end{corollary}

The following claim shows that for a Lipschitz predicate, any two points that violate its affinity must be sufficiently separated. Its proof can be found in \Cref{sec:deferred_dec_rules} as well. 

\begin{claim}\label{clm:y_and_y'_are_far}
Let $\eps \in [0,1]$ and $M \in \R^+$. Let $f: \stochasticVecs \to \zo$ be a randomized predicate such that:
\begin{itemize}
    \item $f$ is $M$-Lipschitz;
    \item $f$ is $\eps$-far from affine: that is, there exists $y,y' \in \stochasticVecs$ and $\gamma \in (0,1)$ such that \[
    \left| \expectation_{f}\left[f\left(\gamma y + (1-\gamma) y'\right) \right]- \left( \gamma \expectation_{f}\left[f(y)\right] + (1-\gamma) \expectation_{f}\left[f(y')\right] \right) \right| > \eps.
    \]
\end{itemize}
Then, $|| y - y' ||_{\infty} \ge \eps / M$. 
\end{claim}

\subsection{Indistinguishability}\label{sec:indistin_subsets}
The impossibility results (\Cref{thm:impossibility_result,thm:impossibility_result_of_loss}) rely on the existence of an indistinguishable subset with respect to the uniform distribution. We present two methods for constructing such a subset: one based on cryptographic assumptions, specifically the existence of one-way functions, and another that is information-theoretic, applicable when no limitation is placed on the complexity of Nature. We begin with a formal definition of indistinguishability.

\begin{definition}[Indistinguishable Subset]\label{def:indistin}
    Let $\X$ be a domain and let $\DX$ be a distribution. For a security parameter $\kappa \in \N$, we say that $\emptyset \subsetneq \X_1 \subsetneq \X$ is a $\kappa$-{\em indistinguishable subset} if for any $x \sim \DX$, membership in $\X_1$ is not efficiently decidable (w.r.t. $\kappa$). That is, for every probabilistic $\poly(\kappa)$-time algorithm $\mathcal{A}$ and all sufficiently large $\kappa$,
    \[
    \left| \Pr_{\mathcal{A}, x \sim \DX}\left[ \mathcal{A}(x) = 1 \mid x \in \X_1 \right] - \Pr_{\mathcal{A}, x \sim \DX}\left[ \mathcal{A}(x) = 1 \mid x \in \X \setminus \X_1 \right] \right| = \negli(\kappa),
    \]
    where $\negli(\cdot)$ denotes a negligible function, i.e., a function $\negli:\mathbb{N}\to[0,1]$ such that for every polynomial $p(\cdot)$, there exists $N\in\mathbb{N}$ for which $\negli(\kappa) < 1/p(\kappa)$ for all $\kappa \ge N$.
\end{definition}
Note that the probabilities are well-defined (i.e., we do not condition on an event of probability zero), since the definition requires $\X_1$ to be a non-empty proper subset. Indeed, both the empty set and the entire universe $\X$ are never indistinguishable.

\paragraph{Cryptographic approach.} In what follows, we show how to use cryptographic assumptions to construct an indistinguishable subset of a given size. In particular, we assume the existence of one-way functions (OWFs), often regarded as the ``minimal'' assumption in cryptography. The construction of the subset is computationally efficient; however, under the assumption that OWFs exist, determining membership in the subset is computationally infeasible. The proof is deferred to \Cref{sec:deferred_indistin}.

\begin{lemma}\label{lem:existence_of_X1}
    Let $\X$ be a domain, and let $\UX$ denote the uniform distribution over $\X$. Assume one-way functions exist w.r.t. security parameter $\kappa \ge \log |\X|$. For any constant $\gamma \in (0,1)$, there exists a subset $\X_1 \subset \X$ such that
    \begin{itemize}
        \item $\left| \Pr_{x \sim \UX}\left[x \in \X_1 \right] - \gamma \right| = \negli(\kappa)$;
        \item $\X_1$ is $\kappa$-indistinguishable w.r.t. $\UX$;
        \item There exists a secret key such that, given the key, membership in $\X_1$ is efficiently computable.
    \end{itemize}
\end{lemma}

\paragraph{Information-theoretic approach.} The alternative approach we present is information-theoretic. If we do not impose any computational constraints on the way we model Nature, then determining membership in a subset defined by this circuit can be computationally infeasible. In particular, if Nature is sufficiently ``complex'' to resemble a random function, it may inherently contain an indistinguishable subset. 

We emphasize that this perspective does not require Nature to be a complex function that runs in superpolynomial time; rather, it reflects that, given our {\em partial view} of the feature set defining an individual, Nature appears indistinguishable from a random function. It remains entirely possible that, with access to the {\em full} set of features, Nature could still be computed efficiently.

In this setting as well, for any constant $\gamma \in (0,1)$, we can assume the existence of a Nature that admits an indistinguishable subset of size $\gamma$, up to a negligible rounding error (arising from the chosen binary representation of individuals). 
\bigskip

The following claim applies in both settings, when considering the uniform distribution over the domain $\X$. Its proof is deferred to \Cref{sec:deferred_indistin} as well.  

\begin{claim}\label{clm:subsets_maintain_gamma_fraction}
    Let $\X$ be a domain, and let $\UX$ denote the uniform distribution over $\X$. 
    Assume $\X$ contains a $\kappa$-indistinguishable subset $\X_1$ as per \Cref{def:indistin} w.r.t. a security parameter $\kappa \ge \log|\X|$. For any $\poly(\kappa)$-time algorithm $\mathcal{A}: \X \to \zo$,
    if $\Pr[\mathcal{A}(x) = 1]$ is non-negligible in $\kappa$, then
    \[
    \left| \Pr_{\substack{x \sim \UX \\ \mathcal{A} \text{ randomness}}}[x \in \X_1 | \mathcal{A}(x) = 1] - \Pr_{x \sim \UX}[x \in \X_1] \right| = \negli(\kappa).
    \]
\end{claim}

Notice that an immediate corollary of the claim shows that any efficiently-recognizable subset $\tilX \subseteq \X$ of non-negligible size intersects an indistinguishable subset $\X_1 \subsetneq \X$ in approximately the same fraction as $\X$ does. This follows by interpreting $\mathcal{A}$ as a (deterministic) procedure that decides membership in a set $\tilX$.

\section{Translating accuracy of predictions to accuracy of classifications}\label{sec:MAD}

We now turn to the study of decision rules, also known as post-processing functions. A typical approach to decision-making is to approximate $R^*$ by an efficient predictor $\tilde{R}$ and then apply a decision rule to $\tilde{R}$ in the same manner that one would apply it to $R^*$. For a collection of sets $\C$, we define {\em multi-accuracy-on-decision} to mean that for any set $S \in \C$, members of $S$ receive the same distribution of decisions when the decision rule is applied to $\tilde{R}$ as they would when it is applied to $R^*$. Formally, a decision rule $\decRule: \stochasticVecs \to \zo$ takes as input a distribution over labels and outputs either $0$ or $1$, potentially in a randomized manner. Any randomized predicate, as defined above, can serve as a (randomized) decision rule.

\begin{definition}[Multi-Accuracy-on-Decision]\label{def:MAD}
    For a collection of sets $\C \subseteq \zo^{|\X|}$ and $\eps \ge 0$, a probabilistic predictor $\tilde{R}: \X \to \stochasticVecs$ satisfies $(\C,\eps)$-multi-accuracy-on-decision w.r.t. $\mD$ and $\decRule$ if for all $S \in \C$,
    \[
    \left| \expectation_{x \sim \DX, \decRule} \left[ \decRule\left(\tilde{R}(x)\right) \mid x \in S \right] - \expectation_{x \sim \DX, \decRule} \left[ \decRule\left(R^*(x)\right) \mid x \in S \right] \right| \le \eps.
    \]
\end{definition}

\paragraph{Characterization of affine decision rules.}  
Before stating the result, we present an alternative characterization of affine decision rules: those that first sample a label and then apply a function that (randomly) maps the label to a decision in $\{0,1\}$. This characterization will be useful in the proof of \Cref{thm:MA_implies_MAD}.  

\begin{definition}\label{def:ITA}
    A randomized decision rule $\decRule: \stochasticVecs \to \zo$ is called an {\em Instantiate-Then-Act} rule if there exists a randomized function $g: \setOfTypes \to \zo$ such that $\decRule$ works as follows: On input $y \in \stochasticVecs$, it samples $\type \sim y$ and outputs $g(\type)$. 
\end{definition}

\begin{lemma}\label{lem:equiv_ITA_and_Affine}
    Instantiate-Then-Act (ITA) rules are equivalent to Affine rules (\Cref{def:close_to_affine}). 
\end{lemma}
\begin{proof}
    First, we show that affineness implies ITA. Let $f: \stochasticVecs \to \zo$ be an affine rule. By \Cref{clm:two_points_in_convex_combi_are_enough}, for any $z_1, \dots, z_k \in \stochasticVecs$ and any $\gamma_1, \dots, \gamma_k \in [0,1]$, 
    \[
    \expectation_{f}\left[f\left(\sum_{i=1}^k \gamma_i z_i\right) \right] = \sum_{i=1}^k \gamma_i \expectation_{f}\left[f(z_i)\right].
    \]
    We define $g: \setOfTypes \to \zo$ as follows: for any $\type \in \setOfTypes$, $g(\type)$ is distributed like $f(e_\type)$ where $e_\type$ is the $\type\tth$ unit vector. Notice that, defining $\decRule: \stochasticVecs \to \zo$ such that on input $y$ it samples $\type \sim y$ and outputs $g(\type)$, we get that
    \[
    \Pr_\decRule[\decRule(y) = 1] = \expectation_{\type \sim y} \left[ g(\type) \right] = \sum_{\type \in \setOfTypes} \Pr_y[\type = \type'] \expectation\left[ g(\type) \right] = \sum_{\type \in \setOfTypes} y_\type \expectation\left[ g(\type) \right] = \sum_{\type \in \setOfTypes} y_\type \expectation\left[ f(e_\type) \right]
    \]
    and on the other hand, 
    \[
    \Pr_f\left[ f(y) = 1 \right] = \expectation\left[ f\left( \sum_{\type \in \setOfTypes} y_\type e_\type \right) \right] = \sum_{\type \in \setOfTypes} y_\type \expectation[f(e_\type)].
    \]
    Hence, $f$ and $\decRule$ are identically distributed.  
    Next, we show that ITA implies affineness. Let $g: \setOfTypes \to \zo$ and let $\decRule$ be as in \Cref{def:ITA}. We define $f:\stochasticVecs\to\zo$ so that, on input $y\in\stochasticVecs$, the random variable $f(y)$ follows a $\text{Ber}\left( \sum_{\type \in \setOfTypes} y_{\type}\expectation[g(\type)] \right)$ distribution. It follows directly that $f$ and $\decRule$ are identically distributed. To see that $\decRule$ is affine, let $y,y' \in \stochasticVecs$ and $\gamma \in [0,1]$.
    Notice that
    \begin{align*}
    \expectation \left[ f\left(\gamma y + (1-\gamma) y'\right) \right] & = \sum_{\type \in \setOfTypes} \left( \gamma y_\type + (1-\gamma) y'_\type \right) \expectation[g(\type)] \\ & =
    \gamma \sum_{\type \in \setOfTypes} y_\type \expectation[g(\type)] + (1-\gamma) \sum_{\type \in \setOfTypes} y'_\type \expectation[g(\type)]
    \\& = \gamma \expectation \left[f(y)\right] + (1-\gamma) \expectation\left[f(y')\right],
    \end{align*}
    and the lemma follows. 
\end{proof}

The following theorem shows that a coordinate-wise-multi-accurate (coordinate-wise-MA) predictor, as defined in \Cref{def:eq_MA_for_types}, implies multi-accuracy-on-decision (MAD) with respect to any decision rule that is close to affine.

\begin{theorem}[coordinate-wise-MA implies MAD]\label{thm:MA_implies_MAD}
Fix $\C \subseteq \zo^{|\X|}$ and let $\delta \defeq \min_{S \in \C} \Pr_{x \sim \DX} [x \in S]$.  
Let $\decRule : \stochasticVecs \to \zo$ be a decision rule that is $\eps$-close to affine. If a predictor $\tilde{R}$ is $(\C,\alpha)$-coordinate-wise-multi-accurate w.r.t. $\mD$, then it is $(\C,\eps')$-multi-accurate-on-decision w.r.t. $\mD$ and $\decRule$, where $\eps' \le k\alpha/\delta + 2\eps$.  

Furthermore, if $\setOfTypes$ is an ordered set, $\tilde{R}$ is $(\C,\alpha)$-threshold-multi-accurate and $\decRule$ is deterministic and monotone on unit vectors, then $\eps' \le \alpha/\delta + 2\eps$.  
\end{theorem}

\begin{proof}
    By \Cref{cor:close_affine_function}, there exists an affine decision rule $\decRule'$ such that $\forall y \in \stochasticVecs$, 
    \[
  \left| \expectation_{\decRule'}[\decRule'(y)] - \expectation_{\decRule}[\decRule(y)] \right| \le 2\eps.
    \]
    Since $\decRule'$ is affine, \Cref{lem:equiv_ITA_and_Affine} implies that it can also be viewed as an Instantiate-Then-Act (ITA) rule. That is, there exists a randomized function $g: \setOfTypes \to \zo$ such that $\decRule'$ works as follows: On input $y \in \stochasticVecs$, it samples $\type \sim y$ and outputs $g(\type)$. This implies that, for any $S \in \C$,
    \begin{align*}
    & \left| \expectation_{x \sim \DX, \decRule'} \left[ \decRule'\left(\tilde{R}(x)\right) \mid x \in S \right] - \expectation_{x \sim \DX, \decRule'} \left[ \decRule'\left(R^*(x)\right) \mid x \in S \right] \right| = \\ & 
    \left| \expectation_{x, \tilde{\type}(x), g} \left[ g\left(\tilde{\type}(x)\right) \mid x \in S \right] - \expectation_{x, \type^*(x), g} \left[ g\left(t^*(x)\right) \mid x \in S \right] \right| = \\ & 
    \left| \sum_{\type \in \setOfTypes} \left(\Pr_{x, \tilde{\type}(x)}[\tilde{\type}(x) = \type \mid x \in S ] -  \Pr_{x, \type^*(x)}[\type^*(x) = \type \mid x \in S ]\right) \expectation_g[g(\type)] \right| \le \\ &
    \sum_{\type \in \setOfTypes} \left|\Pr_{x, \tilde{\type}(x)}[\tilde{\type}(x) = \type \mid x \in S ] -  \Pr_{x, \type^*(x)}[\type^*(x) = \type \mid x \in S ]\right| \le \frac{k \alpha}{\delta}.
    \end{align*}
    This means that 
    \[
    \left| \expectation \left[ \decRule\left(\tilde{R}(x)\right) \mid x \in S \right] - \expectation\left[ \decRule\left(R^*(x)\right) \mid x \in S \right] \right| \le \frac{k \alpha}{\delta} + 2\eps.
    \]
    
    For the furthermore part, without loss of generality let us assume that $\setOfTypes = [k]$. By the assumption, $\decRule$ is deterministic and monotone on unit vectors. Notice that $\decRule'$ as defined in \Cref{cor:close_affine_function} coincides with $\decRule$ on unit vectors. On the other hand, expressing $\decRule'$ as an ITA rule, $\decRule'$ also coincides with $g$ on unit vectors, meaning that for all $\tau \in [k]$,
    \[
    g(\tau) = \decRule(e_\tau).
    \]
    This implies that $g$ is both deterministic and monotone on unit vectors, i.e., 
    \begin{equation}\label{eq:g_is_mono}
    \forall \tau \in [k],\  \expectation_g g(\tau) \in \zo \text{ and } \forall \tau_1 < \tau_2,\ g(\tau_1) \ge g(\tau_2).
    \end{equation}
Therefore, $g$ can be interpreted as a threshold function, meaning there exists $\bar{\tau} \in [k]$ such that\footnote{This does not cover the case where $\decRule' \equiv 0$, but in that case, the argument follows trivially.}  
    \[
    g(\tau) = \begin{cases}
        1 & \text{if } \tau \le \bar{\tau}, \\
        0 & \text{otherwise.}
    \end{cases} 
    \]
    Hence, in this case, MAD translates to
    \begin{align*}
    & \left| \expectation_{x, \tilde{r}(x), g} \left[ g\left(\tilde{r}(x)\right) \mid x \in S \right] - \expectation_{x, r^*(x), g} \left[ g\left(r^*(x)\right) \mid x \in S \right] \right| = \\ &
    \left| \Pr_{x, \tilde{r}(x)}
\left[\tilde{r}(x) \le \bar{\tau} \mid x \in S \right] - \Pr_{x, r^*(x)} \left[r^*(x) \le \bar{\tau} \mid x \in S \right]
 \right|,
    \end{align*}
    which removes the $k$ multiplicative blowup, if $\tilde{R}$ is threshold-multi-accurate according to \Cref{def:MA_for_prob_ranking}.
Importantly, as noted in \Cref{sec:MA_defs}, threshold-multi-accuracy can be achieved with the same sample and time complexity as coordinate-wise multi-accuracy, despite being a mathematically stronger notion.  
\end{proof}

\begin{remark}\label{rem:avoiding_k_with_arbitrary_subset_MA}
As mentioned in \Cref{rem:arbitrary_subset_MA}, multi-accuracy can alternatively be required with respect to a subset or a collection of subsets of labels. Referring back to \Cref{eq:g_is_mono}, such definitions correspond to deterministic functions $g$ (i.e., functions that output either $0$ or $1$ for each label). As in the threshold-based case, this eliminates the multiplicative factor of $k$ in the error bound.
\end{remark}

Next, we define Multi-Accuracy for Classification. To this end, we first formally define loss minimization for an action function. An action function is a randomized mapping $\actionFunction: \X \to \zo$ that assigns individuals to either $0$ or $1$, which can be interpreted as ``accepted'' or ``rejected.'' An action function {\em may} be constructed via a decision rule composed with a predictor, but it can also be defined directly. More generally, decision rules or action functions may have more than two possible actions, but we restrict attention to the binary case for simplicity.

\begin{definition}[Loss Minimization]\label{def:loss_minimization}
Let $\loss: \setOfTypes \times \zo \to [0,1]$ be a loss function, and let $\actionFunctionsClass \subseteq \{\X \to \zo\}$ be a class of action functions. For $\eps \ge 0$, an action function $\actionFunction$ is said to be $\left(\actionFunctionsClass,\loss,\eps \right)$-loss-minimizing if  
\[
\forall \actionFunction' \in \actionFunctionsClass,\ \expectation_{\substack{x 
    \sim \DX \\ \actionFunction\ \text{randomness} \\ \type^*(x) \sim R^*(x)}} \left[\loss\left(\type^*(x), \actionFunction(x)\right) \right] \le \expectation_{\substack{x 
    \sim \DX \\ \actionFunction'\ \text{randomness} \\ \type^*(x) \sim R^*(x)}} \left[\loss(\type^*(x), \actionFunction'(x)) \right] + \eps.
\]
\end{definition}
We denote the expected loss of $\actionFunction$ over $\mD$ by $\expLoss(\actionFunction)$. Any loss function induces a loss-minimizing decision rule, which we formally define as follows:
\begin{definition}[Loss-Minimizing Decision Rule]\label{def:loss_opti_dec_rule}
Let $\loss: \setOfTypes \times \zo \to [0,1]$ be a loss function. The {\em loss-minimizing decision rule} $\decRule^*_\loss : \stochasticVecs \to \zo$ with respect to $\loss$ is defined as
\[
\decRule^*_\loss(y) =
\begin{cases}
    1 & \text{if } \expectation_{\type \sim y} \left[\loss(\type, 1) - \loss(\type, 0) \right] < 0, \\
    0 & \text{otherwise.}
\end{cases}
\]
\end{definition}
Equivalently, we can express this as  
\[
\decRule^*_\loss(y) = \argmin_{a \in \zo} \expectation_{\type \sim y}[\loss(\type,a)],
\]
where ties---cases in which the expected losses for acceptance and rejection are equal---are broken in favor of rejection.
Notice that this decision rule is deterministic, and independent of $\mD$. We additionally define the loss-minimizing action function, which is obtained by applying the loss-minimizing decision rule to Nature's predictor. This action function achieves the minimum possible loss compared to any other action function. In fact, this motivates our definition of the loss-minimizing decision rule: it makes optimal decisions under the assumption that Nature's predictor provides the true underlying probabilities.

\begin{definition}[Loss-Minimizing Action Function]\label{def:loss_opti_action_func}
Let $\decRule^*_\loss : \stochasticVecs \to \zo$ be as in \Cref{def:loss_opti_dec_rule}. The {\em loss-minimizing action function} $\actionFunction^*_\loss : \X \to \zo$ w.r.t. $\loss$ is defined as
\[
\actionFunction^*_\loss = \decRule^*_\loss \circ R^*.
\]
\end{definition}

Note that $\actionFunction^*_\loss$ is deterministic as well. 
The following fairness criterion measures the accuracy of representation across different subpopulations {\em after classification}. We say that an action function $\actionFunction$ is Accurate-on-Classification if the fraction of individuals accepted according to $\actionFunction$ is close to the fraction accepted by $\actionFunction^*_\loss$. We formally define this measure in the general multigroup setting. 

\begin{definition}[Multi-Accuracy-on-Classification] \label{def:MAC}
Let $\loss: \setOfTypes \times \zo \to [0,1]$ be a loss function, and let $\C \subseteq \zo^{|\X|}$ be a collection of sets. We say that $\actionFunction: \X \to \zo$ is $(\C,\loss,\eps)$-multi-accurate-on-classification w.r.t. $\mD$ if for all $S \in \C$, 
    \[
    \left| \expectation_{\substack{x \sim \DX \\ \actionFunction\ randomness}} \left[ \actionFunction(x) \mid x \in S \right] - \expectation_{x \sim \DX} \left[ \actionFunction^*_\loss(x) \mid x \in S \right] \right| \le \eps.
    \]
\end{definition}
We remark that one could define Multi-Accuracy-on-Classification (MAC) with respect to an arbitrary decision rule~$\decRule^*$ instead of $\decRule^*_\loss$, by requiring that the fraction of individuals accepted according to $\actionFunction$ is close to the fraction accepted by $\decRule^* \circ R^*$. Achieving MAC under this alternative definition can be done using the same argument presented next in \Cref{cor:MA_implies_MAC}.

\bigskip
We now show that MAC action functions can be obtained efficiently. As a corollary of \Cref{thm:MA_implies_MAD}, we define an affine decision rule and show that composing it with any multi-accurate predictor yields a MAC action function. In particular, this implies that a single predictor can simultaneously satisfy multi-accuracy-on-decision (MAD) with respect to a decision rule, and satisfy MAC when composed with that rule. As shown in the following corollary, the rule we define can be viewed as an affine analogue of the loss-minimizing decision rule.

Notice that this corollary holds only when Nature is deterministic, i.e., $R^*$ outputs unit vectors for any $x \in \X$. See \Cref{rem:deterministic_nat} for more details on this point.
\begin{corollary}[coordinate-wise-MA implies MAC]\label{cor:MA_implies_MAC}
    Let $\loss$ be a loss function and assume $R^*$ is deterministic. 
    In the setting of \Cref{thm:MA_implies_MAD}, there exists a decision rule $\decRule$ such that for any $(\C,\alpha)$-coordinate-wise-multi-accurate predictor $\tilde{R}$, the action function $\actionFunction = \decRule \circ \tilde{R}$ is $(\C,\loss,\eps)$-multi-accurate-on-classification for $\eps \le k\alpha/\delta$. 
\end{corollary}
\begin{proof}
We define the function $g: \setOfTypes \to \zo$ as follows:
\[
\forall \type \in \setOfTypes,\ g(\type) = \decRule^*_\loss(e_\type) = \argmin_{a \in \zo} \{\loss(\type,a)\},
\]
where $e_\type$ is the $\type\tth$ unit vector.
Let $\decRule : \stochasticVecs \to \zo$ be defined as an ITA rule (\Cref{def:ITA}) w.r.t. $g$, that is: on input $y \in \stochasticVecs$, it samples $\type \sim y$ and outputs $g(\type)$. Since $\decRule$ is affine, then \Cref{thm:MA_implies_MAD} implies that $\tilde{R}$ is $(\C,\eps)$-MAD w.r.t. $\mD$ and $\decRule$, where $\eps \le k\alpha/\delta$. That is, for any $S \in \C$,
\[
\left| \expectation_{x \sim \DX, \decRule} \left[ \decRule\left(\tilde{R}(x)\right) \mid x \in S \right] - \expectation_{x \sim \DX, \decRule} \left[ \decRule\left(R^*(x)\right) \mid x \in S \right] \right| \le \eps.
\]
On the other hand, since Nature is deterministic, it holds for any $x \in S$ that
\[
\decRule\left(R^*(x)\right) = \decRule^*_\loss \left(R^*(x)\right) = \actionFunction^*_\loss(x).
\]
Putting it together, 
\[
\left| \expectation_{x \sim \DX, \decRule} \left[ \actionFunction(x) \mid x \in S \right] - \expectation_{x \sim \DX, \decRule} \left[ \actionFunction^*_\loss(x) \mid x \in S \right] \right| \le \eps.
\]
Once again, if the definition of multi-accuracy implies $\eps$-MAD w.r.t. $\mD$ and $\decRule$ with $\eps \le \alpha/\delta$, then $\actionFunction$ achieves $(\C,\loss,\eps)$-MAC with $\eps \le \alpha/\delta$ as well. 
\end{proof}

To conclude this section, we show that MAD can be obtained efficiently for {\em any} decision rule, not necessarily affine, under the same assumption that Nature is deterministic (see \Cref{rem:deterministic_nat} for further discussion). Compared with \Cref{thm:MA_implies_MAD}, this implies the following: in the deterministic setting, achieving MAD for an arbitrary decision rule reduces to the task of learning a multi-accurate predictor---a task we have already shown can be performed efficiently. In contrast, in the probabilistic setting, the uncertainty can be overcome only for affine rules.

To formally state this argument, we define a standard primitive known as {\em random instantiation}.

\begin{definition}[Random Instantiation]
Let $\mathcal{E} \subseteq \stochasticVecs$ denote the set of unit vectors over $\setOfTypes$. Let $R: \X \to \stochasticVecs$ be a probabilistic predictor. A {\em random instantiation} of $R$ is a function $r: \X \to \mathcal{E}$ defined by independently sampling $r(x) \sim R(x)$ for each $x \in \X$, and returning the corresponding unit vector. We denote this by $r \sim R$.
\end{definition}

The proof is straightforward and follows the same logic presented in \Cref{thm:MA_implies_MAD}. Notice that \Cref{rem:avoiding_k_with_arbitrary_subset_MA} applies to \Cref{lem:MA_implies_MAD_for_non_affine} as well: the multiplicative factor of $k$ in the error bound can be avoided by adopting a different variant of multi-accuracy, such as threshold-multi-accuracy or multi-accuracy with respect to a collection of label subsets.

\begin{lemma}[Coordinate-Wise Multi-Accuracy Implies MAD for Any Rule]\label{lem:MA_implies_MAD_for_non_affine}
Assume $R^*$ is deterministic. 
In the setting of \Cref{thm:MA_implies_MAD}, for any decision rule $\decRule$ and any random instantiation $\tilde{r} \sim \tilde{R}$ of a $(\C,\alpha)$-coordinate-wise-multi-accurate predictor $\tilde{R}$, the predictor $\tilde{r}$ is $(\C,\eps)$-multi-accurate-on-decision with respect to $\mD$ and $\decRule$, where $\eps \le k\alpha/\delta$.
\end{lemma}

\begin{proof}
Let $\tilde{r} \sim \tilde{R}$ be a random instantiation of a $(\C,\alpha)$-coordinate-wise-multi-accurate predictor $\tilde{R}$, and recall that $e_\type \in \mathcal{E}$ denotes the $\type\tth$ unit vector. Then, for any $S \in \C$ and any $\type \in \setOfTypes$, we can write
\[
\left| 
\Pr_{x \sim \DX} \left[R^*(x) = e_\type \land x \in S \right] - 
\Pr_{x \sim \DX} 
\left[\tilde{r}(x) = e_\type \land x \in S \right] \right| \le \alpha,
\]
since $R^*$ is deterministic. 
Hence, for any decision rule $\decRule$,
\begin{align*}
& \left| \expectation_{x \sim \DX, \decRule} \left[ \decRule\left(\tilde{r}(x)\right) \mid x \in S \right] - \expectation_{x \sim \DX, \decRule} \left[ \decRule\left(R^*(x)\right) \mid x \in S \right] \right| = \\
& \left| \sum_{\type \in \setOfTypes} \left(\Pr_{x}[\tilde{r}(x) = e_\type \mid x \in S ] -  \Pr_{x}[R^*(x) = e_\type \mid x \in S ]\right) \expectation_\decRule[\decRule(e_\type)] \right| \le \\
& \sum_{\type \in \setOfTypes} \left| \Pr[\tilde{r}(x) = e_\type \mid x \in S ] -  \Pr[R^*(x) = e_\type \mid x \in S ] \right| \le \frac{k \alpha}{\delta}.
\end{align*}
This implies that the MAD error of $\tilde{r}$ is at most $k\alpha/\delta$, and the lemma follows.
\end{proof}

\begin{remark}[Deterministic Nature] \label{rem:deterministic_nat}
As discussed in the introduction, we consider a rich set of labels that capture the uncertainty of Nature, such as values of the Bayes-optimal predictor, rather than binary outcomes. With such labels, it is appropriate to model Nature as deterministic. A randomized Nature with such rich labels poses a greater challenge than the standard agnostic learning model, and thus we cannot hope to achieve the strong fairness guarantees sought in this work unless Nature is assumed to be deterministic.
\end{remark}

\section{When accuracy of predictions contradicts accuracy of decisions} \label{sec:impossibility_cali_and_AD_requires_affine}
In this section, we present our first impossibility result. The following theorem establishes that if a decision rule is far from affine, then there exists a setting of Nature in which it is computationally infeasible to efficiently learn a predictor that is both calibrated and accurate-on-decision, even for a single subpopulation. 

Notably, the calibration requirement used here follows the relaxed definition in \Cref{def:relaxed_MC_for_types}, thereby strengthening the result by relying on a weaker assumption. Moreover, in contrast to simplified approaches, our proof holds for any discretization parameter $\lambda$ and avoids trivial bounds on the error parameters. See \Cref{rem:gamma_rand,rem:relax_the_lip} for further discussion.

Then, in \Cref{sec:loss_mini_far_from_aff}, we show that the loss-minimizing decision rule is far from affine, and that this implies that it may be computationally infeasible to learn a calibrated predictor that is accurate-on-decision with respect to the loss-minimizing decision rule.

\begin{theorem}[Calibration and Accuracy-on-Decision Requires Affineness] \label{thm:impossibility_result}
    Assume one-way functions exist w.r.t. a security parameter $\kappa \ge \max\left\{\log|\X|, O((M/\eps)^2) \right\}$. Let $\eps \in (0,1)$ and $M \in \R^+$. 
    Let $\lambda \in (0,1)$ be a discretization parameter satisfying $\lambda \le \eps/16M$ and let $\alpha \ge 0$ be an error parameter such that $\alpha = O(\lambda^2 \eps / kM)$. 

    If $\decRule : \stochasticVecs \to \zo$ is $M$-Lipschitz and $\eps$-far from affine, then there exists $\mD$ such that there is no $\poly(\kappa)$-time learner that can output with non-negligible probability a predictor $\tilde{R}$ which is both $(\alpha,\lambda)$-coordinate-wise-calibrated w.r.t. $\mD$ and $\eps/2$-accurate-on-decision w.r.t. $\mD$ and $\decRule$. 
\end{theorem}

For convenience, we instantiate the assumption that a computationally hidden subpopulation exists by assuming one-way functions. However, as discussed in \Cref{sec:indistin_subsets}, the assumption of one-way functions can be removed if no restriction is placed on the complexity of Nature.

\begin{proof}
    Let $\decRule : \stochasticVecs \to \zo $ be a randomized decision rule which is $M$-Lipschitz and $\eps$-far from affine (see \Cref{def:lip_rule,def:close_to_affine}). This means that there exists $y,y' \in \stochasticVecs$ and $\gamma \in (0,1)$ such that 
    \[
    \left| \expectation\left[\decRule\left(\gamma y + (1-\gamma) y'\right)\right] - \left( \gamma \expectation\left[\decRule(y)\right] + (1-\gamma) \expectation\left[\decRule(y')\right] \right) \right| > \eps.
    \]
    Let $\UX$ denote the uniform distribution over $\X$. 
    First, we perturb $\gamma$ by adding uniformly random noise $\nu \sim \mathrm{Uniform} \left[ -\frac{M}{\eps \kappa}, \frac{M}{\eps \kappa}\right]$. Let \begin{equation}\label{eq:noisy_gamma}
    \noisyGa = \gamma + \nu
    \end{equation}
    denote the perturbed version of $\gamma$. This ``randomization'' helps mitigate discretization issues when analyzing how $R^*$ behaves on subsets defined by $\tilde{R}$. A more detailed discussion appears in \Cref{rem:gamma_rand}.

    Under the assumption that one-way functions exist, \Cref{lem:existence_of_X1} implies the existence of a subset $\X_1 \subseteq \X$ of size $|\X_1| = \gamma' |\X|$ for $\left| \gamma' - \noisyGa \right| = \negli(\kappa)$, such that $\X_1$ is $\kappa$-indistinguishable w.r.t. $\UX$ (see \Cref{def:indistin}) and Nature is efficiently computable. 
    Taking $\X_2 \defeq \X \setminus \X_1$, \[ \X = \X_1 \cupdot \X_2. \]
    We define Nature's predictor w.r.t. $\UX$ as follows:
    \[
    R^*|_{\X_1} \equiv y \text{ and } R^*|_{\X_2} \equiv y',
    \]
    and define $\mD = (\UX, R^*(\UX))$.
    Towards contradiction, assume there exists a $\poly(\kappa)$-time learner $\learner$, with access to labeled samples from $\mD$, that outputs with non-negligible probability a probabilistic predictor which is both $(\alpha,\lambda)$-coordinate-wise-calibrated w.r.t. $\mD$ (see \Cref{def:relaxed_MC_for_types}) and $\eps/2$-accurate-on-decision w.r.t. $\mD$ and $\decRule$ (see \Cref{def:MAD}). Take $\tilde{R}: \X \to \stochasticVecs$ to denote this predictor. Let $\left( \X_{\type,\beta} \right)_{\type \in \setOfTypes,\beta\in \Lambda[0,1]}$ be the ``level sets'' as per the definition of coordinate-wise-calibration:
    \[
    \X_{\type, \beta} \defeq \left\{ x \in \X \text{ such that }  \Pr_{\tilde{\type}(x) \sim \tilde{R}(x)}[\tilde{\type}(x) = \type] \in \lambda(\beta) \right\}.
    \]
    We stress that these sets are not necessarily disjoint, however, for a fixed $\type$, they form a partition: $\X = \X_{\type, \frac{\lambda}{2}} \cupdot \dots \cupdot \X_{\type, 1-\frac{\lambda}{2}}$.
    First, we argue that calibration implies that for any label $\type$, only one level set can be large. 
    \begin{lemma}\label{lemma:only_one_set_is_large}
        For any $\type \in \setOfTypes$, with all but a negligible probability, there exists a unique $\beta^*_\type \in \Lambda[0,1]$ for which $\left|\X_{\type,\beta^*_\type}\right| > \frac{2\alpha}{\lambda} |\X|$.
    \end{lemma}
    \begin{proof}
        Let $\type \in \setOfTypes$ and $\beta\in \Lambda[0,1]$ such that $\left|\X_{\type,\beta}\right| > \frac{2\alpha}{\lambda}$.
        By $(\alpha,\lambda)$-coordinate-wise-calibration of $\tilde{R}$,
        \[ 
        \left| 
        \Pr_{\substack{x \sim \UX \\ \type^*(x) \sim R^*(x)}} \left[\type^*(x) = \type \land x \in \X_{\type, \beta} \right] - 
        \Pr_{\substack{x \sim \UX \\ \tilde{\type}(x) \sim \tilde{R}(x)}}
        \left[\tilde{\type}(x) = \type \land x \in \X_{\type, \beta} \right] \right| \le \alpha,
        \]
        which implies that 
        \begin{equation}\label{eq:calibration_of_R_tilde}
        \big| 
        \Pr\left[\type^*(x) = \type \mid x \in \X_{\type, \beta} \right] - 
        \Pr\left[\tilde{\type}(x) = \type \mid x \in \X_{\type, \beta} \right] \big| \le \frac{\alpha}{\Pr\left[x \in \X_{\type, \beta} \right]} < \frac{\lambda}{2}.
        \end{equation}
        We observe that the first term can be expressed in terms of $y$ and $y'$. Taking $y_\type$ to denote the $\type\tth$ coordinate of $y$, 
        \[y_\type = \Pr\left[\type^*(x) = \type \mid x \in \X_{\type, \beta} \cap \X_1 \right] \text{ and } y'_\type = \Pr\left[\type^*(x) = \type \mid x \in \X_{\type, \beta} \cap \X_2 \right],\] hence,
        \begin{equation}\label{eq:decompose_R_star}
        \Pr\left[\type^*(x) = \type \mid x \in \X_{\type, \beta} \right] = 
        y_\type \cdot \frac{\Pr\left[\X_{\type, \beta} \cap \X_1 \right]}{\Pr\left[x \in \X_{\type, \beta} \right] } +
        y'_\type \cdot \frac{\Pr\left[\X_{\type, \beta} \cap \X_2 \right]}{\Pr\left[x \in \X_{\type, \beta} \right] }.
        \end{equation}
    Let us denote 
    \[
    \gamma_{\type,\beta}
    \defeq \frac{\Pr\left[\X_{\type, \beta} \cap \X_1 \right]}{\Pr\left[x \in \X_{\type, \beta} \right] },
    \] 
    so we can rewrite the right-hand side of \Cref{eq:decompose_R_star} as $\gamma_{\type,\beta} \cdot y_\type + (1-\gamma_{\type,\beta}) \cdot y'_\type$.

    Recall that the level sets are defined with respect to {\em discretized} probabilities. Even if, for a fixed $\beta$, the probabilities 
    \[
    \gamma' \cdot y_\type + (1-\gamma') \cdot y'_\type
    \]
    and
    \[
    \gamma_{\type,\beta} \cdot y_\type + (1-\gamma_{\type,\beta}) \cdot y'_\type
    \]
    are very close, they may still be mapped to different $\lambda$-intervals by the discretization (see \Cref{def:lambda}). To prevent this, we introduced random noise to $\gamma$ and defined $R^*$ with respect to $\noisyGa$ (see \Cref{rem:gamma_rand} for a detailed discussion). Consequently, $\gamma'$ and $\left(\gamma_{\type,\beta}\right)_{\beta \in \Lambda[0,1]}$ are also random variables with respect to the noise $\nu$.
    
    Our goal is to show that $\left(\gamma_{\type,\beta} \cdot y_\type + (1-\gamma_{\type,\beta}) \cdot y'_\type \right)$ is independent of $\beta$--that is, although each $\beta$ has a different $\gamma_{\type,\beta}$, all these convex combinations are mapped to the same discretized value $\beta^*_\type$ to which $\left(\gamma' \cdot y_\type + (1-\gamma') \cdot y'_\type \right)$ is mapped.
    To this end, define
    \[
    \gamma_{\type,\text{max}} = \argmax_{\gamma_{\type,\beta},\, \beta \in \Lambda[0,1]} \left| \left(\gamma' \cdot y_\type + (1-\gamma') \cdot y'_\type \right)-  \left(\gamma_{\type,\beta} \cdot y_\type + (1-\gamma_{\type,\beta}) \cdot y'_\type \right) \right|,
    \]
    and let $\beta^*_\type \in \Lambda[0,1]$ be such that
    \[
    \left(\gamma' \cdot y_\type + (1-\gamma') \cdot y'_\type \right) \in \lambda(\beta^*_\type).
    \]
    We seek to bound the probability that the convex combination with respect to $\gamma_{\type,\text{max}}$ is not mapped to $\beta^*_\type$.

    \begin{claim}\label{clm:discretization_is_good_whp}
        $\Pr_{\nu} \left[\left(\gamma_{\type,\text{max}} \cdot y_\type + (1-\gamma_{\type,\text{max}}) \cdot y'_\type \right) \notin \lambda(\beta^*_\type) \right] = \negli(\kappa)$.
    \end{claim}
    \begin{proof}
    First, observe that for all $\beta$,
    \[
    \left|\left(\gamma' \cdot y_\type + (1-\gamma') \cdot y'_\type \right)-  \left(\gamma_{\type,\beta} \cdot y_\type + (1-\gamma_{\type,\beta}) \cdot y'_\type \right)\right| \le 
    \left|\gamma_{\type,\text{max}} - \gamma' \right| = \negli(\kappa),
    \]
    where the last inequality follows from \Cref{clm:subsets_maintain_gamma_fraction}. Similarly,
    \[
    \left| \left(\noisyGa \cdot y_\type + (1-\noisyGa) \cdot y'_\type \right) - \left(\gamma' \cdot y_\type + (1-\gamma') \cdot y'_\type \right) \right| \le 
    \left|\gamma' - \noisyGa \right| = \negli(\kappa).
    \]
    Thus, combining both inequalities,
    \[
    \left| \left(\noisyGa \cdot y_\type + (1-\noisyGa) \cdot y'_\type \right) - \left(\gamma_{\type,\beta} \cdot y_\type + (1-\gamma_{\type,\beta}) \cdot y'_\type \right) \right| = \negli(\kappa).
    \]
    This implies that the ``bad'' event occurs only if $\left(\noisyGa \cdot y_\type + (1-\noisyGa) \cdot y'_\type \right)$ is negligibly close to one of the values in $\Lambda[0,1] = \set{\frac{\lambda}{2},\frac{3\lambda}{2},\dots, 1-\frac{\lambda}{2}}$. 
    Applying \Cref{clm:y_and_y'_are_far}, we get
    \[
    |\gamma y + (1-\gamma)y' - \noisyGa y + (1-\noisyGa)y'| = 
    |(\gamma - \noisyGa) y + (\noisyGa - \gamma) y'| = |\nu| \cdot ||y-y'|| \ge |\nu| \cdot \frac{\eps}{M}.
    \] 
    Since $\nu \sim \mathrm{Uniform} \left[ -\frac{M}{\eps \kappa}, \frac{M}{\eps \kappa}\right]$, the probability of the event occurring is at most $\frac{\negli(\kappa)}{2 / \kappa} = \negli(\kappa)$. The claim follows.
    \end{proof}

    Returning to \Cref{eq:calibration_of_R_tilde}, we established that
    \[
    \left| 
    \left(\gamma_{\type,\beta} \cdot y_\type + (1-\gamma_{\type,\beta}) \cdot y'_\type \right) - 
    \Pr\left[\tilde{\type}(x) = \type \mid x \in \X_{\type, \beta} \right] \right| < \frac{\lambda}{2}.
    \]
    With overwhelming probability, this implies that $\Pr\left[\tilde{\type}(x) = \type \mid x \in \X_{\type, \beta} \right]$ remains within $\frac{\lambda}{2}$ of $\beta^*_\type$ for any $\beta$, ensuring that $\Pr\left[\tilde{\type}(x) = \type \mid x \in \X_{\type, \beta} \right] \in \lambda(\beta^*_\type)$. Consequently, $\X_{\type, \beta}$ is independent of $\beta$, completing the proof of the lemma.
    \end{proof}

Applying \Cref{lemma:only_one_set_is_large}, we conclude that for each $\type \in \setOfTypes$, there exists a unique set $\X_{\type, \beta^*_\type}$ such that $\left|\X_{\type,\beta^*_\type}\right| > \frac{2\alpha}{\lambda} |\X|$. Since the level sets for each $\type$ are disjoint, it follows that 
\[
\left|\X_{\type, \beta^*_\type}\right| \ge \left(1- \frac{2\alpha}{\lambda^2} \right)|\X|.
\]
Defining 
\[
\tilX = \bigcap_{\type \in \setOfTypes} \X_{\type, \beta^*_\type},
\]
we obtain 
\begin{equation}\label{eq:size_of_X_tilde}
\left| \tilX \right| \ge \left(1- \frac{2k\alpha}{\lambda^2}\right)|\X|.
\end{equation}
Now, let
\[
\tilGa \defeq \frac{\left|\X_1 \cap \tilX\right|}{\left|\tilX\right|}
 \text{ and } 
\tilde{y} \defeq \expectation_{x \sim \DX}\left[ \tilde{R}(x) \mid x \in \tilX \right],
\]
where the expectation is taken coordinate-wise (i.e., $\tilde{y} \in \stochasticVecs$).
Following the same decomposition as in \Cref{eq:decompose_R_star}, we have
\[
\Pr\left[\type^*(x) = \type \land x \in \tilX \right] = \tilGa \cdot y_\type + (1-\tilGa) \cdot y_\type'.
\]
By \Cref{clm:accuracy_holds_on_a_large_subset}, $\tilde{R}$ is $\alpha'$-accurate w.r.t. $\mD$ on $\tilX$, where $\alpha' = \alpha + \frac{2k\alpha}{\lambda^2} \le \frac{3k\alpha}{\lambda^2}$. That is,
\[
\left\|\tilde{y} - \left(\tilGa \cdot y + (1-\tilGa) \cdot y'\right) \right\|_{\infty} \le \frac{3k\alpha}{\lambda^2} \cdot \frac{1}{\Pr[x \in \tilX]} \le \frac{3k\alpha}{\lambda^2 - 2k\alpha}.
\]
By the definition of $\tilX$, for any $x \in \tilX$,
\[
\left\|\tilde{y} - \tilde{R}(x)\right\|_\infty \le \lambda.
\]
Thus, for any $x \in \tilX$,
\[
\left\|\tilde{R}(x) - \left(\tilGa \cdot y + (1-\tilGa) \cdot y'\right) \right\|_{\infty} \le \frac{3k\alpha}{\lambda^2 - 2k\alpha} + \lambda.
\]
Since $\decRule$ is $M$-Lipschitz, it follows that for any $x \in \tilX$,
\begin{equation}\label{eq:lip_ex}
\left| \expectation_\decRule\left[\decRule\left(\tilde{R}(x)\right)\right] - \expectation_\decRule\left[\decRule\left(\tilGa \cdot y + (1-\tilGa) \cdot y'\right)\right] \right| \le \frac{3Mk\alpha}{\lambda^2 - 2k\alpha} + \lambda M.
\end{equation}
In particular, taking the expectation over all $x \in \tilX$ gives:
\begin{equation}\label{eq:rule_is_far_on_y_tilde_and_sq_R_star_on_X_tilde}
\left| \expectation_{x,\decRule}\left[\decRule\left(\tilde{R}(x)\right)\right] - \expectation_\decRule\left[\decRule\left(\tilGa \cdot y + (1-\tilGa) \cdot y'\right)\right] \right| \le \frac{3Mk\alpha}{\lambda^2 - 2k\alpha} + \lambda M.
\end{equation}

Recall the assumption that $\decRule$ is $\eps$-far from affine:
\begin{equation}\label{eq:assumption_distance_from_affine}
\left| \expectation\left[\decRule\left(\gamma \cdot y + (1-\gamma) \cdot y'\right)\right] - \left( \gamma \cdot \expectation\left[\decRule(y)\right] + (1-\gamma) \cdot \expectation\left[\decRule(y')\right] \right) \right| > \eps.
\end{equation}
The following claim shows that this also applies to $\tilX$.

\begin{claim}
$\left| \expectation\left[\decRule\left( \tilGa \cdot y + (1-\tilGa) \cdot y' \right)\right] - \left(\tilGa \cdot \expectation\left[\decRule(y)\right] + (1-\tilGa) \cdot \expectation\left[\decRule(y')\right] \right) \right| \ge \eps - \frac{2M^2}{\eps \kappa}$.
\end{claim}
\begin{proof}
Since the $\ell_\infty$ norm of stochastic vectors is at most $1$,
\[
\left\|\left( \gamma \cdot y + (1-\gamma) \cdot y' \right) - \left(\tilGa \cdot y + (1-\tilGa) \cdot y' \right)\right\|_\infty \le |\gamma - \tilGa| \cdot \| y - y' \|_\infty \le |\gamma - \tilGa|.
\]
Since $\decRule$ is $M$-Lipschitz, it follows that
\[
\left| \expectation\left[\decRule\left( \gamma \cdot y + (1-\gamma) \cdot y' \right) \right] - \expectation\left[\decRule\left(\tilGa \cdot y + (1-\tilGa) \cdot y' \right)\right] \right| \le M |\gamma - \tilGa|.
\]
Additionally, since $|\expectation\left[\decRule(y) - \decRule(y')\right]| \le 1$,
\[
\left|\left( \gamma \cdot \expectation\left[\decRule(y)\right] + (1-\gamma) \cdot \expectation\left[\decRule(y')\right] \right)
- \left(\tilGa \cdot \expectation\left[\decRule(y)\right] + (1-\tilGa) \cdot \expectation\left[\decRule(y')\right] \right) \right| \le |\gamma - \tilGa|.
\]
Combining these bounds with the assumption in \Cref{eq:assumption_distance_from_affine} gives
\[
\left| \expectation\left[\decRule\left( \tilGa \cdot y + (1-\tilGa) \cdot y' \right)\right] - \left(\tilGa \cdot \expectation\left[\decRule(y)\right] + (1-\tilGa) \cdot \expectation\left[\decRule(y')\right] \right) \right| \ge \eps - (M+1)|\gamma - \tilGa|.
\]
By \Cref{clm:subsets_maintain_gamma_fraction}, we have 
$\left|\gamma' - \noisyGa \right| = \negli(\kappa)$ and $\left|\tilGa - \gamma' \right| = \negli(\kappa)$. Since $\left|\gamma - \noisyGa \right| = |\nu| \le \frac{M}{\eps \kappa}$, it follows that
\begin{equation}\label{eq:gamma_and_gamma_til_close}
|\gamma - \tilGa| \le |\gamma - \noisyGa| + |\noisyGa - \gamma'| + |\gamma' - \tilGa| \le \negli(\kappa) + \negli(\kappa) + \frac{M}{\eps \kappa},
\end{equation}
and the claim follows.
\end{proof}
Returning to \Cref{eq:rule_is_far_on_y_tilde_and_sq_R_star_on_X_tilde}, the claim implies that
\[
\left| \expectation_{x,\decRule}\left[\decRule\left(\tilde{R}(x)\right)\right] - \left(\tilGa \cdot \expectation_{\decRule}\left[\decRule(y)\right] + (1-\tilGa) \cdot \expectation_{\decRule}\left[\decRule(y')\right] \right) \right| 
\ge \eps - \frac{2M^2}{\eps \kappa} - \frac{3Mk\alpha}{\lambda^2 - 2k\alpha} - \lambda M 
> \frac{3\eps}{4},
\]
by taking $\kappa \ge \frac{30M^2}{\eps^2}$, $\alpha \le \frac{\lambda^2 \eps}{60Mk}$, and $\lambda \le \frac{\eps}{16M}$.

Thus, we have established a lower bound on the accuracy-on-decision (AD) error of $\tilde{R}$
with respect to $\mD$ and $\decRule$ on $\tilX$. To translate this into a lower bound over all
of $\X$, we recall from \Cref{eq:size_of_X_tilde} that 
\[
    \eta \defeq \Pr_{x\sim \UX}[x\notin \tilX] \le \frac{2k\alpha}{\lambda^2}.
\]
By our choice of parameters, $\eta \le \eps/16$, and the lower bound above is at least
$3\eps/4$. Since both $\decRule(\tilde R(x))$ and $\decRule(R^*(x))$ lie in $[0,1]$, the
contribution of $\X\setminus \tilX$ to the AD gap can change the expectation by at most $\eta$.
Therefore, the AD error over $\X$ is at least
\[
    (1-\eta)\cdot \frac{3\eps}{4} - \eta > \frac{\eps}{2},
\]
contradicting the assumption that $\tilde{R}$ is $\eps/2$-AD with respect to $\mD$ and
$\decRule$.
\end{proof}

The following remarks highlight two key technical choices in the proof strategy. They clarify how the argument remains valid for any choice of discretization parameter, and why we adopt a Lipschitz assumption over weaker continuity conditions.

\begin{remark}[Avoiding the randomization of $\gamma$]\label{rem:gamma_rand}
Getting back to \Cref{lemma:only_one_set_is_large}, our goal was to find a large subset $\tilX$ where $\tilde{R}$ is close (in $\ell_\infty$ norm) to the expected value of $R^*$ on $\X$. To achieve this, we prove that for any label $\type$, at most one of the $1/\lambda$ level sets of $\type$ can be large, and we define $\tilX$ as the intersection of these large level sets. However, the $\type\tth$ coordinate of the expected value of $R^*$ over each level set $\X_{\type,\beta}$ is not exactly $\left(\gamma' \cdot y_\type + (1-\gamma') \cdot y'_\type\right)$, but rather $\left(\gamma_{\type,\beta} \cdot y_\type + (1-\gamma_{\type,\beta}) \cdot y'_\type\right)$, where $\gamma_{\type,\beta}$ is negligibly close to $\gamma'$. Nevertheless, they may still fall into different $\lambda$-intervals. 

To address this, we introduce a small random perturbation to $\gamma$, which propagates to $\gamma'$ and $\gamma_{\type,\beta}$, ensuring that the issue arises with only negligible probability, as shown in \Cref{clm:discretization_is_good_whp}. 

An alternative approach is to define $\lambda$ accordingly to prevent this issue. By assuming a uniform bound on the negligible term controlling this difference, we can set $\lambda$ so that all values remain within the same $\lambda$-interval. This bound can be achieved, for instance, by restricting the learner that outputs $\tilde{R}$ to run within a fixed polynomial time. This approach would eliminate the need for the randomization introduced in \Cref{eq:noisy_gamma} and significantly reduce the technical overhead in \Cref{lemma:only_one_set_is_large}. 

However, any method that merely proves the existence of a suitable $\lambda$, even without additional assumptions, is weaker than our approach, which guarantees that the proof holds for {\em any} $\lambda$.
\end{remark}

\begin{remark}[Relaxing the Lipschitz condition]\label{rem:relax_the_lip}
    The theorem applies to Lipschitz rules that are far from affine. One may ask if requiring continuity is enough. Indeed, since $\stochasticVecs$ is a compact set, continuity on $\stochasticVecs$ implies uniform continuity. Let us revisit, for example, \Cref{eq:lip_ex}, and denote the two vectors there by $z$ and $z'$. Using uniform continuity, for any desired bound $\mu$, we know that there exists $\delta > 0$ such that if $||y - y'|| < \delta$, then $\expectation[\decRule(z)] - \expectation[\decRule(z')] < \mu$. However, there is no guarantee of how small $\delta$ should be. In particular, it can be inverse exponential in $\eps$, and lead to trivial bounds on $\alpha$ and $\lambda$, such that the result only holds in the regime where, e.g., they are both smaller than $2^{-\eps}$. To obtain meaningful bounds, we introduce the Lipschitzness requirement.
\end{remark}

\subsection{Loss-minimizing decision rule is far from affine}\label{sec:loss_mini_far_from_aff}
Returning to \Cref{def:loss_opti_dec_rule}, we now prove that the loss-minimizing decision rule $\decRule^*_\loss$ is far from affine for any non-trivial loss function. In \Cref{rem:loss_min_dec_rule_not_affine_as_cor}, we will show how this statement can also be derived from the arguments presented earlier.

The class of loss functions we consider as non-trivial includes all loss functions that do not trivialize the tasks of loss minimization or being accurate-on-classification by making them either too easy or impossible.\footnote{To see why the second condition in \Cref{def:loss_family} ensures that the task of being multi-accurate-on-classification is not trivialized, recall that the loss-minimizing decision rule (\Cref{def:loss_opti_dec_rule}) breaks ties in favor of rejection. Hence, if $\loss$ violates the second condition, then the loss-minimizing action function is constant: either $\actionFunction^*_\loss \equiv 0$ or $\actionFunction^*_\loss \equiv 1$.}

\begin{definition}[Family of Non-Trivial Loss Functions $\lossFamily_\alpha$]\label{def:loss_family} 
    The family of non-trivial loss functions $\lossFamily_\alpha \subseteq \{\setOfTypes \times \zo \to [0,1] \}$ is defined such that any $\loss \in \lossFamily_\alpha$ satisfies the following two conditions:
    \begin{enumerate}
    \item there exists $\type \in \setOfTypes$ such that $|\loss(\type,1) - \loss(\type,0)| \ge \alpha$.
    \item there exist $\type', \type'' \in \setOfTypes$ such that $\loss(\type',1) - \loss(\type',0) \ge 0$ and $\loss(\type'',1) - \loss(\type'',0) < 0$.
    \end{enumerate}
\end{definition}

\begin{lemma}[Loss-Minimizing Decision Rule is Far from Affine]\label{lem:loss_min_dec_rule_not_affine}
    The decision rule $\decRule^*_\loss$ is $1/2$-far from affine for any non-trivial $\loss \in \lossFamily_\alpha$.
\end{lemma}
\begin{proof}
Since $\loss$ is non-trivial (\Cref{def:loss_family}), there exists $\type, \type' \in \setOfTypes$ such that $\loss(\type,1) - \loss(\type,0) \ge 0$ and $\loss(\type',1) - \loss(\type',0) < 0$ (these are $\type',\type''$ from the second condition). 
Consider the unit vectors $e_\type$ and $e_{\type'}$, where each vector has a 1 at coordinate $\type$ (or $\type'$, respectively) and 0 elsewhere. By definition, the loss-minimizing decision rule satisfies
\[
\decRule^*_\loss(e_\type) = 0 \text{ and } \decRule^*_\loss(e_{\type'}) = 1.
\]
Now, consider their convex combination:
\[
v = \frac{1}{2}e_\type + \frac{1}{2}e_{\type'}.
\]
On one hand, the convex combination of their decision rule values is
\[
\frac{1}{2} \decRule^*_\loss(e_\type) + \frac{1}{2} \decRule^*_\loss(e_{\type'}) = \frac{1}{2}.
\]
On the other hand, $\decRule^*_\loss(v) \in \zo$ by its definition --- it is a deterministic decision rule that outputs a bit for each fixed vector. This means that 
\[
\left|  \decRule^*_\loss(v)  - \left( \frac{1}{2} \decRule^*_\loss(e_\type) + \frac{1}{2} \decRule^*_\loss(e_{\type'}) \right) \right| = \frac{1}{2},
\]
so $\decRule^*_\loss$ is at least $1/2$-far from affine.
\end{proof}

Although \Cref{lem:loss_min_dec_rule_not_affine} establishes that the loss-minimizing decision rule $\decRule^*_\loss$ is far from affine, it is not Lipschitz, and therefore \Cref{thm:impossibility_result} does not apply to it directly.
The following claim extends the impossibility to $\decRule^*_\loss$ by applying \Cref{thm:impossibility_result} to a Lipschitz smoothing of this rule and showing that the resulting lower bound transfers to the original, non-Lipschitz rule.

\begin{claim} \label{clm:loss_min_dec_rule_impossibility}
For any non-trivial loss $\loss \in \lossFamily_\alpha$, the conclusion of \Cref{thm:impossibility_result} extends to $\decRule^*_\loss$ and gives accuracy-on-decision error of $1/8$. 
\end{claim}
\begin{proof}[Proof sketch.]
First suppose that the expected loss difference is strictly positive at some point and strictly negative at another. By linearity, we may choose a smoothing parameter $\eta$ satisfying $\eta \le \alpha / 8$ where $\alpha$ is the non-triviality parameter, and $y_0,y_s\in\stochasticVecs$ such that
\[
\expectation_{\type\sim y_0}[\loss(\type,1)-\loss(\type,0)]=-2\eta,
\qquad
\expectation_{\type\sim y_s}[\loss(\type,1)-\loss(\type,0)]=16\eta,
\]
or with both signs reversed. Let $\bar{\decRule}_{\loss,\eta}$ be the symmetric smoothing of $\decRule^*_\loss$: its acceptance probability is $1$ below $-\eta$, $0$ above $\eta$, and linear in between, that is, 
\[
\Pr[\bar{\decRule}_{\loss,\eta}(y)=1] =
\begin{cases}
1
&
\text{if }
\expectation_{\type\sim y}[\loss(\type,1)-\loss(\type,0)]
\le -\eta,
\\
\frac{
\eta-\expectation_{\type\sim y}[\loss(\type,1)-\loss(\type,0)]
}{2\eta}
&
\text{if }
\left|
\expectation_{\type\sim y}[\loss(\type,1)-\loss(\type,0)]
\right|
<\eta,
\\
0
&
\text{if }
\expectation_{\type\sim y}[\loss(\type,1)-\loss(\type,0)]
\ge \eta.
\end{cases}
\]
This decision rule is
$k/(2\eta)$-Lipschitz (see \Cref{def:lip_rule}). For the linear combination
\[
z \defeq \frac{3}{4}y_0+\frac{1}{4}y_s,
\]
the loss difference equals $5\eta/2$ (or $-5\eta/2$), and the acceptance probabilities of $\bar{\decRule}_{\loss,\eta}$ at $(y_0,y_s,z)$ are respectively $(1,0,0)$ or $(0,1,1)$. Thus, $\bar{\decRule}_{\loss,\eta}$ is $1/2$-far from affine (or any other constant strictly smaller than $3/4$). Note that comparing to \Cref{thm:impossibility_result}, $y_0$ and $y_s$ are analogous to $y$ and $y'$ respectively, and $\gamma$ is analogous to $3/4$. 

Apply \Cref{thm:impossibility_result} with $\eps=1/2$. The constructed $R^*$ takes only the values $y_0,y_s$ on which $\bar{\decRule}_{\loss,\eta}$ and $\decRule^*_\loss$ agree. Moreover, the bound preceding \Cref{eq:lip_ex}, together with $\tilGa$ being close to $3/4$ by \Cref{eq:gamma_and_gamma_til_close}, implies that, for every discretization parameter and calibration error satisfying the explicit hypotheses of \Cref{thm:impossibility_result}, the expected loss difference at $\tilde R(x)$ remains outside the smoothing interval $[-\eta,\eta]$ for every $x\in\tilX$. Hence, the two rules also agree on $\tilde R(x)$ throughout $\tilX$. Taking $\Pr[x\notin\tilX]\le1/32$, the $1/4$ lower bound for $\bar{\decRule}_{\loss,\eta}$ yields an accuracy-on-decision error greater than $1/4-1/32>1/8$ for $\decRule^*_\loss$.

Finally, if the only available non-negative point $y_0$ is a tie, the symmetric smoothing disagrees with $\decRule^*_\loss$ at $y_0$. In this case, use the one-sided smoothing whose acceptance probability is $1$ below $-\eta$, $0$ from $0$ onward, and linear on $[-\eta,0]$. Choose $y_s$ with loss difference at most $-8\eta$. Then the loss difference at $\frac34y_0+\frac14y_s$ is at most $-2\eta$, the acceptance probabilities are $(0,1,1)$, and the same argument applies.
\end{proof}

\section{When loss minimization contradicts accuracy of classifications} \label{sec:impossibility_AC_and_loss_min}

As we have argued, achieving accuracy-on-decision may sometimes be impossible, implying that accuracy-on-classification might not be attainable when relying on predictors. However, this does not preclude the possibility of classifying individuals directly---classification methods need not operate by applying a decision rule to a risk predictor. In this section, we show that loss minimization may not align with accuracy-on-classification, even for general action functions. Specifically, for any non-trivial loss function, we construct a setting in which any (possibly randomized) action function that satisfies accuracy-on-classification necessarily fails to minimize the expected loss. This holds even when the hypothesis class is restricted to just the two constant functions that either accept or reject each individual with probability $1$.  

The following theorem establishes the central argument of this section: achieving both optimal loss minimization and accurate representation in classification---even for a single subpopulation---may be computationally infeasible. However, these objectives can be attained independently. In \Cref{cor:MA_implies_MAC}, we show that a multi-accurate predictor enables MAC, while in \Cref{thm:MC_for_loss_mini}, we demonstrate that loss minimization is possible given a multi-calibrated predictor. This confirms that while loss minimization and MAC can each be achieved separately---and even with the same predictor---they may not necessarily be achieved by applying the same decision rule to that predictor.

\begin{theorem}[Loss Minimization Contradicts Accuracy-on-Classification] \label{thm:impossibility_result_of_loss}
Assume one-way functions exist w.r.t.\ a security parameter $\kappa \ge \log|\X|$. Let $\eps, \alpha \in (0,1)$, and let $\actionFunctionsClass = \{\actionFunction_0,\actionFunction_1\}$ where $\actionFunction_0 \equiv 0$ and $\actionFunction_1 \equiv 1$.
For any $\loss \in \lossFamily_\alpha$, there exists a Nature $\mD$ such that for any $\poly(\kappa)$-time computable $\actionFunction: \X \to \zo$, if $\actionFunction$ is $(\loss,\eps)$-accurate-on-classification then $\actionFunction$ incurs loss-minimization error at least $\Omega(\alpha) - O(\eps)$ with respect to $\loss$ and $\actionFunctionsClass$.
\end{theorem}

As discussed in \Cref{sec:indistin_subsets} and before \Cref{thm:impossibility_result}, the assumption of one-way functions is a modeling choice and can be removed if no restriction is placed on the complexity of Nature.

\begin{proof}
Recall from \Cref{def:loss_family} that $\lossFamily_\alpha$ denotes the family of non-trivial loss functions. Let $\loss \in \lossFamily_\alpha$, and let $\UX$ be the uniform distribution over $\X$. Define two constants $\ACEps, \lossEps \in (0,1)$ to be specified later.
Since $\loss$ is non-trivial, there exist types $\type, \type', \type'' \in \setOfTypes$ such that
\[
\left| \loss(\type,1) - \loss(\type,0) \right| \ge \alpha, \quad
\loss(\type',1) - \loss(\type',0) \ge 0, \quad
\loss(\type'',1) - \loss(\type'',0) < 0.
\]
Depending on the sign of $\loss(\type,1) - \loss(\type,0)$, we define $\beta = \loss(\type',1) - \loss(\type',0)$ (when the sign is negative) or $\beta = \loss(\type'',0) - \loss(\type'',1)$ (when the sign is positive), so that $\beta \ge 0$ or $\beta > 0$, respectively.
The comparison between $\alpha$ and $\beta$ induces four analogous cases. We focus on the case where
\[
\loss(\type,1) - \loss(\type,0) \ge \alpha, \quad
\loss(\type',0) - \loss(\type',1) = \beta > 0,\footnote{Here we rename $\type''$ as $\type'$ for notational simplicity.} \quad
\alpha \ge \beta.
\]
In this case, rejecting an individual of type $\type$ results in a significantly lower loss than accepting them, while for type $\type'$, acceptance yields lower loss than rejection, though the gap may be smaller.
The remaining cases follow analogously by interchanging the roles of $\type$ and $\type'$ or $\type''$, and yield the same lower bound.

We define Nature according to the pair of labels $\type$ and $\type'$. Under the assumption that one-way functions exist, \Cref{lem:existence_of_X1} implies the existence of a subset $\X_1 \subseteq \X$ of size  
\[
|\X_1| = \left(\frac{3}{4} + \negli(\kappa)\right) |\X|,
\]
such that $\X_1$ is $\kappa$-indistinguishable w.r.t. $\UX$ (see \Cref{def:indistin}) and Nature is efficiently computable.

Defining $\X_2 \defeq \X \setminus \X_1$, we write  
\[
\X = \X_1 \cupdot \X_2.
\]
We specify Nature's deterministic prediction as follows:
\[
R^*|_{\X_1} = \type, \quad R^*|_{\X_2} = \type',
\]
and take $\mD = \UX \times R^*(\UX)$.\footnote{So far, we defined a deterministic Nature such that it outputs unit vectors $e_\type, e_{\type'} \in \stochasticVecs$, rather than labels $\type, \type' \in \setOfTypes$. For notational convenience, we now define it to output a label in this case, and denote it by $R^*$.
}

Let $\actionFunction$ be a $\poly(\kappa)$-time computable randomized action function that is $(\loss,\ACEps)$-accurate-on-classification. By our definition of Nature,  
\[
\actionFunction^*_\loss(x) =
\begin{cases}
0, & x \in \X_1, \\
1, & x \in \X_2.
\end{cases}
\]
Therefore,  
\[
\expectation\left[ \actionFunction^*_\loss(x) \right] = \Pr[x \in \X_2] = \frac{1}{4} + \negli(\kappa).
\]
By $(\loss,\ACEps)$-accuracy-on-classification (see \Cref{def:MAC}) and since $\expectation\left[ \actionFunction(x) \right] = \Pr[\actionFunction(x) = 1]$, it follows that
\begin{equation}\label{eq:MAC_cond}  
\Pr_{x,\actionFunction}[\actionFunction(x) = 1] \ge \frac{1}{4} - \ACEps + \negli(\kappa).
\end{equation}
In particular, this implies that $\Pr[\actionFunction(x) = 1]$ is non-negligible. Since $\actionFunction$ is $\poly(\kappa)$-time computable and $\X_1$ is $\kappa$-indistinguishable, \Cref{clm:subsets_maintain_gamma_fraction} applies, yielding  
\begin{equation}\label{eq:indistin_cond}
    \Pr\left[ x \in \X_1 \middle| \actionFunction(x) =1 \right] = \frac{3}{4} + \negli(\kappa),
\end{equation}
and, analogously,  
\[
\Pr\left[ x \in \X_2 \middle| \actionFunction(x) =1 \right] = \frac{1}{4} + \negli(\kappa).
\]
Finally, we proceed to compute the expected loss of the loss minimizer $\actionFunction_0$ and of $\actionFunction$. 
    Notice that $\actionFunction_0 \in \actionFunctionsClass$, and its expected loss is
\[
\expectation_{x} \left[\loss\left(R^*(x), \actionFunction_0(x)\right) \right] = \expectation \left[\loss\left(R^*(x), 0\right) \right] = 
\frac{3}{4} \cdot \loss(\type,0) + \frac{1}{4}  \cdot \loss(\type',0) \defeq L_0.
\]
Similarly, the expected loss of $\actionFunction_1$ is 
\[
\expectation_{x} \left[\loss\left(R^*(x), \actionFunction_1(x)\right) \right] = \expectation \left[\loss\left(R^*(x), 1\right) \right] = 
\frac{3}{4} \cdot \loss(\type,1) + \frac{1}{4}  \cdot \loss(\type',1) \defeq L_1.
\]
This implies that
\begin{equation}\label{eq:diff_L0_L1}
L_1 - L_0 = \frac{3}{4} \left( \loss(\type,1) - \loss(\type,0)\right) - \frac{1}{4} \left( \loss(\type',0) - \loss(\type',1)\right) \ge \frac{3}{4} \alpha - \frac{1}{4} \beta \ge \frac{\alpha}{2},
\end{equation}
where the last inequality follows from the assumption that $\alpha \ge \beta$.

Next, we compute the expected loss of $\actionFunction$ over $\UX$:
\begin{align}\label{eq:loss_of_h}
     \expectation_{x,\actionFunction} \left[\loss\left(R^*(x), \actionFunction(x)\right) \right] \nonumber 
    &= \expectation\left[\loss\left(R^*(x), 0\right) \middle| \actionFunction(x) = 0 \right] \cdot \Pr[\actionFunction(x) = 0] \nonumber \\
    & + \expectation\left[\loss\left(R^*(x), 1\right) \middle| \actionFunction(x) = 1 \right] \cdot \Pr[\actionFunction(x) = 1] \nonumber \\ 
    &\ge \expectation\left[\loss\left(R^*(x), 0\right) \middle| \actionFunction(x) = 0 \right] \cdot \left(\frac{3}{4} - \ACEps \right) \nonumber \\
    & + \expectation\left[\loss\left(R^*(x), 1\right) \middle| \actionFunction(x) = 1 \right] \cdot \left(\frac{1}{4} - \ACEps \right) + \negli(\kappa),
\end{align}
where the last inequality follows from \Cref{eq:MAC_cond}.

We argue that this quantity is close to $\frac{3}{4}L_0 + \frac{1}{4}L_1$. First, observe that
\begin{align*} 
\expectation\left[\loss\left(R^*(x), 1\right) \middle| \actionFunction(x) = 1 \right] &= \loss\left(\type, 1\right) \cdot \Pr\left[x \in \X_1 \mid \actionFunction(x) = 1 \right] \\
&+ \loss\left(\type', 1\right) \cdot \Pr\left[x \in \X_2 \mid \actionFunction(x) = 1 \right] \\ 
&= \frac{3}{4} \cdot \loss\left(\type, 1\right)  + \frac{1}{4} \cdot \loss\left(\type', 1\right) + \negli(\kappa) = L_1 + \negli(\kappa),
\end{align*}
where the second-to-last equality follows from \Cref{eq:indistin_cond}. Similarly, we obtain
\[
\expectation\left[\loss\left(R^*(x), 0\right) \middle| \actionFunction(x) = 0 \right] = \frac{3}{4} \cdot \loss\left(\type, 0\right)  + \frac{1}{4} \cdot \loss\left(\type', 0\right) + \negli(\kappa) = L_0 + \negli(\kappa).
\]
Substituting into \Cref{eq:loss_of_h}, we conclude that
\[
\expectation \left[\loss\left(R^*(x), \actionFunction(x)\right) \right] \ge  \left(\frac{3}{4} - \ACEps \right) \cdot L_0  + \left(\frac{1}{4} - \ACEps \right) \cdot L_1 + \negli(\kappa).
\]
By $(\actionFunctionsClass,\loss,\lossEps)$-loss-minimization of $\actionFunction$ (see \Cref{def:loss_minimization}), we have
\begin{align*}
    \lossEps &\ge \expectation_{x,\actionFunction} \left[\loss\left(R^*(x), \actionFunction(x)\right) \right] - \expectation_{x} \left[\loss\left(R^*(x), \actionFunction_0(x)\right) \right]  \\ 
    &\ge \left(\frac{3}{4} - \ACEps \right) \cdot L_0  + \left(\frac{1}{4} - \ACEps \right) \cdot L_1 + \negli(\kappa) - L_0 \\ 
    &= \frac{1}{4} \left(L_1 - L_0 \right) - \ACEps \left(L_1 + L_0 \right) + \negli(\kappa) \\ 
    &\ge \frac{\alpha}{8} - \ACEps \left(L_1 + L_0 \right) + \negli(\kappa),
\end{align*}
where the last inequality follows from \Cref{eq:diff_L0_L1}.
Since $L_1 + L_0 \le 2 - \frac{3}{4}\alpha$, we conclude that
\[
\lossEps \ge \frac{\alpha}{8} - \ACEps \left(2 - \frac{3}{4}\alpha \right) + \negli(\kappa).
\]
The theorem follows by setting $\eps = \ACEps$, but the analysis allows for a more general tradeoff between $\ACEps$ and $\lossEps$. For instance, requiring the accuracy-on-classification error to be smaller than $\frac{\alpha}{80}$ implies that $(\actionFunctionsClass,\loss,\frac{\alpha}{10})$-loss-minimization is impossible.
\end{proof}

\section{Translating accuracy of predictions to loss minimization}\label{sec:MC_for_loss}
Recall the definitions of the loss-minimizing decision rule (\Cref{def:loss_opti_dec_rule}) and full-multi-calibration (\Cref{def:full_MC_for_types}). The following theorem shows that composing a multi-calibrated predictor with the loss-minimizing decision rule results in a loss-minimizing action function. The hypothesis class with respect to which loss minimization is measured corresponds to the class of subpopulations for which the predictor is multi-calibrated. Notably, this implies that the loss function need not be known during learning; one can first learn a multi-calibrated predictor and later post-process it using the loss-minimizing decision rule to obtain an action function that minimizes any given loss. 

This guarantee aligns with the notion of {\em omnipredictors} introduced by \citet{omni}, which are predictors that admit post-processing to minimize any loss function within a given class. Specifically, Theorem~8.1 in \citet{omni} shows that a multi-calibrated predictor is an omnipredictor. Our proof follows their high-level approach and proof technique, with a few adaptations to fit our setting. First, the class of loss functions we consider, as well as the assumptions imposed on them, differ from those in~\citet{omni}. Second, in \Cref{lem:not_in_omni_paper} we extend the argument by showing that composing the original predictor with a loss-minimizing decision rule yields an action function that minimizes loss, whereas the original paper establishes this property for a different primitive. Moreover, in \Cref{sec:relaxing_the_MC}, we show how to relax the multi-calibration requirement. We note that the notion of multi-calibration used in the following proof is identical to that of~\citet{omni}, though their presentation is phrased in terms of covariance.

\begin{theorem}[Multi-Calibration for Loss Minimization]\label{thm:MC_for_loss_mini}
    Let $\C \subseteq \zo^{|\X|}$ be a collection of sets, and let $\actionFunctionsClass = \actionFunctionsClass_\C \subseteq \{ \X \to \zo \}$ be the hypothesis class consisting of all functions that compute membership in sets from $\C$. 
    Let $\lambda \in (0,1)$ be a discretization parameter, and let $\loss: \setOfTypes \times \zo \to [0,1]$ be a loss function. If $\tilde{R}$ is $(\C,\alpha,\lambda)$-full-multi-calibrated w.r.t. $\mD$, then $\decRule^*_\loss \circ \tilde{R}$ is $\left(\actionFunctionsClass,\loss,3k(\alpha+\lambda) \right)$-loss-minimizing. 
\end{theorem}
\begin{proof}
Throughout the proof, it will be convenient to use the equivalence between two sampling processes: the standard process where we first sample an element from $\X$ and then a label, and the alternative of first sampling a label and then an element.  
Given a distribution $\DX$ over $\X$, each element $x \in \X$ is associated with a stochastic vector $R^*(x) \in \stochasticVecs$, which defines a distribution over the set of types $\setOfTypes$. The {\em marginal label distribution} $\mD_\setOfTypes$ is given by  
\[
\mD_\setOfTypes(\type) = \sum_{x \in \X} \DX(x) R^*(x)_{\type},
\] 
where $R^*(x)_{\type}$ is the $\type\tth$ coordinate of $R^*(x)$. 
The {\em distribution over elements given a label} $\mD_{\X | \setOfTypes}$ is  
\[
\mD_{\X | \setOfTypes}(x \mid \type) = \frac{\DX(x) R^*(x)_{\type}}{\mD_\setOfTypes(\type)}.
\]  
Thus, sampling $(x, \type)$ by first drawing $x \sim \DX$ and then $\type \sim R^*(x)$ is equivalent to first drawing $\type \sim \mD_\setOfTypes$ and then $x \sim \mD_{\X | \setOfTypes}(\cdot \mid \type)$.
For clarity, we now fix the notation used throughout the proof.
\begin{itemize}
    \item In what follows, we define a partition $T = (T_v)_{v \in \left(\Lambda[0,1]\right)^k}$ of $\X$. For sampling a subset $T_v$ from the partition $T$ according to its mass under $\DX$, that is, $\sum_{x \in T_v} \DX(x)$, we use $T_v \sim T$.

    \item We use $x \sim T_v$ to denote sampling $x \sim \DX$ conditioned on the event that $x \in T_v$. 

    \item We use $\type \sim \mD_\setOfTypes|_{T_v}$ to denote sampling from the marginal label distribution restricted to the set $T_v$. 

    \item We use $x \sim \mD_{\X | \setOfTypes = \type, T_v}$ for sampling an element from a set $T_v$ according to the distribution over elements given a label $\type$.

    \item For any $\actionFunction \in \actionFunctionsClass$, we let $\hat{\actionFunction} : T \to \zo$ denote a random variable following a $\text{Ber} \left( \expectation_{h, x \sim T_v} [\actionFunction(x)] \right)$ distribution for any $T_v \in T$.

    \item We restate full multi-calibration in the covariance formulation, following~\citet{omni}. If $\tilde{R}$ is $(\C,\alpha,\lambda)$-full-multi-calibrated with respect to $\mD$, as defined in \Cref{def:full_MC_for_types}, then there exists a partition $T = (T_v)_{v \in \left(\Lambda[0,1]\right)^k}$ of $\X$ such that for any $\actionFunction \in \actionFunctionsClass$, $\type \in \setOfTypes$,
    \begin{equation}\label{eq:cov_MC}
    \expectation_{T_v \sim T} \left[ \Pr_{\type' \sim \mD_{\setOfTypes | T_v}}[\type' = \type] \cdot \left| \expectation_{\actionFunction , x \sim T_v} \left[\actionFunction(x)\right] - \expectation_{\actionFunction , x \sim \mD_{\X \mid \setOfTypes = \type, T_v}} \left[\actionFunction(x)\right] \right| \right] \le \alpha + \lambda.
    \end{equation}

    \item We define the predictor $R^T$ as follows:
    \[\forall x \in T_v,\quad R^{T_v}(x) \defeq \expectation_{x \sim T_v} \left[ R^*(x) \right],
    \] where the expectation is taken coordinate-wise, so that $R^T$ outputs a vector. That is, $R^T$ is constant on each $T_v$ and equals the coordinate-wise expectation of $R^*$ over that part.

    \item We define its corresponding loss-minimizing action function as $\actionFunction^T_\loss \defeq \decRule^*_\loss \circ R^T$.

    \item Following the terminology set in \citet{omni}, we refer to $R^T$ as the {\em canonical predictor} and to $\actionFunction^T_\loss$ as the {\em canonical action function}. 

    \item We denote $\tilde{\actionFunction} \defeq \decRule^*_\loss \circ \tilde{R}$.
\end{itemize}

The proof consists of two steps. First, we show that the canonical action function minimizes the loss. We begin by proving that its loss is at most that of any function that remains constant on each partition set (such as the canonical predictor). We then extend this result to all functions in the hypothesis class. Second, we demonstrate that its expected loss is close to that of $\tilde{\actionFunction}$.

The following claim relates the expected losses of $\actionFunction$ and $\hat{\actionFunction}$ over $T_v$ and a fixed label to their expected values.
\begin{claim}
    Let $\type \in \setOfTypes$ and $T_v \in T$.
    \[
\expectation_{\hat{\actionFunction} , x \sim \mD_{\X | \setOfTypes = \type, T_v}} \left[\loss \left(\type, \hat{\actionFunction}(x) \right)\right] - 
\expectation_{\actionFunction , x \sim \mD_{\X | \setOfTypes = \type, T_v}} \left[\loss \left(\type, \actionFunction(x) \right)\right] 
\le \left|\expectation_{\actionFunction , x \sim T_v} \left[\actionFunction(x) \right] - 
\expectation_{\actionFunction , x \sim \mD_{\X | \setOfTypes = \type, T_v}} \left[\actionFunction(x) \right] \right| .
    \]
\end{claim}
\begin{proof}
    The first term on the left-hand side can be expressed as
    \begin{align*}
    \expectation_{\hat{\actionFunction} , x \sim \mD_{\X | \setOfTypes = \type, T_v}} \left[\loss \left(\type, \hat{\actionFunction}(x) \right)\right] & = 
    \Pr_{\hat{\actionFunction} , x \sim \mD_{\X | \setOfTypes = \type, T_v}} \left[\hat{\actionFunction}(x) = 1 \right] \cdot \loss(\type,1) +  \Pr_{\hat{\actionFunction} , x \sim \mD_{\X | \setOfTypes = \type, T_v}} \left[\hat{\actionFunction}(x) = 0\right]\cdot \loss(\type,0)  
     \\ & =
    \expectation_{\actionFunction , x \sim T_v} \left[\actionFunction(x) \right] \cdot \loss(\type,1) + \left( 1 - \expectation_{\actionFunction , x \sim T_v} \left[\actionFunction(x) \right] \right) \cdot \loss(\type,0),
    \end{align*}
    since $\hat{\actionFunction}$ is fixed on $T_v$.
    The second term on the left-hand side can be expressed as 
    \[
    \expectation_{\actionFunction , x \sim \mD_{\X | \setOfTypes = \type, T_v}} \left[\loss \left(\type, \actionFunction(x) \right)\right] = \expectation_{\actionFunction , x \sim \mD_{\X | \setOfTypes = \type, T_v}} \left[\actionFunction(x) \right] \cdot \loss(\type,1) + \left( 1 - \expectation_{\actionFunction , x \sim \mD_{\X | \setOfTypes = \type, T_v}} \left[\actionFunction(x) \right] \right) \cdot \loss(\type, 0).
    \]
    Hence, their difference is bounded by
    \begin{align*}
     \expectation_{\actionFunction , x \sim T_v} \left[\actionFunction(x) \right] \cdot \loss(\type,1) & + \left( 1 - \expectation_{\actionFunction , x \sim T_v} \left[\actionFunction(x) \right] \right) \cdot \loss(\type,0)  \\ & - \expectation_{\actionFunction , x \sim \mD_{\X | \setOfTypes = \type, T_v}} \left[\actionFunction(x) \right] \cdot \loss(\type,1) - \left( 1 - \expectation_{\actionFunction , x \sim \mD_{\X | \setOfTypes = \type, T_v}} \left[\actionFunction(x) \right] \right) \cdot \loss(\type, 0) 
    \\ & = \left( \loss(\type,1) - \loss(\type,0) \right) \cdot \left( \expectation_{\actionFunction , x \sim T_v} \left[\actionFunction(x) \right] - 
    \expectation_{\actionFunction , x \sim \mD_{\X | \setOfTypes = \type, T_v}} \left[\actionFunction(x) \right] \right)
    \\ & \le \left| \expectation_{\actionFunction , x \sim T_v} \left[\actionFunction(x) \right] - \expectation_{\actionFunction , x \sim \mD_{\X | \setOfTypes = \type, T_v}} \left[\actionFunction(x) \right] \right|.
    \end{align*}
\end{proof}
We use the claim to bound the difference between the expected losses of $\hat{\actionFunction}$ and $\actionFunction$. For any $\actionFunction \in \actionFunctionsClass$, we write this difference as 
\begin{align} \label{eq:loss_diff_from_canonical_is_k_alpha}
& \nonumber \expLoss(\hat{\actionFunction}) - \expLoss(\actionFunction) \le \\ & \nonumber
\expectation_{T_v \sim T} \expectation_{\type \sim \mD_\setOfTypes|_{T_v}} \left| \expectation_{\hat{\actionFunction} , x \sim \mD_{\X | \setOfTypes = \type, T_v}} \left[\loss \left(\type, \hat{\actionFunction}(x) \right)\right] - 
    \expectation_{\actionFunction , x \sim \mD_{\X | \setOfTypes = \type, T_v}} \left[\loss \left(\type, \actionFunction(x) \right)\right]  \right| \le \\ &  \nonumber\expectation_{T_v \sim T} \expectation_{\type \sim \mD_\setOfTypes|_{T_v}} \left| \expectation_{\actionFunction , x \sim T_v} \left[\actionFunction(x) \right] - 
    \expectation_{\actionFunction , x \sim \mD_{\X | \setOfTypes = \type, T_v}} \left[\actionFunction(x) \right] \right| = \\ &\nonumber
    \expectation_{T_v \sim T} \sum_{\type \in \setOfTypes} \Pr_{\type' \sim \mD_\setOfTypes|_{T_v}}[\type' = \type] \left| \expectation_{\actionFunction , x \sim T_v} \left[\actionFunction(x) \right] - 
    \expectation_{\actionFunction , x \sim \mD_{\X | \setOfTypes = \type, T_v}} \left[\actionFunction(x) \right] \right| = \\ &
     \sum_{\type \in \setOfTypes} \expectation_{T_v \sim T} \left[ \Pr_{\type' \sim \mD_\setOfTypes|_{T_v}}[\type' = \type] \left| \expectation_{\actionFunction , x \sim T_v} \left[\actionFunction(x) \right] - 
    \expectation_{\actionFunction , x \sim \mD_{\X | \setOfTypes = \type, T_v}} \left[\actionFunction(x) \right] \right| \right] \le k(\alpha+\lambda),
\end{align}
where the last inequality follows from the $(\C,\alpha,\lambda)$-full-multi-calibration of $\tilde{R}$, as stated in \Cref{eq:cov_MC}.

Next, recall the definitions of the canonical predictor $R^T$ and the canonical action function $\actionFunction^T_\loss$. The following claim achieves our first goal, that is, proving that the canonical action function minimizes the loss. 
\begin{claim}[{Corollary 6.2 in \citet{omni}}]
    $\forall \actionFunction: T \to \zo,\ \expLoss\left(\actionFunction^T_\loss \right) \le \expLoss\left(\actionFunction \right)$.
\end{claim}
\begin{proof}
First, notice that
    \[
    \expLoss\left(\actionFunction^T_\loss \right) = \expectation_{T_v \sim T} \expectation_{x \sim T_v} \expectation_{\type \sim R^{T_v}(x)} \left[ \loss(\type, \actionFunction^T_\loss (x)) \right]. 
    \]    
    A useful fact throughout is that sampling $t\sim R^{T_v}$ gives the same marginal distribution over labels as first sampling $x\sim D_X$ conditioned on $x\in T_v$ and then sampling $t\sim R^*(x)$. This implies that
    \[
    \decRule^*_\loss\left(R^{T_v}(x)\right) = \argmin_{a \in \zo} \expectation_{\type \sim R^{T_v}(x)} \left[ \loss(\type, a) \right]  = \argmin_{a \in \zo} \expectation_{\type \sim R^*|_{T_v}(x)} \left[ \loss(\type, a) \right],
    \]
    hence, 
    \[
    \expectation_{\type \sim R^{T_v}(x)} \left[ \loss(\type, \actionFunction^T_\loss (x)) \right] \le \min \left\{ \expectation_{\type \sim R^*|_{T_v}(x)}[\loss(\type,0)], \expectation_{\type \sim R^*|_{T_v}(x)}[\loss(\type,1)] \right\}.
    \]
Next, let $\actionFunction: T \to \zo$. Once again,
\[
\expLoss\left(\actionFunction \right) = \expectation_{T_v \sim T} \expectation_{x \sim T_v} \expectation_{\actionFunction, \type \sim R^*|_{T_v}(x)} \left[ \loss(\type, \actionFunction(x)) \right].
\]
Focusing on the last term,
\begin{align*}
\expectation_{\actionFunction, \type \sim R^*|_{T_v}(x)} \left[ \loss(\type, \actionFunction(x)) \right] = & \expectation_\actionFunction[\actionFunction] \cdot \expectation_{\type \sim R^*|_{T_v}(x)} \left[ \loss(\type,1) \right] + \left(1-\expectation_\actionFunction[\actionFunction]\right) \cdot \expectation_{\type \sim R^*|_{T_v}(x)} \left[ \loss(\type,0) \right] \\ & \ge \min \left\{ \expectation_{\type \sim R^*|_{T_v}(x)}[\loss(\type,0)], \expectation_{\type \sim R^*|_{T_v}(x)}[\loss(\type,1)] \right\}  \\ & \ge  \expectation_{\type \sim R^{T_v}(x)} \left[ \loss(\type, \actionFunction^T_\loss (x)) \right],
\end{align*}
and the claim follows.
\end{proof}

Recall that $\expLoss(\hat{\actionFunction}) - \expLoss(\actionFunction) \le k(\alpha+\lambda)$ by \Cref{eq:loss_diff_from_canonical_is_k_alpha}. 
Combining it with the claim yields that $\actionFunction^T_\loss$ is $\left(\actionFunctionsClass,\loss,k(\alpha+\lambda) \right)$-loss-minimizing. The following lemma completes the proof, by showing that the difference between $\expLoss\left(\actionFunction^T_\loss\right)$ and $\expLoss\left(\Tilde{\actionFunction}\right)$ is small. 
\begin{lemma}\label{lem:not_in_omni_paper}
    $\expLoss\left(\tilde{\actionFunction}\right) - \expLoss\left(\actionFunction^T_\loss\right) \le 2k(\alpha+\lambda)$.
\end{lemma}
\begin{proof}
    Let us denote 
    \[
    L_1 \defeq \expectation_{\substack{x \sim T_v \\ \type \sim R^*|_{T_v}(x)}} \left[ \loss(\type, 1) \right],
    \]
    and, analogously,
    \[
    L_0 \defeq \expectation_{\substack{x \sim T_v \\ \type \sim R^*|_{T_v}(x)}} \left[ \loss(\type, 0) \right].
    \]
    This means that for any $R$, 
    \[
\expLoss\left((\decRule^*_\loss(R)\right) = \expectation_{x \sim T_v}\left[\decRule^*_\loss(R(x)) \right] \cdot L_1 + \left(1-\expectation_{x \sim T_v}\left[\decRule^*_\loss(R(x)) \right]\right) \cdot L_0,
    \]
    since $\decRule^*_\loss$ is not randomized. Hence,
    \begin{align}\label{eq:bounding_the_exp_loss_diff}
\expLoss(\tilde{\actionFunction}) - \expLoss(\actionFunction^T_\loss) = & \expectation_{x \sim T_v}\left[\decRule^*_\loss(\tilde{R}(x)) \right] \cdot L_1 + \left(1-\expectation_{x \sim T_v}\left[\decRule^*_\loss(\tilde{R}(x)) \right]\right) \cdot L_0  \nonumber \\ & -
    \expectation_{x \sim T_v}\left[\decRule^*_\loss(R^{T_v}(x)) \right] \cdot L_1 + \left(1-\expectation_{x \sim T_v}\left[\decRule^*_\loss(R^{T_v}(x)) \right]\right) \cdot L_0  \nonumber 
    \\ & =
    (L_1-L_0) \left( \expectation\left[\decRule^*_\loss(\tilde{R}(x)) \right] - \expectation\left[\decRule^*_\loss(R^{T_v}(x)) \right] \right) \nonumber  \\ & \le |L_1-L_0| \cdot  \left| \expectation\left[\decRule^*_\loss(\tilde{R}(x)) \right] - \expectation\left[\decRule^*_\loss(R^{T_v}(x)) \right] \right|.
    \end{align}
    We proceed by considering two cases. If $|L_1 - L_0| \le 2k(\alpha+\lambda)$, the claim follows immediately. Otherwise, $|L_1 - L_0| > 2k(\alpha+\lambda)$. Let us assume, without loss of generality, that $L_0 - L_1 > 2k(\alpha+\lambda)$. First, notice that 
    \begin{align}\label{eq:difference_from_L0_L1}
    & \left| \expectation_{\substack{x \sim T_v \\ \type^*(x) \sim R^*(x)}}\left[\loss(\type^*(x), 0) - \loss(\type^*(x), 1)\right] - \expectation_{\substack{x \sim T_v \\ \tilde{\type}(x) \sim \tilde{R}(x)}}\left[\loss(\tilde{\type}(x), 0) - \loss(\tilde{\type}(x), 1)\right] \right| = \\ & \nonumber \left| \expectation_{x \sim T_v} \left[ \sum_{\type \in \setOfTypes} \left( \Pr[\type^*(x) = \type] - \Pr[\tilde{\type}(x) = \type] \right) \cdot \left( \loss(\type, 0) - \loss(\type,1) \right) \right] \right| \le \\ & \nonumber \expectation_{x \sim T_v} \sum_{\type \in \setOfTypes} \left| \Pr[\type^*(x) = \type] - \Pr[\tilde{\type}(x) = \type] \right| \cdot \left| \loss(\type, 0) - \loss(\type,1) \right| \le 2k\alpha,
    \end{align}
    by $(\C,\alpha,\lambda)$-full-multi-calibration of $\tilde{R}$.
    Notice that the first term in \Cref{eq:difference_from_L0_L1} is exactly $L_0-L_1$, hence we conclude that 
    \[
    \mu \defeq \expectation_{\substack{x \sim T_v \\ \tilde{\type}(x) \sim \tilde{R}(x)}}\left[\loss(\tilde{\type}(x), 0) - \loss(\tilde{\type}(x), 1)\right] \ge L_0-L_1 - 2k\alpha \ge 2k\lambda.
    \]
    Next, we argue that this leads to $\decRule^*_\loss(\tilde{R}) = 1$. Specifically, observe that for any $x, x' \in T_v$, we have $||\tilde{R}(x) - \tilde{R}(x')||_\infty \leq \lambda$ by full-multi-calibration.
    Consequently, this implies that
    \[
    \forall x \sim T_v, \left|\expectation_{\tilde{\type}(x) \sim \tilde{R}(x)}\left[\loss(\tilde{\type}(x), 0) - \loss(\tilde{\type}(x), 1)\right] - \mu \right| \le k \lambda,
    \]
Namely, for each $x \sim T_v$, the expected difference between the loss when rejecting and when accepting cannot deviate by more than $k \lambda$ from its expectation over all of $T_v$. Since $\mu \ge 2k\lambda$, it follows that  
\[
\expectation_{\tilde{r}(x) \sim \tilde{R}(x)}\left[\loss(\tilde{r}(x), 0) - \loss(\tilde{r}(x), 1) \right] > 0
\]  
for any $x \sim T_v$. Consequently, we obtain 
\[
\expectation_{x \sim T_v}\left[\decRule^*_\loss(\tilde{R}(x)) \right] = 1.
\]
On the other hand, by definition of $R^{T_v}$, sampling $\type\sim R^{T_v}$ gives the same marginal distribution as first sampling $x\sim D_X$ conditioned on $x\in T_v$ and then sampling $\type\sim R^*(x)$. Since $L_0>L_1$, it follows that
\[
\expectation_{x \sim T_v}\left[\decRule^*_\loss(R^{T_v}(x)) \right] = 1.
\]  
Returning to \Cref{eq:bounding_the_exp_loss_diff}, the case where $|L_1 - L_0| > 2k(\alpha+\lambda)$ ensures that  
\[
\expectation\left[\decRule^*_\loss(\tilde{R}(x)) \right] = \expectation\left[\decRule^*_\loss(R^{T_v}(x)) \right],
\]  
which completes the proof.
\end{proof}
Recalling that $\actionFunction^T_\loss$ is 
$\left(\actionFunctionsClass,\loss,k(\alpha+\lambda) \right)$-loss-minimizing, 
we conclude that $\tilde{\actionFunction}$ is 
$\left(\actionFunctionsClass,\loss,3k(\alpha+\lambda) \right)$-loss-minimizing, and the theorem follows.
\end{proof}

\begin{remark}\label{rem:loss_min_dec_rule_not_affine_as_cor}
    Returning to \Cref{lem:loss_min_dec_rule_not_affine}, which shows that the loss-minimizing decision rule is far from affine, we observe that this claim also follows as a corollary of the arguments just presented. Suppose, for the sake of contradiction, that $\decRule^*_\loss$ were close to affine. Then, for any calibrated predictor $\tilde{R}$, the action function $\actionFunction = \decRule^*_\loss(\tilde{R})$ would satisfy MAC by \Cref{cor:MA_implies_MAC}. By \Cref{thm:MC_for_loss_mini}, $\actionFunction$ would also be loss-minimizing. However, this contradicts \Cref{thm:impossibility_result_of_loss}, which shows that an action function cannot simultaneously satisfy MAC and minimize loss.
\end{remark}

\subsection{Relaxation of the multi-calibration requirement}\label{sec:relaxing_the_MC}
As discussed in \Cref{sec:intro:tech-overview,sec:related_work}, full-multi-calibration (\Cref{def:full_MC_for_types}) is often infeasible due to exponential sample complexity in $k$, the number of labels. This section is devoted to relaxing the full-multi-calibration requirement of \Cref{thm:MC_for_loss_mini} in order to achieve efficient sample complexity. We identify three settings in which this relaxation is possible: (i) when the label space is $\setOfTypes = \Lambda[0,1]$; (ii) when the loss function is known in advance during training of the predictor $\tilde{R}$; and (iii) when a {\em family} of loss functions is known during training. In the first two settings, the training time can be reduced to that of weak agnostic learning over the hypothesis class.

\subsubsection{Probabilities and linear losses}\label{sec:relaxing_the_MC_for_probs}

We first consider the case where the label space is $\setOfTypes = \Lambda[0,1]$, and the predictor $\tilde{R}$ maps each input $x$ to a distribution over $\Lambda[0,1]$. In this setting, we argue that all natural loss functions $\loss: \Lambda[0,1] \times \zo \to [0,1]$ are linear in their first argument. This corresponds to a modeling where the loss depends only on the realized outcome, rather than the probability assigned to it. For instance, in the context of loan approvals, the incurred loss depends solely on whether the borrower repays the loan; once the outcome is revealed, the specific predicted probability (e.g., $0.3$ vs.\ $0.8$) becomes irrelevant.

We show that under linear loss functions, probabilistic predictors offer no advantage over deterministic ones that output the expected prediction. In particular, the predictor that maps $x$ to the expected value under $\tilde{R}(x)$ incurs the same loss as $\tilde{R}$ itself. 

We define the {\em expectation-predictor}, which replaces a randomized prediction with its mean. Formally, for a predictor $\tilde{R}$ and input $x$, we set
\begin{equation}\label{eq:def_of_q}
\tilde{q}(x) = \sum_{\beta \in \Lambda[0,1]} \beta \cdot \tilde{R}(x)_\beta,
\end{equation}
and define
\begin{equation}\label{eq:def_of_Eq}
\expectation_{\tilde{R}}(x) = (\tilde{q}(x), 1-\tilde{q}(x)).
\end{equation}
We refer to $\expectation_{\tilde{R}}$ as the {\em expectation-predictor}. This transformation maps a $k$-dimensional predictor to a 2-dimensional one, consistent with the binary label space $\zo$. The expectation is taken over the distribution induced by $\tilde{R}$ with $x$ fixed. Although we could equivalently define the expectation-predictor as a scalar-valued function outputting $\tilde{q}(x)$, we retain the stochastic vector form for consistency with our prior notation.
Analogously, we define 
\[
q^*(x) = \sum_{\beta \in \Lambda[0,1]} \beta \cdot R^*(x)_\beta,
\]
and define Nature's expectation-predictor $\expectation_{R^*}$ in the same way. We measure Multi-Calibration with respect to $\mD = \DX \times \expectation_{R^*}(\DX)$, using the formulation in \Cref{def:full_MC_for_types} with $k = 2$, consistent with~\citet{omni}. 
Note that this is equivalent to training $\tilde{q}$ to be multi-calibrated with respect to $\mD = \DX \times q^*(\DX)$, and we later show how to achieve this efficiently using samples from $R^*$. 
We now state the main consequence of this reduction:

\begin{corollary} \label{cor:MC_for_loss_mini_for_probs}
Let $\C$, $\actionFunctionsClass$, and $\lambda$ be as in \Cref{thm:MC_for_loss_mini}, with $\setOfTypes = \Lambda[0,1]$. Let $\loss: \Lambda[0,1] \times \zo \to [0,1]$ be a loss function that is linear in its first argument. If the expectation-predictor $\expectation_{\tilde{R}}$ is $(\C, \alpha, \lambda)$-full-multi-calibrated with respect to $\mD$, then the action function $\decRule^*_{\loss} \circ \tilde{R}$ is $\left(\actionFunctionsClass, \loss, 6(\alpha + \lambda)\right)$-loss-minimizing.
\end{corollary}

\begin{proof}
The idea is to leverage the linearity of $\loss$, which is natural when $\setOfTypes = \Lambda[0,1]$, to reduce it to a ``binary'' loss function $\loss' : \zo \times \zo \to [0,1]$. We define $\loss'$ to coincide with $\loss$ on binary inputs: for all $b, a \in \zo$, let $\loss'(b,a) = \loss(b,a)$.\footnote{Strictly speaking, $\loss$ is not defined on inputs like $\loss(0,a)$ and $\loss(1,a)$, since $0,1 \notin \Lambda[0,1]$. We slightly abuse notation by interpreting $\loss'(b,a) = \loss(b,a)$ as shorthand for $\loss'(0,a) = \loss(\lambda/2,a)$ and $\loss'(1,a) = \loss(1 - \lambda/2,a)$, where $\lambda/2$ and $1 - \lambda/2$ are the smallest and largest elements in $\Lambda[0,1]$, respectively.}

We now apply \Cref{thm:MC_for_loss_mini} to the expectation-predictor $\expectation_{\tilde{R}}$, with respect to $\loss'$ and $\actionFunctionsClass$. By the theorem, the induced action function
\[
\tilde{f} = \decRule^*_{\loss'} \circ \expectation_{\tilde{R}}
\]
is $\left(\actionFunctionsClass, \loss', 6(\alpha + \lambda)\right)$-loss-minimizing, where $\decRule^*_{\loss'}$ denotes the optimal decision rule for $k = 2$ (as in \Cref{def:loss_opti_dec_rule}).

Our goal is to show that
\[
\tilde{\actionFunction} = \decRule^*_{\loss} \circ \tilde{R}
\]
is $\left(\actionFunctionsClass, \loss, 6(\alpha + \lambda)\right)$-loss-minimizing. To this end, we first establish that for any $x \in \X$, any predictor $R : \X \to \stochasticVecs$ over $\setOfTypes = \Lambda[0,1]$ (with corresponding $q$ and expectation-predictor $\expectation_{R}$ as defined in \Cref{eq:def_of_q,eq:def_of_Eq}), and any action $a \in \zo$, the following holds:
\begin{equation}\label{eq:goal}
\expectation_{b \sim \expectation_{R}(x)} \left[ \loss'(b,a) \right] = \expectation_{\beta \sim R(x)} \left[ \loss(\beta,a) \right].
\end{equation}
Indeed, the left-hand side of \Cref{eq:goal} expands to
\[
q(x) \cdot \loss'(1,a) + (1 - q(x)) \cdot \loss'(0,a) = q(x) \cdot \left( \loss'(1,a) - \loss'(0,a) \right) + \loss'(0,a).
\]
The right-hand side is
\[
\sum_{\beta \in \Lambda[0,1]} R(x)_\beta \cdot \loss(\beta,a).
\]
Since $\loss$ is linear in its first argument, we have
\[
\loss(\beta,a) = \beta \cdot \left( \loss(1,a) - \loss(0,a) \right) + \loss(0,a),
\]
and thus
\begin{align*}   
& \expectation_{\beta \sim R(x)} \left[ \loss(\beta,a) \right]  =\left( \sum_{\beta \in \Lambda[0,1]} \beta \cdot R(x)_\beta \right) \cdot \left( \loss(1,a) - \loss(0,a) \right) + \loss(0,a) \\& = q(x) \cdot \left( \loss(1,a) - \loss(0,a) \right) + \loss(0,a).
\end{align*}

By the definition of $\loss'$, the two sides are equal.

\Cref{eq:goal} implies that for any $a \in \zo$,
\[
\expectation_{b \sim \expectation_{\tilde{R}}(x)} \left[ \loss'(b,a) \right] = \expectation_{\beta \sim \tilde{R}(x)} \left[ \loss(\beta,a) \right],
\]
and, in particular,
\[
\argmin_{a \in \zo} \expectation_{b \sim \expectation_{\tilde{R}}(x)} \left[ \loss'(b,a) \right] = \argmin_{a \in \zo} \expectation_{\beta \sim \tilde{R}(x)} \left[ \loss(\beta,a) \right],
\]
which implies that $\tilde{f} = \tilde{\actionFunction}$. Applying \Cref{eq:goal} again yields
\[
\expectation_{b \sim \expectation_{R^*}(x)} \left[ \loss'(b,\tilde{f}(x)) \right] = \expectation_{\beta \sim R^*(x)} \left[ \loss(\beta,\tilde{f}(x)) \right],
\]
so
\begin{equation}\label{eq:exp_loss_of_squash_tilde_is_equal}
\expectation_{b \sim \expectation_{R^*}(x)} \left[ \loss'(b,\tilde{f}(x)) \right] = \expectation_{\beta \sim R^*(x)} \left[ \loss(\beta,\tilde{\actionFunction}(x)) \right].
\end{equation}
Similarly, for any $\actionFunction \in \actionFunctionsClass$, \Cref{eq:goal} implies
\begin{equation}\label{eq:exp_loss_of_squash_is_equal}
\expectation_{b \sim \expectation_{R^*}(x)} \left[ \loss'(b,\actionFunction(x)) \right] = \expectation_{\beta \sim R^*(x)} \left[ \loss(\beta,\actionFunction(x)) \right].
\end{equation}
Since $\tilde{f}$ is $(\actionFunctionsClass, \loss', 6(\alpha + \lambda))$-loss-minimizing, it holds for all $\actionFunction \in \actionFunctionsClass$ that
\[
\expectation_{b \sim \expectation_{R^*}(x)} \left[ \loss'(b,\tilde{f}(x)) \right] \le \expectation_{b \sim \expectation_{R^*}(x)} \left[ \loss'(b,\actionFunction(x)) \right] + 6(\alpha + \lambda).
\]
Substituting from \Cref{eq:exp_loss_of_squash_tilde_is_equal,eq:exp_loss_of_squash_is_equal}, we conclude that for all $\actionFunction \in \actionFunctionsClass$,
\[
\expectation_{\beta \sim R^*(x)} \left[ \loss(\beta,\tilde{\actionFunction}(x)) \right] \le 
\expectation_{\beta \sim R^*(x)} \left[ \loss(\beta,\actionFunction(x)) \right] + 6(\alpha + \lambda),
\]
which completes the proof.
\end{proof}

\paragraph{Efficient training of $\tilde{q}$.} 
We consider the standard one-dimensional multi-calibration setting, consistent with \Cref{def:full_MC_for_types} instantiated with $k = 1$.\footnote{Formally, \Cref{def:full_MC_for_types} defines a stochastic vector-valued predictor, but the same definition applies when the predictor outputs a scalar.} 
We reduce samples from $R^*$ to samples from a one-dimensional predictor $q^*$: for each $x \sim \DX$, we sample $\beta^*(x) \sim R^*(x)$ and then draw $y^*(x) \sim \text{Ber}(\beta^*(x))$.
By the law of total probability,
\begin{align*} 
\Pr\left[y^*(x) = 1\right] 
&= \sum_{\beta \in \Lambda[0,1]} \Pr\left[\beta^*(x) = \beta \right] \cdot \Pr\left[y^*(x) = 1 \mid \beta^*(x) = \beta\right] \\
&= \sum_{\beta \in \Lambda[0,1]} R^*(x)_\beta \cdot \beta = q^*(x).
\end{align*}
To efficiently train $\tilde{q}$, we apply the algorithm of Theorem 3.10 in \citet{MC}. This algorithm achieves sample complexity 
\[
O\left( \frac{\log(|\C| / \alpha \lambda \delta)}{\alpha^4 \lambda^{3/2}} \right),
\]
where $\delta$ bounds the failure probability of the algorithm. As for the training time, Theorem 4.2 in \citet{MC} shows that the task of learning a multi-calibrated predictor reduces to weak agnostic learning over $\C$, following the same framework described at the end of \Cref{sec:learning_MA_MC}.

\subsubsection{Types and a single loss}\label{sec:relaxing_the_MC_for_one_loss}

Next, we present an alternative way to relax the multi-calibration assumption in a different setting. Here, the loss function is known in advance, before training the predictor. On the other hand, this approach applies to the more general case of types, where we make no assumptions on the outcome space. 
The relation between \Cref{cor:MC_for_loss_mini_for_probs} and \Cref{cor:MC_for_loss_mini_when_loss_is_known} can be viewed as a tradeoff between the generality of the outcome space and the prior knowledge of the loss function: 
the former supports arbitrary losses but requires the outcome space to be a discrete set of probabilities, whereas the latter accommodates arbitrary outcome spaces at the cost of assuming that the loss function is fixed during training.

We define the {\em loss-weighted predictor} with respect to a loss function $\loss$, which replaces randomized predictions with scores that reflect the expected loss. Formally, for a loss function $\loss: \setOfTypes \times \zo \to [0,1]$, a predictor $R$, and an input $x \in \X$, we define:
\begin{equation}\label{eq:def_of_p}
p_{R,\loss}(x) \defeq \frac{1 - \sum_{\type \in \setOfTypes} \left( \loss(\type,1) - \loss(\type,0) \right) \cdot R(x)_\type}{2}.
\end{equation}
This implies that $\forall x \in X$,
\[
\decRule^*_\loss(R(x)) = 1 \iff \expectation_{\type(x) \sim R(x)} \left[\loss(\type(x), 1) - \loss(\type(x), 0)  \right] < 0 \iff p_{R,\loss}(x) > \frac{1}{2}. 
\]
We denote
\[
\tilde{p}(x) \defeq p_{\tilde{R},\loss}(x), \text{ and } p^*(x) \defeq p_{R^*,\loss}(x).
\]
We refer to $\tilde{p}$ as the {\em loss-weighted predictor} and measure (one-dimensional) multi-calibration of $\tilde{p}$ w.r.t. $\mD = \DX \times p^*(\DX)$. Note that $\tilde{p}(x), p^*(x) \in [0,1]$ for all $x$ by construction. 
The following corollary shows that multi-calibration of $\tilde{p}$ suffices to guarantee that $\decRule^*_{\loss} \circ \tilde{R}$ is loss-minimizing. This result can be viewed as a simplified special case of \Cref{thm:MC_for_loss_mini}.
For simplicity, we trivially assume that $\X \in \C$ and that the discretization is aligned with the threshold $1/2$.

\begin{corollary} \label{cor:MC_for_loss_mini_when_loss_is_known}
    Let $\C$, $\actionFunctionsClass$, and $\lambda$ be as in \Cref{thm:MC_for_loss_mini}, and let $\loss: \setOfTypes \times \zo \to [0,1]$ be a loss function. If the loss-weighted predictor $\tilde{p}$ is $(\C, \alpha, \lambda)$-full-multi-calibrated with respect to $\mD$, then the action function $\decRule^*_{\loss} \circ \tilde{R}$ is $\left(\actionFunctionsClass, \loss, 4\alpha/\lambda\right)$-loss-minimizing.
\end{corollary}

\begin{proof}
Let $\actionFunction \in \actionFunctionsClass$ and write
$\tilde{\actionFunction}_\loss \defeq \decRule^*_{\loss} \circ \tilde{R}$.
For any $\actionFunction' : \X \to \{0,1\}$, define
\[
    \Delta_{\actionFunction'}(x)
    \defeq
    \expectation_{\type^*(x)\sim R^*(x)}
    \left[
        \loss(\type^*(x),\actionFunction'(x))
    \right].
\]
By the definition of $p^*$ and $\tilde p$,
\[
    \expectation_{\type^*(x)\sim R^*(x)}
    \left[\loss(\type^*(x),1)-\loss(\type^*(x),0)\right]
    = 1-2p^*(x),
\]
and similarly the loss difference according to $\tilde p$ is $1-2\tilde p(x)$.
Therefore,
\[
\begin{aligned}
&\expectation_{x\sim\DX}
\left[
    \Delta_{\tilde{\actionFunction}_\loss}(x)
    -
    \Delta_{\actionFunction}(x)
\right] \\
&=
\expectation_{x\sim\DX}
\left[
    (1-2p^*(x))
    \cdot
    \left(\tilde{\actionFunction}_\loss(x)-\actionFunction(x)\right)
\right] \\
&=
\expectation_{x\sim\DX}
\left[
    (1-2\tilde p(x))
    \cdot
    \left(\tilde{\actionFunction}_\loss(x)-\actionFunction(x)\right)
\right]
+
2\expectation_{x\sim\DX}
\left[
    (\tilde p(x)-p^*(x))
    \cdot
    \left(\tilde{\actionFunction}_\loss(x)-\actionFunction(x)\right)
\right].
\end{aligned}
\]
The first term is non-positive, since
$\tilde{\actionFunction}_\loss(x)=1$ iff $\tilde p(x)>1/2$, and hence
$\tilde{\actionFunction}_\loss(x)$ minimizes the loss according to $\tilde p(x)$.
Thus,
\[
\begin{aligned}
&\expectation_{x\sim\DX}
\left[
    \Delta_{\tilde{\actionFunction}_\loss}(x)
    -
    \Delta_{\actionFunction}(x)
\right] \\
&\le
2\left|
\expectation_{x\sim\DX}
\left[
    (\tilde p(x)-p^*(x))\tilde{\actionFunction}_\loss(x)
\right]
\right|
+
2\left|
\expectation_{x\sim\DX}
\left[
    (\tilde p(x)-p^*(x))\actionFunction(x)
\right]
\right|.
\end{aligned}
\]
The second term is at most $2\alpha/\lambda$, since
$\{x:\actionFunction(x)=1\}\in\C$ and summing the full-multi-calibration guarantee
over all $\lambda$-level sets of $\tilde p$ contributes at most $\alpha/\lambda$.
The first term is also at most $2\alpha/\lambda$, since
$\{x:\tilde{\actionFunction}_\loss(x)=1\}=\{x:\tilde p(x)>1/2\}$ is a union of
$\lambda$-level sets of $\tilde p$ inside $\X$, and $\X\in\C$.
Altogether,
\[
\expectation_{\substack{x\sim\DX\\ \type^*(x)\sim R^*(x)}}
\left[
    \loss\left(\type^*(x),\tilde{\actionFunction}_\loss(x)\right)
\right]
\le
\expectation_{\substack{x\sim\DX\\ \type^*(x)\sim R^*(x)}}
\left[
    \loss\left(\type^*(x),\actionFunction(x)\right)
\right]
+
\frac{4\alpha}{\lambda}.
\]
Since this holds for every $\actionFunction\in\actionFunctionsClass$,
$\tilde{\actionFunction}_\loss$ is
$(\actionFunctionsClass,\loss,4\alpha/\lambda)$-loss-minimizing.
\end{proof}

\paragraph{Efficient training of $\tilde{p}$.}
As in \Cref{sec:relaxing_the_MC_for_probs}, we consider the standard one-dimensional multi-calibration setting and apply the algorithm of~\citet{MC} to efficiently train $\tilde{p}$. To do so, we reduce samples from $R^*$ to samples from $p^*$: for each $x \sim \DX$, we first sample $\type^*(x) \sim R^*(x)$, and then define $y^*(x) \sim \text{Ber} \left( \frac{1 - \left( \loss(\type^*(x),1) - \loss(\type^*(x),0) \right)}{2} \right)$. By the law of total probability,
\begin{align*}
\Pr_{y^*(x)}\left[y^*(x) = 1\right] 
&= \sum_{\type \in \setOfTypes} \Pr_{\type^*(x) \sim R^*(x)}\left[\type^*(x) = \type \right] \cdot \frac{1 - \left( \loss(\type,1) - \loss(\type,0) \right)}{2} \\
&= \frac{1 - \expectation_{\type^*(x) \sim R^*(x)} \left[ \loss(\type^*(x),1) - \loss(\type^*(x),0) \right]}{2} = p^*(x).
\end{align*}

\subsubsection{Types and a family of losses}\label{sec:relaxing_the_MC_for_many_losses}

\Cref{cor:MC_for_loss_mini_for_probs,cor:MC_for_loss_mini_when_loss_is_known} can be viewed as reductions from the task of loss minimization to that of training a one-dimensional predictor to be multi-calibrated. Generalizing \Cref{cor:MC_for_loss_mini_when_loss_is_known} to a {\em family} of loss functions requires a stronger guarantee: the learned predictor must be multi-calibrated not only with respect to a single loss, but simultaneously for every loss in a given family $\lossFamily$.

To that end, we aim to satisfy multi-calibration with respect to all convex combinations induced by $\lossFamily$, rather than a fixed weighting. We reduce this problem to identifying, in each round, a pair $(\loss, \actionFunction)$ with $\loss \in \lossFamily$ and $\actionFunction \in \actionFunctionsClass$ such that the expected loss of $\decRule^*_\loss \circ \tilde{R}$ exceeds that of $\actionFunction$. We refer to the procedure that searches for such violating pairs as an {\em auditor}. In our setting, the auditor is realized by evaluating a family of distinguishers --- each parameterized by $(\loss, \actionFunction)$ --- that attempt to distinguish Nature's outcomes from those generated by $\tilde{R}$. A distinguisher that succeeds reveals a violation, and the corresponding pair is used to update the predictor. As in previous settings, query complexity remains low as long as each update affects a non-negligible portion of the domain. However, the total runtime is dominated by the auditing step: namely, the search over $\lossFamily \times \actionFunctionsClass$. Unlike the single-loss case, this setting does not reduce to weak agnostic learning over $\C$.

\begin{corollary} \label{cor:MC_for_loss_mini_when_family_of_losses_is_known}
    Let $\C$ and $\actionFunctionsClass$ be as in \Cref{thm:MC_for_loss_mini}, and let $\lossFamily \subseteq \{ \setOfTypes \times \zo \to [0,1] \}$ be a family of loss functions. Given $\tilde{O}\left( \log k \cdot (\log |\C| + \log |\lossFamily|)/ \eps^4 \right)$ samples, one can train a predictor $\tilde{R}$ such that, for every $\loss \in \lossFamily$, the composed rule $\decRule^*_{\loss} \circ \tilde{R}$ is $\left(\actionFunctionsClass, \loss, \eps \right)$-loss-minimizing.
\end{corollary}

\begin{proof}
In \Cref{sec:learning_MA_MC}, we present {\em Outcome Indistinguishability} (OI)~\citep{OI} and discuss an algorithm for training an Outcome Indistinguishable predictor. At a high level, the idea behind OI is to define a family of computationally bounded distinguishers $\mathcal{A}$, such that a target objective (in our case, loss minimization) reduces to the distinguishers' inability to tell apart Nature's outcomes from those generated based on the predictions. A probabilistic predictor $\tilde{R}: \X \to \stochasticVecs$ is said to be $(\mathcal{A},\eps)$-OI with respect to $\mD$ if for any $A \in \mathcal{A}$,
\[
\left| \Pr_{(x,\type^*(x)) \sim \mD} \left[ A\left(x, \type^*(x); \tilde{R}(x) \right) = 1 \right] - 
\Pr_{(x, \tilde{\type}(x)) \sim \DX \times \tilde{R}(x)} \left[ A\left(x, \tilde{\type}(x); \tilde{R}(x) \right) = 1 \right]
\right| \le \eps.
\]
This variant is called Sample-Access-OI. \Cref{thm:beyond} (from~\citet{OI}) shows that Sample-Access-OI can be achieved using 
\[
\tilde{O} \left( \frac{\log k \cdot \log |\mathcal{A}|}{\eps^4} \right)
\]
samples. We refer the reader to \Cref{sec:learning_MA_MC} for a detailed exposition.

We define a family of distinguishers such that if $\tilde{R}$ is OI with respect to them, then $\decRule^*_{\loss} \circ \tilde{R}$ is $\left(\actionFunctionsClass, \loss, \eps \right)$-loss-minimizing. Each distinguisher is parameterized by a loss function $\loss \in \lossFamily$ and a hypothesis $\actionFunction \in \actionFunctionsClass$, and aims to detect whether $\actionFunction$ outperforms $\decRule^*_{\loss} \circ \tilde{R}$ in terms of loss. Formally, the randomized Boolean distinguisher $A_{\loss,\actionFunction}$ operates as follows:
\begin{align*}
A_{\loss,\actionFunction}\left(x,\type(x),\tilde{R}(x)\right) =\ & \text{Compute } p(x) := \loss(\type(x), \decRule^*_{\loss}(\tilde{R}(x))) - \loss(\type(x), \actionFunction(x)) \\
& \text{Output } 1 \text{ with probability } \frac{p(x) + 1}{2}.
\end{align*}
While the distinguishers could have been defined to output $p(x)$ directly in $[-1,1]$, we instead normalize and sample a bit, to be consistent with the formulation of~\citet{OI}.

We define
\[
\tilde{p}(x) := \loss(\tilde{\type}(x), \decRule^*_{\loss}(\tilde{R}(x))) - \loss(\tilde{\type}(x), \actionFunction(x)) 
\quad \text{and} \quad
p^*(x) := \loss(\type^*(x), \decRule^*_{\loss}(\tilde{R}(x))) - \loss(\type^*(x), \actionFunction(x)).
\]
By \Cref{thm:beyond}, given
\[
\tilde{O} \left( \frac{\log k \cdot \log |\mathcal{A}|}{(\eps')^4} \right)
\]
samples from $R^*$, we can learn a predictor $\tilde{R}$ such that for all $A_{\loss,\actionFunction} \in \mathcal{A}$,
\[
\left| 
\Pr_{(x,\type^*(x)) \sim \mD} \left[ A_{\loss,\actionFunction}(x, \type^*(x); \tilde{R}(x)) = 1 \right] - 
\Pr_{(x, \tilde{\type}(x)) \sim \DX \times \tilde{R}(x)} \left[ A_{\loss,\actionFunction}(x, \tilde{\type}(x); \tilde{R}(x)) = 1 \right]
\right| \le \eps'.
\]
Equivalently,
\begin{equation}\label{eq:exps_are_close}
\left| 
\mathbb{E}_{x,\type^*(x)} \left[ A_{\loss,\actionFunction}(x, \type^*(x); \tilde{R}(x)) \right] - 
\mathbb{E}_{x,\tilde{\type}(x)} \left[ A_{\loss,\actionFunction}(x, \tilde{\type}(x); \tilde{R}(x)) \right]
\right| \le \eps'.
\end{equation}
Let us first focus on the right term. By the definition of $\decRule^*_{\loss}$ (see \Cref{def:loss_opti_dec_rule}),
\[
\expectation_{\tilde{\type}(x) \sim \tilde{R}(x)}\left[\loss\left(\tilde{\type}(x), \decRule^*_{\loss}(\tilde{R}(x)) \right) \right] \le  \expectation_{\tilde{\type}(x) \sim \tilde{R}(x)}\left[\loss\left(\tilde{\type}(x), \actionFunction(x) \right) \right],
\]
so
\[
\expectation_{\tilde{\type}(x) \sim \tilde{R}(x)}\left[\loss\left(\tilde{\type}(x), \decRule^*_{\loss}(\tilde{R}(x)) \right) - \loss\left(\tilde{\type}(x), \actionFunction(x) \right) \right] \le 0.
\]
Recalling the definition of $\tilde{p}(x)$, we get that
\[
\forall x \in \X,\ \expectation_{\tilde{\type}}\left[ \tilde{p}(x) \right] \le 0 \implies \frac{\expectation_{\tilde{\type}}\left[ \tilde{p}(x) \right] + 1}{2} \le \frac{1}{2},
\]
and, overall,
\[
\expectation_{x,\tilde{\type}(x), A_{\loss,\actionFunction}} \left[ A_{\loss,\actionFunction}\left(x, \tilde{\type}(x); \tilde{R}(x) \right) \right] = \frac{\expectation_{\tilde{\type}}\left[ \tilde{p}(x) \right] + 1}{2} \le \frac{1}{2}.
\]
Returning to \Cref{eq:exps_are_close}, this implies that
\[
\expectation_{x, \type^*(x), A_{\loss,\actionFunction}} \left[ A_{\loss,\actionFunction}\left(x, \type^*(x); \tilde{R}(x) \right) \right] \le \frac{1}{2} + \eps',
\]
that is,
\[
\expectation_{x,\type^*(x)} \left[ \frac{p^*(x) + 1}{2} \right] \le \frac{1}{2} + \eps',
\]
so $\expectation_{x,\type^*(x)} [p^*(x)] \le 2\eps'$. By the definition of $p^*$, this means
\[
\expectation_{x,\type^*(x)} \left[\loss(\type^*(x), \decRule^*_{\loss}(\tilde{R}(x))) - \loss(\type^*(x), \actionFunction(x))\right] \le 2\eps',
\]
hence,
\[
\expectation_{x,\type^*(x)} \left[\loss(\type^*(x), \decRule^*_{\loss}(\tilde{R}(x))) \right]  \le  \expectation_{x,\type^*(x)} \left[\loss(\type^*(x), \actionFunction(x))\right] + 2\eps'.
\]
Setting $\eps' = \eps / 2$ concludes the proof.
\end{proof}

\section{Deferred proofs}
This section provides the rather standard proofs deferred from \Cref{sec:prelims}.

\subsection{Multi-accuracy and multi-calibration}\label{sec:deferred_MA_MC}

We start with the proof of \Cref{clm:accuracy_holds_on_a_large_subset}. 
\begin{claim}
    Let $S, S'$ be two sets such that $S' \subseteq S$. If $\tilde{R}$ is $\alpha$-coordinate-wise-accurate w.r.t. $\mD$ on $S$, then it is $\alpha'$-coordinate-wise-accurate w.r.t. $\mD$ on $S'$, where $\alpha' = \alpha + \Pr_{x \sim \DX} \left[ x \in S \setminus S' \right]$.
\end{claim}
\begin{proof} 
For any $\type \in \setOfTypes$,
\begin{align*}   
& \Pr\left[\type^*(x) = \type \land x \in S' \right] - 
\Pr\left[\tilde{\type}(x) = \type \land x \in S' \right]  \\ 
& = \Pr\left[\type^*(x) = \type \land x \in S \right] - 
\Pr\left[\tilde{\type}(x) = \type \land x \in S \right]  \\ 
& - \Pr\left[\type^*(x) = \type \land x \in S \setminus S' \right] + 
\Pr\left[\tilde{\type}(x) = \type \land x \in S \setminus S' \right],
\end{align*}
where the probabilities are taken over $x \sim \DX, \type^*(x) \sim R^*(x)$ and $x \sim \DX, \tilde{\type}(x) \sim \tilde{R}(x)$, respectively.  
By assumption, the difference is bounded by $\Pr_{x \sim \DX} \left[ x \in S \setminus S' \right] + \alpha$.
\end{proof}

Next, we prove \Cref{clm:MC_implies_MA}. 
\begin{claim}[Coordinate-Wise-Multi-Calibration Implies Coordinate-Wise-Multi-Accuracy]
    In the setting of \Cref{def:relaxed_MC_for_types}, if $\tilde{R}$ is $(\C,\alpha,\lambda)$-coordinate-wise-multi-calibrated, then $\tilde{R}$ is also $(\C,\alpha/\lambda)$-coordinate-wise-multi-accurate.
\end{claim}
\begin{proof}
    Let $S \in \C$ and $\type \in \setOfTypes$.
    By the definition of $S_{\type, \beta}$, we have $S \subseteq \bigcup_{\beta \in \Lambda[0,1]} S_{\type, \beta}$, so 
    \begin{align*}
        &\left|
        \Pr[\type^*(x)=\type \land x\in S]-\Pr[\tilde{\type}(x)=\type \land x\in S]
        \right|  \\ & \le\sum_{\beta\in\Lambda[0,1]}
        \left|
            \Pr[\type^*(x)=\type \land x\in S_{\type,\beta}]
            -\Pr[\tilde{\type}(x)=\type \land x\in S_{\type,\beta}]
        \right|
        \le \frac{\alpha}{\lambda}.
            \end{align*}
\end{proof}

\subsubsection{Notions of multi-accurate rankings}\label{sec:MA_for_rankings}
In this section, we examine the relationship between two definitions of Multi-Accuracy for Probabilistic Rankings, previously introduced in \Cref{sec:MA_defs}.  
The first and stronger definition, \Cref{def:MA_for_prob_ranking}, evaluates accuracy with respect to {\em thresholds}. Specifically, for every set $S$ in the collection $\C$ and every rank $\tau$, the probabilities assigned by $R^*$ and $\tilde{R}$ to an individual from $S$ being above threshold $\tau$ are similar.  

The second definition, which is also natural, enforces {\em equality} on ranks, namely, coordinate-wise- equality. This alternative is motivated by decision rules that operate on specific segments---for example, a particular bank or university might target a specific segment of the population. The following definition restates \Cref{def:eq_MA_for_types}, with a modified notation suited to the case of rankings, where $\setOfTypes = [k]$.

\begin{definition}[Coordinate-Wise-Multi-Accuracy for Probabilistic Rankings] \label{def:MA_for_prob_ranking_with_equality}
In the setting of \Cref{def:MA_for_prob_ranking}, a probabilistic ranking $\tilde{R}: \X \to \stochasticVecs$ satisfies $(\C,\alpha)$-coordinate-wise-multi-accuracy w.r.t. $\mD$ if for every $S \in \C$ and every $\tau \in [k]$,
\[ 
\left| 
\Pr_{\substack{x \sim \DX \\ r^*(x) \sim R^*(x)}} \left[r^*(x) = \tau \land x \in S \right] - 
\Pr_{\substack{x \sim \DX \\ \tilde{r}(x) \sim \tilde{R}(x)}}
\left[\tilde{r}(x) = \tau \land x \in S \right] \right| \le \alpha.
\]
\end{definition}

The following claim shows that this definition is strictly weaker than the threshold definition.
\begin{claim} \label{clm:MA_implies_equality_MA}
    $(\C,\alpha)$-threshold-multi-accuracy implies $(\C,2\alpha)$-coordinate-wise-multi-accuracy.
\end{claim}
\begin{proof}
Let $S \in \C$ and $\tau \in [k]$. If $\tau = k$, then $(\C,\alpha)$-coordinate-wise-multi-accuracy follows directly, since $r(x) = k \iff r(x) > k-1$. Otherwise, by the assumption,
\[ \left| \Pr \left[r^*(x) > \tau \land x \in S \right] - \Pr
\left[\tilde{r}(x) > \tau \land x \in S \right] \right| \le \alpha\]
and
\[ \left| \Pr \left[r^*(x) > \tau-1 \land x \in S \right] - 
\Pr \left[\tilde{r}(x) > \tau-1 \land x \in S \right] \right| \le \alpha.\]
Hence,
\begin{align*}
&|\Pr \left[r^*(x) = \tau \land x \in S \right] - \Pr \left[\tilde{r}(x) = \tau \land x \in S \right]| = \\
& |\Pr \left[r^*(x) > \tau-1 \land x \in S \right] - 
\Pr \left[\tilde{r}(x) > \tau-1 \land x \in S \right] - \\
&\left(\Pr \left[r^*(x) > \tau \land x \in S \right] - 
\Pr \left[\tilde{r}(x) > \tau \land x \in S \right]\right)|
\le 2\alpha 
\end{align*}
follows from triangle inequality. 

On the other hand, note that an error of $2\alpha$ is tight. Let $k \ge 3$, $S \in \C$ and let $\tilde{R}$ be such that
\begin{align*}
& \Pr \left[\tilde{r}(x) = k \land x \in S \right] = \Pr \left[r^*(x) = k \land x \in S \right] + \alpha, \\
& \Pr \left[\tilde{r}(x) = k-1 \land x \in S \right] = \Pr \left[r^*(x) = k-1 \land x \in S \right] - 2\alpha, \\
& \Pr \left[\tilde{r}(x) = k-2 \land x \in S \right] = \Pr \left[r^*(x) = k-2 \land x \in S \right] + \alpha, \\
\forall \tau \le k-3,\ & \Pr \left[\tilde{r}(x) = \tau \land x \in S \right] = \Pr\left[r^*(x) = \tau \land x \in S \right],
\end{align*}
where all probabilities are taken over sampling an element $x \sim \DX$ and a rank from either $\tilde{R}(x)$ or $R^*(x)$.
This ranking $\tilde{R}$ is $\alpha$-threshold-multi-accurate on $S$, yet its coordinate-wise-multi-accuracy error is $2\alpha$.
\end{proof}

On the other hand, the following claim shows that \Cref{def:MA_for_prob_ranking_with_equality} implies \Cref{def:MA_for_prob_ranking} with multiplicative $k$ blowup in the error parameter:
\begin{claim}\label{clm:coordinate_wise_implies_thres_with_k_blowup}
    $(\C,\alpha)$-coordinate-wise-multi-accuracy implies $(\C,\Theta(k \cdot \alpha))$-threshold-multi-accuracy. 
\end{claim}
\begin{proof}
First, we show that $(\C,\alpha)$-coordinate-wise-multi-accuracy implies $(\C,O(k \cdot \alpha))$-threshold-multi-accuracy.
Let $S \in \C$ and $\tau \in [k-1]$. By assumption,
\begin{align*}
&|\Pr \left[r^*(x) > \tau \land x \in S \right] - \Pr \left[\tilde{r}(x) > \tau \land x \in S \right]| = \\ 
& |\Pr \left[r^*(x) \in \{\tau+1, \dots, k\} \land x \in S \right] - \Pr \left[\tilde{r}(x) \in \{\tau+1, \dots, k\}  \land x \in S \right]| = \\
& \left| \sum_{i = \tau+1}^k \left(\Pr \left[r^*(x) = i \land x \in S \right] - \Pr \left[\tilde{r}(x) = i \land x \in S \right]  \right)\right| \le (k-1)\cdot \alpha.
\end{align*}
Next, we demonstrate that this bound is tight by presenting an example where a ranking is $(\C,\alpha)$-coordinate-wise-multi-accurate but incurs a threshold-multi-accuracy error of $(k-1) \cdot \alpha$. Consider the following example. Let $k \in \N_{\mathsf{even}}$ and let $R^*$ be a deterministic Nature that is uniform over the ranks $\forall S \in \C$, namely,
\[
\forall \tau \in [k],\ \Pr \left[ r^*(x) = \tau \land x \in S \right] = \frac{1}{k}.
\]
Consider the following deterministic ranking $\tilde{R}$: For $S \in \C$,
\[
\tau \in \{1,\dots, k/2\},\ \Pr\left[ \tilde{r}(x) = \tau \land x \in S \right] = \frac{1}{k} + \alpha,
\]
and
\[
\tau \in \{k/2+1,\dots, k\},\ \Pr \left[ \tilde{r}(x) = \tau \land x \in S \right] = \frac{1}{k} - \alpha.
\]
By definition, it follows immediately that $\tilde{R}$ is $(\C,\alpha)$-coordinate-wise-multi-accurate. On the other hand, consider $\tau = \frac{k}{2} - 1$ in \Cref{def:MA_for_prob_ranking}:
\[ 
\left| 
\Pr \left[r^*(x) > \frac{k}{2}-1 \land x \in S \right] - 
\Pr
\left[\tilde{r}(x) > \frac{k}{2}-1 \land x \in S \right] \right| = (k/2) \cdot \alpha.
\]
This means that $\tilde{R}$ is $(\C,\Omega(k \cdot \alpha))$-threshold-multi-accurate, and the claim follows.
\end{proof}

\subsubsection{Learning multi-accurate and multi-calibrated predictors}\label{sec:learning_MA_MC}

We use the framework of {\em Outcome Indistinguishability}, introduced by  \citet{OI} in the context of binary prediction tasks. It articulates the goals of prediction using the language of computational indistinguishability: a predictor is outcome-indistinguishable (OI) if no computationally bounded observer can distinguish Nature's outcomes from those generated based on the predictions. This framework ensures that predictive models yield outcomes that cannot be efficiently refuted using real-life observations. 

Dwork et al.\ also provide algorithms for learning OI predictors. The intuition behind their learning approach is as follows: if there exists a computationally bounded distinguisher $A$ that distinguishes between Nature $R^*$ and the learned predictor $\tilde{R}$, use $A$ to update $\tilde{R}$; otherwise, $R^*$ and $\tilde{R}$ are indistinguishable, and the process terminates. This strategy can be interpreted from different perspectives---as a form of boosting or gradient descent---and is closely related to the approach used by \citet{MC} to learn multi-calibrated predictors.  

In a subsequent work (\citep{BeyondBer}), Dwork et al.\ extend the concept of outcome indistinguishability beyond binary (Bernoulli) outcomes to more complex, discrete, or continuous domains. They provide learning algorithms for constructing OI predictors, which we use next to achieve multi-accurate and multi-calibrated probabilistic predictors.

Formally, let $\mathcal{A}$ be a class of distinguishers and $\eps$ an indistinguishability parameter. A probabilistic predictor $\tilde{R}: \X \to \stochasticVecs$ is said to be $\left(\mathcal{A},\eps\right)$-{\em no-access}-OI with respect to $\mD$ if for any $A \in \mathcal{A}$,   
\[
\left| 
\Pr_{\substack{(x,\type^*(x)) \sim \mD \\ \text{$A$'s randomness}}} \left[ A\left(x, \type^*(x) \right) = 1 \right] - 
\Pr_{\substack{(x, \tilde{\type}(x)) \sim \DX \times \tilde{R}(x) \\ \text{$A$'s randomness}}} \left[ A\left(x, \tilde{\type}(x) \right) = 1 \right]
\right| \le \eps,
\]
and \emph{sample-access}-OI with respect to $\mD$ if for any $A \in \mathcal{A}$,   
\[
\left| 
\Pr_{\substack{(x,\type^*(x)) \sim \mD \\ \text{$A$'s randomness}}} \left[ A\left(x, \type^*(x); \tilde{R}(x) \right) = 1 \right] - 
\Pr_{\substack{(x, \tilde{\type}(x)) \sim \DX \times \tilde{R}(x) \\ \text{$A$'s randomness}}} \left[ A\left(x, \tilde{\type}(x); \tilde{R}(x) \right) = 1 \right]
\right| \le \eps.
\]
Dwork et al.\ present other, stronger variants of outcome indistinguishability, but we focus exclusively on these two. Moreover, we comment that our representation of distributions as stochastic vectors, which explicitly encode the probability mass function, coincides with the representation used by Dwork et al.\ for discrete random variables over a finite domain $|\setOfTypes| = k < \infty$.

\begin{theorem}[{Learning OI Predictors, Theorem 20 in \citet{BeyondBer}}]\label{thm:beyond}
There exists an algorithm for learning an $\left(\mathcal{A},\eps\right)$-outcome-indistinguishable probabilistic predictor $\tilde{R}: \X \to \stochasticVecs$ with respect to $\mD$. The sample complexity is bounded by  
\[
\tilde{O} \left( \frac{\log k \cdot \log |\mathcal{A}|}{\eps^4} \right)
\]
for the case of Sample-Access-OI, and by 
$O \left( \frac{\log |\mathcal{A}|}{\eps^2} \right)$  
for the case of No-Access-OI. Assuming that the distinguishers in $\mathcal{A}$ can be evaluated in time $\text{time}_{\mathcal{A}}$, the running time is  
\[
\tilde{O} \left( \frac{k \cdot \text{time}_{\mathcal{A}} \cdot \log |\mathcal{A}|}{\eps^6} + \frac{|\mathcal{A}| \cdot \text{time}_{\mathcal{A}} \cdot \log k}{\eps^4} \right).
\]  
With overwhelming probability over its samples, the algorithm produces a probabilistic predictor $\tilde{R}$ that can be evaluated in time  
\[
\tilde{O} \left( \frac{k \cdot \text{time}_{\mathcal{A}} \cdot \log |\mathcal{A}|}{\varepsilon^2} \right)
\]
for the case of Sample-Access-OI, and by $O \left( k \cdot \text{time}_{\mathcal{A}} \cdot \log |\mathcal{A}| \right)$
for the case of No-Access-OI.
\end{theorem}

Recall \Cref{def:eq_MA_for_types}, where $k = |\setOfTypes|$. Using \Cref{thm:beyond}, we derive our result for learning coordinate-wise-multi-accurate probabilistic predictors:  

\begin{theorem}[Learning Coordinate-Wise-Multi-Accurate Predictors]\label{thm:learning_MA}
Let $\C \subseteq \zo^{|\X|}$ be a collection of sets such that the set membership evaluation time for any $S \in \C$ is $T$.  
There exists an algorithm for learning a $(\C,\alpha)$-coordinate-wise-multi-accurate probabilistic predictor $\tilde{R}: \X \to \stochasticVecs$ with respect to $\mD$. The sample complexity is bounded by  
\[
O \left( \frac{\log |\C| + \log k}{\alpha^2} \right).
\]  
The running time is  
\[
\tilde{O} \left( \frac{k \cdot T \cdot |\C|}{\alpha^6} \right).
\]  
With overwhelming probability over its samples, the algorithm produces a probabilistic predictor $\tilde{R}$ that can be evaluated in time $O \left( k \log k \cdot T \cdot \log |\C| \right)$.
\end{theorem}

\begin{proof}
We define the family of no-access-distinguishers $\mathcal{A} = \left(A_{S,\type}\right)_{S \in \C, \type \in \setOfTypes}$ as follows:
\[
A_{S,\type}(x,y) = \begin{cases}
    1 & \text{if } y = \type \text{ and } x \in S, \\
    0 & \text{otherwise}.
\end{cases}
\]
Let $\tilde{R}: \X \to \stochasticVecs$ be an $(\mathcal{A},\alpha)$-no-access-OI predictor with respect to $\mD$, guaranteed by \Cref{thm:beyond}. This implies that for any $A_{S,\type} \in \mathcal{A}$,  
\[
\left| \Pr_{(x,\type^*(x)) \sim \mD} \left[ A_{S,\type}\left(x, \type^*(x) \right) = 1 \right] - 
\Pr_{\left(x, \tilde{\type}(x)\right) \sim \DX \times \tilde{R}(x)} \left[ A_{S,\type}\left(x, \tilde{\type}(x) \right) = 1 \right]
\right| \le \alpha.
\]  
Expanding the definition of $A_{S,\type}$, this ensures that for any $S \in \C$ and any $\type \in \setOfTypes$,
\[
\left| \Pr_{(x,\type^*(x)) \sim \mD} \left[ \type^*(x) = \type \land x \in S \right] - 
\Pr_{\left(x, \tilde{\type}(x)\right) \sim \DX \times \tilde{R}(x)} \left[ \tilde{\type}(x) = \type \land x \in S \right]
\right| \le \alpha.
\]    
Thus, $\tilde{R}$ satisfies $(\mathcal{C}, \alpha)$-coordinate-wise-multi-accuracy, which completes the proof. The claimed complexity bounds follow from \Cref{thm:beyond}.
\end{proof}
To learn threshold-multi-accurate rankings directly, as per \Cref{def:MA_for_prob_ranking}, which is stronger than \Cref{def:eq_MA_for_types} in case of rankings, we can alternatively define the distinguishers 
\[
A_{S,\tau}(x,y) = \begin{cases}
    1 & \text{if } y > \tau \text{ and } x \in S, \\
    0 & \text{otherwise},
\end{cases}
\]
for any $\tau \in [k]$, and achieve threshold-multi-accuracy without any increase in complexity. 

\bigskip
Next, we address the more complex task of learning multi-calibrated predictors. The following theorem shows that coordinate-wise-multi-calibration, as defined in \Cref{def:relaxed_MC_for_types}, can be achieved efficiently.  

\begin{theorem}[Learning Multi-Calibrated Predictors]\label{thm:learning_MC}
Let $\C \subseteq \zo^{|\X|}$ be a collection of sets such that the set membership evaluation time for any $S \in \C$ is $T$, and let $\lambda \in (0,1)$ be a discretization parameter. 
There exists an algorithm for learning a $(\C,\alpha,\lambda)$-coordinate-wise-multi-calibrated probabilistic predictor $\tilde{R}: \X \to \stochasticVecs$ with respect to $\mD$. The sample complexity is bounded by  
\[
\tilde{O} \left( \frac{\log^2 k \cdot \left( \log |\C| + \log\left(\frac{1}{\lambda}\right) \right)}{\alpha^4} \right).
\]  
The running time is  
\[
\tilde{O} \left( \frac{k \cdot T \cdot \left( |\C| \cdot \log(k) + \log \frac{1}{\lambda} \right) }{\lambda \cdot \alpha^6} \right).
\]  
With overwhelming probability over its samples, the algorithm produces a probabilistic predictor $\tilde{R}$ that can be evaluated in time  
\[
\tilde{O} \left( \frac{k \cdot T \cdot \log \left( \log k + \log |\C| + \log \frac{1}{\lambda} \right)}{\alpha^2} \right).
\]
\end{theorem}
\begin{proof}
We define the family of sample-access-distinguishers $\mathcal{A} = \left(A_{S,\type,\beta}\right)_{S \in \C, \type \in \setOfTypes, \beta \in \Lambda[0,1]}$ with respect to the discretization $\lambda$ as follows:
\[
A_{S,\type,\beta}(x,y;\tilde{R}(x)) = \begin{cases}
    1 & \text{if } y = \type \text{ and } x \in S \text{ and } \tilde{R}(x)_\type \in \lambda(\beta), \\
    0 & \text{otherwise},
\end{cases}
\]
where $\tilde{R}(x)_\type$ denotes the $\type\tth$ coordinate of the stochastic vector $\tilde{R}(x)$.
Notice that evaluating $A_{S,\type,\beta}$ can be done in time $T$, since $\Lambda[0,1]$ is known in advance. 

Let $\tilde{R}: \X \to \stochasticVecs$ be $(\mathcal{A},\alpha)$-sample-access-OI predictor with respect to $\mD$, guaranteed by \Cref{thm:beyond}. This implies that for any $A_{S,\type,\beta} \in \mathcal{A}$,  
\[
\left| \Pr_{(x,\type^*(x)) \sim \mD} \left[ A_{S,\type,\beta}\left(x, \type^*(x);\tilde{R}(x) \right) = 1 \right] - 
\Pr_{\left(x, \tilde{\type}(x)\right) \sim \DX \times \tilde{R}(x)} \left[ A_{S,\type,\beta}\left(x, \tilde{\type}(x);\tilde{R}(x) \right) = 1 \right]
\right| \le \alpha.
\]  
Recall the definition of $S_{\type, \beta}$ w.r.t. $\tilde{R}$: 
\[
S_{\type, \beta} \defeq \left\{ x \in S \text{ such that }  \Pr_{\tilde{\type}(x) \sim \tilde{R}(x)}[\tilde{\type}(x) = \type] \in \lambda(\beta) \right\}.
\]
By definition of $A_{S,\type,\beta}$, this ensures that for any $S_{\type, \beta}$ and any $\type \in \setOfTypes$,
\[
\left| \Pr_{(x,\type^*(x)) \sim \mD} \left[ \type^*(x) = \type \land x \in S_{\type, \beta} \right] - 
\Pr_{\left(x, \tilde{\type}(x)\right) \sim \DX \times \tilde{R}(x)} \left[ \tilde{\type}(x) = \type \land x \in S_{\type, \beta} \right]
\right| \le \alpha.
\]    
Thus, $\tilde{R}$ satisfies $(\C,\alpha,\lambda)$-coordinate-wise-multi-calibration w.r.t. $\mD$.

Notice that the number of sets to consider is $\left|(S_{\type,\beta})_{\type \in \setOfTypes, \beta \in \Lambda[0,1]} \right| =  k / \lambda$, therefore $|\mathcal{A}| = \frac{k \cdot |\C|}{\lambda}$ and the claimed complexity bounds follow from \Cref{thm:beyond}.
\end{proof}

\paragraph{Replacing the linear dependence on $|\C|$ via weak agnostic learning.}
The running time bounds in \Cref{thm:learning_MA,thm:learning_MC} scale linearly with $|\C|$ due to the need to evaluate every distinguisher $A \in \mathcal{A}$ in each iteration, where each set $S \in \C$ corresponds to a distinguisher. 
To better understand this complexity, we revisit Theorem 20 in \citet{BeyondBer}, which decomposes the learning time into three components: sample generation, searching, and reweighting. The most computationally expensive step is searching, where, in each iteration, we must identify a distinguisher $A \in \mathcal{A}$ that differentiates between $R^*$ and the current predictor $\tilde{R}$. This is done via exhaustive search: iterating over all distinguishers in $\mathcal{A}$ and computing their distinguishing advantage. To estimate the distinguishing advantage, each distinguisher requires  
$O\left( (\log |\mathcal{A}| + \log N) / \eps^2 \right)$ 
samples per iteration, where $N$ is the iteration complexity. Since in the case of a discrete label distribution we have $N \le O(\log k / \eps^2)$, we obtain the complexity  
\[
\tilde{O} \left(\frac{|\mathcal{A}| \cdot \text{time}_{\mathcal{A}} \cdot \log k}{\eps^4} \right),
\]  
as presented in \Cref{thm:beyond}, where $\text{time}_{\mathcal{A}}$ is an upper bound on the time required to evaluate a single distinguisher.

To eliminate the need for exhaustive search over $\mathcal{A}$, we follow a standard approach and replace this step with a black-box auditing procedure. Given access to the current predictor $\tilde{R}$ and a sample from $\mD$, the auditor returns a set $S \in \C$ and a label $\type \in \setOfTypes$ such that the predictor violates the calibration or accuracy condition on $(S, t)$ by at least $\eps = \alpha$, if such a violation exists. If no such pair is found, the predictor is guaranteed to be $(\mathcal{C}, \alpha)$-multi-accurate. A similar approach applies to multi-calibration, where the auditor additionally returns a probability $\beta \in \Lambda[0,1]$. In this framework, the overall time complexity is determined by the number of correction steps and the running time of the auditor. 

In the case of coordinate-wise multi-accuracy and multi-calibration, the auditing step can be implemented via a weak agnostic learner. This follows from Theorem 4.2 in \citet{MC}, which shows how to achieve multi-calibration using weak agnostic learning, and thus, multi-accuracy as well. Although their setting is one-dimensional, our framework operates coordinate-wise rather than globally on the entire predictor. Consequently, the same argument applies, albeit with a multiplicative factor of $k$ in the running time. Thus, under the assumption that the class $\C$ admits an efficient weak agnostic learner, the total running time of the learning algorithm can be expressed in terms of the time complexity of this learner, effectively removing the linear dependence on $|\C|$. \Cref{thm:learning_MA_simplified,thm:learning_MC_simplified} are stated in terms of weak agnostic learning. We refer the reader to \citet{EfficientMultiClassMC} for a detailed discussion on how weak agnostic learning can be leveraged to implement efficient auditing procedures for multi-calibration and multi-accuracy.

\subsection{Decision rules}\label{sec:deferred_dec_rules}

We start with the proof of \Cref{clm:two_points_in_convex_combi_are_enough}.
\begin{claim}
    Let $\numOfVecs \in \N$, $\gamma_1,\dots,\gamma_\numOfVecs \in [0,1]$ s.t. $\sum_{i=1}^\numOfVecs \gamma_i = 1$, and $y_1,\dots,y_\numOfVecs \in \stochasticVecs$. Let $\eps \in [0,1]$ and assume that $f: \stochasticVecs \to \zo$ is $\eps$-close to affine. Then, 
    \[
    \left| \expectation_{f}\left[f\left(\sum_{i=1}^\numOfVecs \gamma_i y_i\right) \right]- \sum_{i=1}^\numOfVecs \gamma_i \expectation_{f}\left[f(y_i)\right] \right| \le 2 \eps.
    \]
\end{claim}
\begin{proof}
First, let us rewrite the second sum: 
\begin{align*}
 \sum_{i=1}^\numOfVecs \gamma_i \expectation[f(y_i)] = \gamma_\numOfVecs \expectation[f(y_\numOfVecs)] +\sum_{i=1}^{\numOfVecs-1} \gamma_i \expectation[f(y_i)] & = \gamma_\numOfVecs \expectation[f(y_\numOfVecs)] + \left( \sum_{i=1}^{\numOfVecs-1} \gamma_i \right) \cdot \sum_{i=1}^{\numOfVecs-1} \frac{\gamma_i}{\sum_{i=1}^{\numOfVecs-1} \gamma_i} \expectation[f(y_i)] \\ & = \gamma_\numOfVecs \expectation[f(y_\numOfVecs)] + \left( 1 - \gamma_\numOfVecs \right) \cdot \sum_{i=1}^{\numOfVecs-1} \frac{\gamma_i}{1 - \gamma_\numOfVecs } \expectation[f(y_i)].
\end{align*}

Since $\left(\gamma_i / 1 - \gamma_\numOfVecs\right)_{i \in [\numOfVecs-1]}$ sums to $1$, the expression $\sum_{i=1}^{\numOfVecs-1} \left(\gamma_i / 1 - \gamma_\numOfVecs\right) \cdot y_i$ forms a convex combination of vectors in $\stochasticVecs$. As $\stochasticVecs$ is a convex set, this combination remains within $\stochasticVecs$.  
Thus, the assumption from \Cref{def:close_to_affine} applies, yielding that
\[
\left|\sum_{i=1}^{\numOfVecs-1} \frac{\gamma_i}{1 - \gamma_\numOfVecs} \expectation[f(y_i)] -  \expectation\left[f\left(\sum_{i=1}^{\numOfVecs-1} \frac{\gamma_i}{1 - \gamma_\numOfVecs} \cdot y_i \right) \right] \right| \le \eps.
\]
This means that 
\begin{align}\label{eq:after_one_epsilon}
& \nonumber \left| \expectation\left[f\left(\sum_{i=1}^\numOfVecs \gamma_i y_i\right)\right] - \sum_{i=1}^\numOfVecs \gamma_i \expectation[f(y_i)] \right| = \\
& \nonumber \left| \expectation\left[f\left(\sum_{i=1}^\numOfVecs \gamma_i y_i\right)\right] - \gamma_\numOfVecs \expectation[f(y_\numOfVecs)] - \left( 1 - \gamma_\numOfVecs \right) \cdot \sum_{i=1}^{\numOfVecs-1} \frac{\gamma_i}{1 - \gamma_\numOfVecs } \expectation[f(y_i)] \right| \le  \\ & \left( 1 - \gamma_\numOfVecs \right) \cdot \eps + \left| \expectation\left[f\left(\sum_{i=1}^\numOfVecs \gamma_i y_i\right)\right] - \gamma_\numOfVecs \expectation[f(y_\numOfVecs)] - 
\left( 1 - \gamma_\numOfVecs \right) \cdot 
\expectation\left[f\left(\sum_{i=1}^{\numOfVecs-1} \frac{\gamma_i}{1 - \gamma_\numOfVecs} \cdot y_i \right)\right]  \right|.
\end{align}
Following the same arguments, we rewrite the first sum:
\begin{align*}
& f\left(\sum_{i=1}^\numOfVecs \gamma_i y_i\right) = f\left(\gamma_\numOfVecs \cdot y_\numOfVecs + \left(1-\gamma_\numOfVecs \right) \cdot \sum_{i=1}^{\numOfVecs-1} \frac{\gamma_i }{1-\gamma_\numOfVecs} \cdot y_i \right),
\end{align*}
and by the assumption,
\begin{align*}
 \Bigg| & \expectation\left[f\left(\gamma_\numOfVecs \cdot y_\numOfVecs + \left(1-\gamma_\numOfVecs \right) \cdot \sum_{i=1}^{\numOfVecs-1} \frac{\gamma_i}{1-\gamma_\numOfVecs} \cdot y_i \right)\right] \\ & - \gamma_\numOfVecs \cdot \expectation[f(y_\numOfVecs)] - \left(1-\gamma_\numOfVecs \right) \cdot \expectation\left[f\left(\sum_{i=1}^{\numOfVecs-1} \frac{\gamma_i}{1-\gamma_\numOfVecs} \cdot y_i\right) \right] \Bigg| \le \eps.
\end{align*}
Plugging in into \Cref{eq:after_one_epsilon},
\begin{align*}
& \left| \expectation\left[f\left(\sum_{i=1}^\numOfVecs \gamma_i y_i\right)\right] - \sum_{i=1}^\numOfVecs \gamma_i \expectation[f(y_i)] \right| \\ & \le 
 \left( 1 - \gamma_\numOfVecs \right) \cdot \eps + \eps +  \Bigg|  \gamma_\numOfVecs \cdot \expectation[f(y_\numOfVecs)] + \left(1-\gamma_\numOfVecs \right) \cdot \expectation\left[f\left(\sum_{i=1}^{\numOfVecs-1} \frac{\gamma_i}{1-\gamma_\numOfVecs} \cdot y_i\right)\right]  \\& - \gamma_\numOfVecs \expectation[f(y_\numOfVecs)] - 
\left( 1 - \gamma_\numOfVecs \right) \cdot 
\expectation\left[f\left(\sum_{i=1}^{\numOfVecs-1} \frac{\gamma_i}{1 - \gamma_\numOfVecs} \cdot y_i \right)\right]  \Bigg| \\&  = \left(2- \gamma_\numOfVecs\right)\cdot \eps \le 2\eps.
\end{align*}
\end{proof}

Next, we derive \Cref{cor:close_affine_function}.
\begin{corollary}
    If $f: \stochasticVecs \to \zo$ is $\eps$-close to affine, then there exists an affine randomized predicate $f': \stochasticVecs \to \zo$ such that $\forall y \in \stochasticVecs$,
    \[
    |\expectation_{f}[f(y)] - \expectation_{f'}[f'(y)]| \le 2\eps.
    \]
\end{corollary}
\begin{proof}
Recall that $e_i$ denotes the $i\tth$ unit vector in $k$ dimensions. Any vector $y$ can be expressed as $y = \sum_{i \in [k]} y_i \cdot e_i$, where $y_i$ is its $i\tth$ coordinate.  
We define $f'$ as follows:  
\[
f'(y) \sim \text{Ber} \left( \sum_{i \in [k]} y_i \cdot \mathbb{E}[f(e_i)] \right).
\]
Clearly, $f'$ is affine, and the corollary follows from \Cref{clm:two_points_in_convex_combi_are_enough}.
\end{proof}

Finally, we provide the proof of \Cref{clm:y_and_y'_are_far}.
\begin{claim}
Let $\eps \in [0,1]$ and $M \in \R^+$. Let $f: \stochasticVecs \to \zo$ be a randomized predicate such that:
\begin{itemize}
    \item $f$ is $M$-Lipschitz;
    \item $f$ is $\eps$-far from affine: that is, there exists $y,y' \in \stochasticVecs$ and $\gamma \in (0,1)$ such that \[
    \left| \expectation_{f}\left[f\left(\gamma y + (1-\gamma) y'\right) \right]- \left( \gamma \expectation_{f}\left[f(y)\right] + (1-\gamma) \expectation_{f}\left[f(y')\right] \right) \right| > \eps.
    \]
\end{itemize}
Then, $|| y - y' ||_{\infty} \ge \eps / M$. 
\end{claim}
\begin{proof}
    First, since $f$ is $M$-Lipschitz,
    \[
    \left|\expectation\left[f\left(\gamma y + (1-\gamma)y' \right)\right] - \expectation\left[f(y)\right] \right| \le M ||\gamma y + (1-\gamma)y' - y|| = M(1-\gamma)||y-y'|| \le M||y-y'||,
    \]
    and following the same argument, 
    \[
    \left|\expectation\left[f\left(\gamma y + (1-\gamma)y' \right)\right] - \expectation\left[f(y')\right] \right| \le  M||y-y'||.
    \]
This implies, by the assumption,
\begin{align*}
    \eps  < &\left| \expectation\left[f\left(\gamma y + (1-\gamma) y'\right)\right] - \left( \gamma \expectation\left[f(y)\right] + (1-\gamma) \expectation\left[f(y')\right] \right) \right| \\ = 
    & \left| \gamma \left(\expectation\left[f\left(\gamma y + (1-\gamma)y' \right)\right] - \expectation\left[f(y)\right] \right) + (1-\gamma) \left( \expectation\left[f\left(\gamma y + (1-\gamma)y' \right)\right] - \expectation\left[f(y')\right]  \right) \right| \\ \le 
    & \gamma M||y-y'|| + (1-\gamma) M||y-y'|| = M ||y-y'||,
\end{align*}
and the claim follows.
\end{proof}

\subsection{Indistinguishability}\label{sec:deferred_indistin}

First, we provide the proof of \Cref{lem:existence_of_X1}.
\begin{lemma}
    Let $\X$ be a domain, and let $\UX$ denote the uniform distribution over $\X$. Assume one-way functions exist w.r.t. security parameter $\kappa \ge \log |\X|$. For any constant $\gamma \in (0,1)$, there exists a subset $\X_1 \subset \X$ such that
    \begin{itemize}
        \item $\left| \Pr_{x \sim \UX}\left[x \in \X_1 \right] - \gamma \right| = \negli(\kappa)$;
        \item $\X_1$ is $\kappa$-indistinguishable w.r.t. $\UX$;
        \item There exists a secret key such that, given the key, membership in $\X_1$ is efficiently computable.
    \end{itemize}
\end{lemma}
\begin{proof}
    It is well-known in cryptography that the existence of one-way functions implies the existence of pseudorandom functions (PRFs) \citep{HILL99,GGM86}. In fact, we will show that weak-PRFs are sufficient for the construction, though the existence of OWFs remains a necessary assumption in both cases. 
    
    Recall that $\kappa \ge \log |\X| = n$, where $n$ is the length of the Boolean string representing any $x \in \X$. 
    Let $F_s: \zo^n \to \zo^\ell$ denote the $\poly(\kappa)$-size circuit computing $F_s$, where $\ell = \lceil \log(1/\gamma) + \kappa \rceil$ and $s \in \zo^\kappa$ denotes the secret key, which is hard-wired to the circuit. No efficient computation can access $s$ or learn anything about it. 
    
    $F_s$ defines membership in the subset $\X_1$ in the following manner: On input $x \sim \UX$, we interpret the output $F_s(x)$ as a real number in $[0,1]$. We define $x \in \X_1$ if and only if $F_s(x) \le \gamma$.\footnote{For convenience, we write $F_s(x) \le \gamma$ to indicate that the real value in $[0,1]$ represented by the binary string $F_s(x)$ is at most $\gamma$.} By our choice of $\ell$, the precision error of this length-$\ell$ binary string is at most $2^{-\ell}$, therefore
    \[
    \left| \Pr_{x \sim \UX}\left[x \in \X_1 \right] - \gamma \right| = \negli(\kappa).
    \]
The $\kappa$-indistinguishability of $\X_1$ follows from the pseudorandomness of $F_s$. Consider a probabilistic $\poly(\kappa)$-time algorithm $\mathcal{A}$ such that
    \begin{equation} \label{eq:indistin_of_x1}
    \left| \Pr_{\mathcal{A}, x \sim \UX}\left[ \mathcal{A}(x) = 1 \mid x \in \X_1 \right] - \Pr_{\mathcal{A}, x \sim \UX}\left[ \mathcal{A}(x) = 1 \mid x \in \X \setminus \X_1 \right] \right| = \eps(\kappa),
    \end{equation}
    where $\eps(\cdot)$ is a non-negligible function in $\kappa$. Recall that, since $\gamma$ is a constant, $\Pr[x \in \X_1] \in (0,1)$, hence \Cref{eq:indistin_of_x1} is well-defined.
We reduce $\mathcal{A}$ to an adversary $\mathcal{B}$ for the pseudorandom function $F_s$. Conceptually, the adversary $\mathcal{B}$ has oracle access to an unknown function $H$, which can either be the PRF or a truly random function. The goal of $\mathcal{B}$ is to distinguish between the two, using its oracle access to $\mathcal{A}$.
\begin{itemize}
    \item \textbf{Input:} oracle access to $\mathcal{A}: \zo^n \to \zo$, $H: \zo^n \to \zo^\ell$. 
    
    \item $\mathcal{B}$ samples inputs $x_1, \dots, x_N \sim \UX$ for $N = O\left(1/\eps^2(\kappa)\right)$. Let $S = \{x_i\}_{i \in [N]}$.

    \item $\mathcal{B}$ makes $N$ oracle calls to $H$ and defines
    \[
        S_1 = \left\{ x_i \in S : H(x_i) \le \gamma \right\},
        \qquad
        S_0 = S \setminus S_1.
    \]
    If either $S_0$ or $S_1$ is empty, $\mathcal B$ returns $0$.
    
    \item $\mathcal{B}$ makes $N$ oracle calls to $\mathcal{A}$ using fresh randomness, and computes
    \[
        \hat p_1 \defeq \frac{1}{|S_1|}\sum_{x_i \in S_1} \mathcal{A}(x_i),
        \qquad
        \hat p_0 \defeq \frac{1}{|S_0|}\sum_{x_i \in S_0} \mathcal{A}(x_i).
    \]
    $\mathcal{B}$ returns $1$ if $|\hat p_1-\hat p_0|>\eps(\kappa)/2$, and $0$ otherwise.

    
\end{itemize}
By construction, $\mathcal{B}$ is a probabilistic algorithm that runs in $\poly(\kappa)$ time. Moreover, observe that $\mathcal{B}$ queries its oracle $H$ only on uniformly random inputs, which allows us to prove weak-PRF security as commented above.
To demonstrate that $\mathcal{B}$ breaks the security of $F_s$, we proceed by showing that
\[
\left| \Pr_{s,\mathcal{B},\mathcal{A}} \left[\mathcal{B}^{\mathcal{A},F_s}(1^\kappa) = 1 \right] - \Pr_{G_\kappa,\mathcal{B},\mathcal{A}} \left[\mathcal{B}^{\mathcal{A},G_\kappa}(1^\kappa) = 1 \right] \right| \ge \frac{1}{3},
\]
where $G_\kappa: \zo^n \to \zo^\ell$ is a truly random function, namely, it is uniformly distributed over the set of all functions mapping $n(\kappa)$-bit-long strings to $\ell(\kappa)$-bit-long strings. 
Notice that, when $H = F_s$,
\begin{align*}
& \left| \Pr\left[ \mathcal{A}(x_i) = 1 | x_i \in S_1 \right] - \Pr\left[ \mathcal{A}(x_i) = 1 | x_i \in S\setminus S_1 \right] \right| = \\
& \left| \Pr\left[ \mathcal{A}(x_i) = 1 | F_s(x_i) \le \gamma \right] - \Pr\left[ \mathcal{A}(x_i) = 1 | F_s(x_i) > \gamma \right] \right|.
\end{align*}

Hence, by a Chernoff bound, applying \Cref{eq:indistin_of_x1} to the samples
$x_i \sim \UX$ implies that, for a sufficiently large constant in the choice
$N=O(1/\eps^2(\kappa))$, with probability at least $2/3$,
\[
    |\hat p_1-\hat p_0| > \frac{\eps(\kappa)}{2}.
\]
Therefore,
\[
    \Pr_{s,\mathcal{B},\mathcal{A}}
    \left[\mathcal{B}^{\mathcal{A},F_s}(1^\kappa)=1\right]
    \ge \frac{2}{3}.
\]

On the other hand, when $H=G_\kappa$, the event $G_\kappa(x)\le \gamma$ is
independent of $x$ and of the output of $\mathcal{A}(x)$. Therefore,
\[
\Pr\left[ \mathcal{A}(x)=1 \mid G_\kappa(x)\le \gamma \right]
=
\Pr\left[ \mathcal{A}(x)=1 \mid G_\kappa(x)> \gamma \right].
\]
Again by a Chernoff bound, with probability at least $2/3$,
\[
    |\hat p_1-\hat p_0| \le \frac{\eps(\kappa)}{2}.
\]
Thus,
\[
\Pr_{G_\kappa,\mathcal{B},\mathcal{A}}
    \left[\mathcal{B}^{\mathcal{A},G_\kappa}(1^\kappa)=1\right]
    \le \frac{1}{3}.
\]
All in all, the distinguishing advantage of $\mathcal{B}$ is at least $1/3$,
which contradicts the assumed security of $F_s$.

\end{proof}

Next, we restate \Cref{clm:subsets_maintain_gamma_fraction} together with its proof. 
\begin{claim}
    Let $\X$ be a domain, and let $\UX$ denote the uniform distribution over $\X$. 
    Assume $\X$ contains a $\kappa$-indistinguishable subset $\X_1$ as per \Cref{def:indistin} w.r.t. a security parameter $\kappa \ge \log|\X|$. For any $\poly(\kappa)$-time algorithm $\mathcal{A}: \X \to \zo$,
    if $\Pr[\mathcal{A}(x) = 1]$ is non-negligible in $\kappa$, then
    \[
    \left| \Pr_{\substack{x \sim \UX \\ \mathcal{A}\ randomness }}[x \in \X_1 | \mathcal{A}(x) = 1] - \Pr_{x \sim \UX}[x \in \X_1] \right| = \negli(\kappa).
    \]
\end{claim}
\begin{proof}
    Let $\mathcal{A}: \X \to \zo$ be a $\poly(\kappa)$-time algorithm and let $\X_2 \defeq \X \setminus \X_1$. Let us denote 
    \[
    \tilGa \defeq \Pr_{\substack{x \sim \UX \\ \mathcal{A}\ randomness }}\left[x \in \X_1 \mid \mathcal{A}(x) = 1 \right] \text{ and } \gamma' \defeq \Pr_{x \sim \UX}\left[x \in \X_1 \right].
    \]
    The advantage of $\mathcal{A}$ in distinguishing membership in $\X_1$ is:
    \begin{align*}
        & \left| \Pr\left[\mathcal{A}(x) = 1 \mid x \in \X_1 \right] - \Pr\left[\mathcal{A}(x) = 1 \mid x \in \X_2 \right] \right| = \\
        & \Pr[\mathcal{A}(x) = 1] \cdot \left| 
        \frac{\Pr\left[x \in \X_1 \mid \mathcal{A}(x) = 1 \right]}{\Pr\left[x \in \X_1 \right]} - 
        \frac{\Pr\left[x \in \X_2 \mid \mathcal{A}(x) = 1 \right]}{\Pr\left[x \in \X_2 \right]} \right|  = \\
        & \Pr[\mathcal{A}(x) = 1] \cdot \left| \frac{\tilGa}{\gamma'} - \frac{1 - \tilGa}{1 - \gamma'}  \right| \ge 4 \Pr[\mathcal{A}(x) = 1]  \cdot \left| \tilGa - \gamma' \right|,
    \end{align*}
where the last inequality follows from the fact that for all $\gamma' \in [0,1]$, we have $\gamma'(1 - \gamma') \leq \frac{1}{4}$. Recall that, by assumption, $\Pr[\mathcal{A}(x) = 1]$ is non-negligible in $\kappa$. Since the advantage of any efficient algorithm in recognizing membership in $\X_1$ is negligible, we conclude that $\left| \tilGa - \gamma' \right| = \negli(\kappa)$.
\end{proof}


\end{document}